\documentclass[twoside]{article}

\usepackage[accepted]{aistats2023}
\usepackage{apacite}
\newcommand{\citet}[1]{\citeauthor{#1} \shortcite{#1}}
\newcommand{\citep}{\cite}
\newcommand{\citealp}[1]{\shortciteauthor{#1} \citeyear{#1}}

\usepackage{colortbl}
\usepackage{graphicx}
\graphicspath{{figures/}}
\usepackage{amsmath}
\usepackage[table,xcdraw]{xcolor}
\usepackage{multirow}
\usepackage{pifont}

\usepackage{algpseudocode}
\usepackage{algorithm} 
\usepackage{diagbox}
\usepackage{url}
\usepackage[symbol]{footmisc}

\begin{document}

\twocolumn[

\aistatstitle{Introducing Intermediate Domains for Effective Self-Training during Test-Time}

\aistatsauthor{ Robert A. Marsden\footnote[1]{Equal contribution.} \And Mario D\"obler\footnotemark[1] \And  Bin Yang }

\aistatsaddress{ University of Stuttgart \And  University of Stuttgart \And University of Stuttgart } ]

\footnotetext[1]{Equal contribution.}

\begin{abstract}
Experiencing domain shifts during test-time is nearly inevitable in practice and likely results in a severe performance degradation. To overcome this issue, test-time adaptation continues to update the initial source model during deployment. A promising direction are methods based on self-training which have been shown to be well suited for gradual domain adaptation, since reliable pseudo-labels can be provided. In this work, we address two problems that exist when applying self-training in the setting of test-time adaptation. First, adapting a model to long test sequences that contain multiple domains can lead to error accumulation. Second, naturally, not all shifts are gradual in practice. To tackle these challenges, we introduce GTTA. By creating artificial intermediate domains that divide the current domain shift into a more gradual one, effective self-training through high quality pseudo-labels can be performed. To create the intermediate domains, we propose two independent variations: mixup and light-weight style transfer. We demonstrate the effectiveness of our approach on the continual and gradual corruption benchmarks, as well as ImageNet-R. To further investigate gradual shifts in the context of urban scene segmentation, we publish a new benchmark: CarlaTTA. It enables the exploration of several non-stationary domain shifts. \footnote[2]{Code is available at: \url{https://github.com/mariodoebler/test-time-adaptation}}
\end{abstract}


\section{Introduction} \label{sec:introduction}
\begin{figure}[t]
\centering
\def\svgwidth{210pt}    
\begingroup%
  \makeatletter%
  \providecommand\color[2][]{%
    \errmessage{(Inkscape) Color is used for the text in Inkscape, but the package 'color.sty' is not loaded}%
    \renewcommand\color[2][]{}%
  }%
  \providecommand\transparent[1]{%
    \errmessage{(Inkscape) Transparency is used (non-zero) for the text in Inkscape, but the package 'transparent.sty' is not loaded}%
    \renewcommand\transparent[1]{}%
  }%
  \providecommand\rotatebox[2]{#2}%
  \newcommand*\fsize{\dimexpr\f@size pt\relax}%
  \newcommand*\lineheight[1]{\fontsize{\fsize}{#1\fsize}\selectfont}%
  \ifx\svgwidth\undefined%
    \setlength{\unitlength}{195.57490601bp}%
    \ifx\svgscale\undefined%
      \relax%
    \else%
      \setlength{\unitlength}{\unitlength * \real{\svgscale}}%
    \fi%
  \else%
    \setlength{\unitlength}{\svgwidth}%
  \fi%
  \global\let\svgwidth\undefined%
  \global\let\svgscale\undefined%
  \makeatother%
  \begin{picture}(1,0.47618949)%
    \lineheight{1}%
    \setlength\tabcolsep{0pt}%
    \put(0,0){\includegraphics[width=\unitlength,page=1]{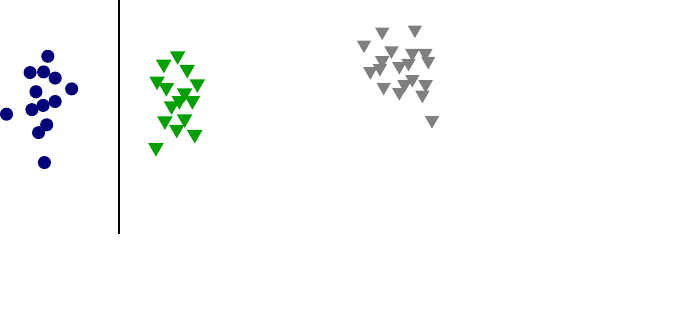}}%
    \put(0.16060344,0.07314371){\makebox(0,0)[lt]{\lineheight{1.25}\smash{\begin{tabular}[t]{l}$\beta_0$\end{tabular}}}}%
    \put(0.13557869,0.01009121){\makebox(0,0)[lt]{\lineheight{1.25}\smash{\begin{tabular}[t]{l}$t = 0$\end{tabular}}}}%
    \put(0,0){\includegraphics[width=\unitlength,page=2]{GDA.pdf}}%
    \put(0.50777342,0.0731436){\makebox(0,0)[lt]{\lineheight{1.25}\smash{\begin{tabular}[t]{l}$\beta_0$\end{tabular}}}}%
    \put(0.48071504,0.01076918){\makebox(0,0)[lt]{\lineheight{1.25}\smash{\begin{tabular}[t]{l}$t = 1$\end{tabular}}}}%
    \put(0,0){\includegraphics[width=\unitlength,page=3]{GDA.pdf}}%
    \put(0.85228317,0.07314369){\makebox(0,0)[lt]{\lineheight{1.25}\smash{\begin{tabular}[t]{l}$\beta_0$\end{tabular}}}}%
    \put(0.82585144,0.0107691){\makebox(0,0)[lt]{\lineheight{1.25}\smash{\begin{tabular}[t]{l}$t = 2$\end{tabular}}}}%
    \put(0,0){\includegraphics[width=\unitlength,page=4]{GDA.pdf}}%
  \end{picture}%
\endgroup%
  
\caption{Illustration of a gradually evolving domain. In the beginning $t=0$, perfect classification is possible. At time step $t=1$, the domain has gradually changed and the original source decision boundary $\beta_0$ still separates both classes. When the domain further evolves ($t=2$), the original decision boundary cannot separate the data anymore. When updating the model through self-training at time step $t=1$, a good classification would still be possible at time step $t=2$, as indicated by the dashed line.}
\label{fig:gradual_motivation}
\end{figure}
Deep neural networks achieve remarkable performance under the assumption that training and test data originate from the same distribution. However, when a neural network is deployed in the real world, this assumption is often violated. This effect is known as data shift \shortcite{quinonero2008dataset} and leads to a potentially large drop in performance on the test data. While it is possible to improve robustness and generalization directly during training \shortcite{hendrycks2019augmix, hendrycks2021many, muandet2013domain, tobin2017domain, tremblay2018training}, the effectiveness remains limited due to the wide range of potential data shifts \shortcite{mintun2021interaction} that are unknown during training. Thus, another area of research, namely test-time adaptation (TTA), follows the idea to adapt the pre-trained source model during deployment, as the encountered test data provides information about the current distribution shift.

Recent work on TTA focuses on the setting where the model only has to adapt to a single test domain. Needless to say, in practice this setting is very unlikely; it is much more likely that a model encounters different domains without the knowledge when a change occurs. \citealp{wang2022continual} denotes the setting where a model is adapted during deployment to a sequence of test domains as \textit{continual test-time adaptation}. Due to potentially infinitely long test sequences and the encounter of different domain shifts, test-time adaptation, which is usually based on self-training and entropy minimization, is prone to error accumulation \shortcite{wang2022continual}. 

Clearly, the larger a shift is, the more likely it becomes that a model introduces errors. In the case of self-training, this likely results in an unsuccessful model adaptation due to the lack of reliable pseudo-labels \shortcite{kumar2020understanding}. On the contrary, \citealp{kumar2020understanding} showed both theoretically and empirically that a model can be adapted successfully if the experienced shifts are small enough. Hence, adaptation to large shifts can be successful if divided into smaller gradual shifts, as illustrated in Figure~\ref{fig:gradual_motivation}.

Looking at the nature of shifts in reality, for many applications they do not occur abruptly, but evolve gradually over time. While the change from day to night is only one example, \citealp{kumar2020understanding} mentions, among others, evolving road conditions \shortcite{bobu2018adapting} and sensor aging \shortcite{vergara2012chemical}. Of course, gradual shifts are not given in all settings or the gap of the experienced gradual shift is still too large for a successful model adaptation. Therefore, we propose to leverage source data to artificially create intermediate domains where, optimally, correct labels can be utilized to prevent the incorporation of additional errors. Even though requiring source data can be a limitation, we argue that having access to the initial source data is commonly the case. Now, for the creation of the intermediate domains we suggest two independent approaches: the first is based on mixup where the intermediate domains are created by linearly interpolating source and test images. The second idea uses a content-preserving light-weight style transfer model that is adapted online to new target styles. Since mixup and style transfer have their limitations and only mitigate the current domain gap, we rely on self-training to close the remaining gap. Assuming that mixup or style transfer moves the model closer to the test distribution, better self-training through more reliable pseudo-labels can be performed.

To demonstrate the effectiveness of our approach, we consider the continual and gradual corruption benchmark, as well as ImageNet-R. Due to the lack of datasets containing non-stationary domain shifts, we introduce and publish a new benchmark for the task of urban scene segmentation: CarlaTTA. It includes various non-stationary domain shifts in the setting of autonomous driving. We achieve new state-of-the-art results on all benchmarks. We summarize our main contributions as follows: 

\begin{itemize}
 \item We introduce a new framework \textit{Gradual Test-time Adaptation} (GTTA), which conducts effective self-training by converting the current arbitrary domain shift into a gradual one. This is achieved by generating artificial intermediate domains using either mixup or light-weight style transfer.
 \item We publish a new benchmark for urban scene segmentation that enables the exploration of several non-stationary domain shifts during test-time in the field of autonomous driving.
\end{itemize}

\section{Related Work}
\textbf{Unsupervised Domain Adaptation}
Recently, there has been a growing interest in mitigating the distributional discrepancy between two domains using unsupervised domain adaptation (UDA). Common approaches for UDA try to align either the input space \shortcite{CYCADA, CACE, ACE}, the feature space \shortcite{CAG, marsden2022contrastive}, the output space \shortcite{AdaptSegNet, PatchAlign}, or several spaces in parallel \shortcite{Bidirectional, FDA}. One line of work relies on adversarial learning, where a domain classifier tries to discriminate whether some feature maps \shortcite{DANN} or network outputs \shortcite{AdaptSegNet, ADVENT} belong to the source or target domain. It is also possible to exploit adversarial learning or adaptive instance normalization (AdaIN) \shortcite{AdaIN} for transferring the target style to source images \shortcite{CYCADA, CACE}. Lately, self-training has gained a lot of attraction \shortcite{DACS, IAST, ProDA, hoyer2022daformer, CCM}. Self-training utilizes a pre-trained (source) model to create predictions for the unlabeled target data. These predictions can then be treated as pseudo-labels to minimize, for example, the cross-entropy. Since high quality pseudo-labels are essential to this approach, most methods differ in how they select or create reliable pseudo-labels.

\textbf{One-shot Unsupervised Domain Adaptation}
As pointed out in \citealp{ASM}, even collecting unlabeled target data can be challenging. Therefore, \citealp{ASM} introduced one-shot UDA, where only one single target image is available during the model adaptation. To address this problem, \citealp{ASM} extends the adaptive instance normalization framework of \citealp{AdaIN} with a variational autoencoder. By selecting styles for which the segmentation model is uncertain, the domain gap is mitigated. Differently, \citealp{SM-PPM} uses a style mixing component within the segmentation model and further adds patch-wise prototypical matching. 

\textbf{Test-time Adaptation}
Although generalizing to any test distribution would solve many problems, the lack of information about the test environment during training imposes a great challenge. However, during model deployment, one can gain some insight into the test distribution by using the current test sample(s). This circumstance is also exploited in recent work, where \citealp{schneider2020improving} showed that even adapting the batch normalization (BN) statistics during test-time can significantly improve the performance on corrupted data. More sophisticated approaches perform source model optimization during test-time. For example, \citealp{TENT} update the BN layers by entropy minimization. \citealp{MEMO} create an ensemble prediction through test-time augmentation \shortcite{krizhevsky2009learning} and then minimize the entropy with respect to all parameters. Other methods rely on self-supervised learning, using either pre-text tasks to adapt the model \shortcite{TTT, liu2021ttt++, MT3} or apply contrastive learning \shortcite{chen2022contrastive}. Recent works make use of diversity regularizers \shortcite{liang2020we, mummadi2021test} to prevent the collapse to trivial solutions potentially caused by confidence maximization.

\textbf{Continual Test-time Adaptation}
Continual test-time adaptation considers online TTA with continually changing target domains. While some of the existing methods can be applied to the continual setting, such as the online version of TENT \shortcite{TENT}, they are often prone to error accumulation due to miscalibrated predictions \shortcite{wang2022continual}. CoTTA \shortcite{wang2022continual} uses weight and augmentation-averaged predictions to reduce error accumulation and stochastic restore to circumvent catastrophic forgetting \shortcite{mccloskey1989catastrophic}.

\textbf{Gradual Domain Adaptation}
Recent work has indicated that when the domain discrepancy is too large, adapting a model through self-training can be very challenging due to noisy pseudo-labels \shortcite{kumar2020understanding}. Therefore, numerous methods consider the setting of gradual domain adaptation \shortcite{hoffman14, IDOL, AuxSelfTrain}, where several intermediate domains exist between source and target. While some of the proposed approaches successively adapt the model using adversarial learning \shortcite{wulfmeier, bobu2018adapting}, it has been shown that self-training can be very powerful in this setting \shortcite{kumar2020understanding}.

\section{Methodology} \label{sec:gda}
\begin{figure}[t]
\centering
\def\svgwidth{235pt}    
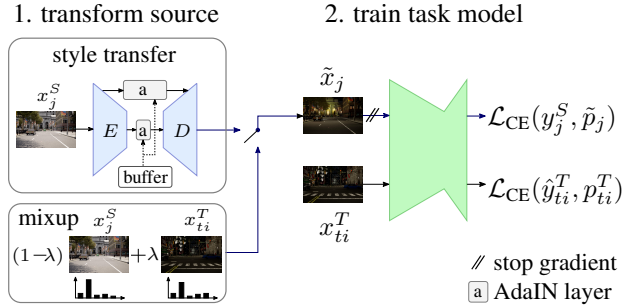  
\caption{Our framework GTTA performs test-time adaptation in two steps. First, utilizing the current test samples, we create intermediate domains by transforming a batch of source samples. This is accomplished by relying either on mixup or style transfer based on AdaIN layers. The transformed samples and the corresponding source labels are subsequently used to minimize a cross-entropy loss $\mathcal{L}_\text{CE}^S(y^S,\tilde{p}^S)$. Second, self-training is performed with filtered pseudo-labels by minimizing the cross-entropy $\mathcal{L}_\text{CE}^T(\hat{y}_{t}^T,p_{t}^T)$.}
\label{fig:framework}
\end{figure}

Since in many practical applications environmental conditions can change over time, a model pre-trained on source data $(\mathcal{X}^S, \mathcal{Y}^S)$ can quickly become sub-optimal for the current test data $x_t^T$ at time step $t$. Online test-time adaptation counteracts the performance deterioration by updating the model based on the current test data $x_t^T$. As already presented by the theory for gradual domain adaptation \shortcite{kumar2020understanding}, self-training can be particularly successful when guaranteed that the domain shift is small enough. Clearly, in reality this is not always given, since the domain shift can occur at different rates and severities. Therefore, in this work, we present a framework, depicted in Figure \ref{fig:framework}, that performs TTA in two steps: First, current test images $x_t^T$ and a batch of randomly sampled source images $x^S$ are utilized to generate an intermediate domain. Since we rely on content-preserving methods to create intermediate domains, the transformed images $\tilde x(x_t^T, x^S)$ and the corresponding source labels $y^S$ are used to minimize the cross-entropy loss $\mathcal{L}_\text{CE}^S(y^S,\tilde p)$, where $\tilde p$ are the softmax predictions of the transformed images $\tilde x$. In a second step, reliable self-training can now be carried out to close the remaining domain gap. This is accomplished by minimizing the cross-entropy $\mathcal{L}_\text{CE}^T(\hat{y}_{t}^T,p_{t}^T)$, using sharpened softmax outputs $\hat{y}_t^T$ from the current test images $x_t^T$ and the corresponding softmax predictions $p_{t}^T$.

\subsection{Generating intermediate domains}
To generate intermediate domains, we propose two ideas: mixup known for improving robustness \shortcite{zhang2017mixup} and content-preserving light-weight style transfer. Either mixup or style transfer can be chosen depending on the type of domain shifts and computational requirements.

\textbf{Mixup}
The original idea of mixup is that linear interpolations in the input space should lead to linear interpolations in the output space. Since we do not want to introduce additional label-noise during test-time, we adapt the original idea of mixup in the sense that we do not interpolate labels. Instead we only rely on our noise-free source labels and linearly interpolate between source and test samples to close the gap between these two domains:
\begin{equation}
    \tilde{x}_j = (1 - \lambda_\text{mix})x_j^S + \lambda_\text{mix}x_{ti}^T.
\end{equation}
To reduce the mixup of samples belonging to different classes, we interpolate source sample $x_j^S$ with the test sample $x_{ti}^T$ from the current test batch $x_{t}^T$ that has the highest similarity in terms of the largest dot product in the output softmax probability space. Investigations about the mixup strength $\lambda_\text{mix}$ are presented in Appendix \ref{appendix:mixup_strength}.

\textbf{Style transfer}
Another possibility to create intermediate domains is to leverage style transfer. However, performing style transfer during test-time imposes some challenges. It should be of light weight to enable real-time processing and the network should be easily trainable during test-time, even when only having one test sample at a time. While \citealp{kim2021deep} introduces a method for photo-realistic style transfer during test-time, it is not suitable for our setting since it takes tens of seconds to transfer one single image-pair. This is similar to style transfer based on adversarial learning \shortcite{isola2017image, CycleGAN}, which can be unstable during training. Therefore, we follow \citealp{AdaIN} and use a VGG19 based network that performs style transfer through an adaptive instance normalization (AdaIN) layer. This layer assumes that the style is mostly contained in the first two moments. In our case, the AdaIN layer re-normalizes a content feature map $z_j^S$ belonging to source image $x_j^S$ to have the same channel-wise mean $\mu$ and standard deviation $\sigma$ as a style feature map $z_{ti}^T$ extracted from the $i$-th test image $x_{ti}^T$:
\begin{equation}
\tilde{z}_j = \sigma(z_{ti}^T) \frac{z_j^S - \mu(z_j^S)}{\sigma(z_j^S)} + \mu(z_{ti}^T).
\label{eq:adain}
\end{equation}
Since $z_j^S$ and $z_{ti}^T$ are extracted with an ImageNet pre-trained and frozen VGG19 encoder, the network only needs to be trained with respect to the decoder's parameters. Now, let $E$, $D$, and $\tilde{x}_j = D(\tilde{z}_j)$ be the encoder, the decoder, and the transferred source image, respectively, then the loss minimized by the decoder can be written as:
\begin{align}
\mathcal{L} &= \lambda_\mathrm{s} \sum_{l=1}^L \bigg[ \mathrm{MSE}\Big( \mu \big (E_l(\tilde{x}_j) \big), \mu \big(E_l(x_{ti}^T) \big) \Big) \label{eq:adain_loss}
\\ 
&+ \mathrm{MSE} \Big( \sigma \big(E_l(\tilde{x}_j) \big ), \sigma \big (E_l(x_{ti}^T) \big) \Big) \bigg] + \mathrm{MSE}(E(\tilde{x}_j), \tilde{z}_j) \nonumber
\end{align}
where MSE is the mean-squared-error, $E_l()$ represents the output of the $l$-th layer of the encoder, and $\lambda_\mathrm{s}$ is a weighting term, set to $\lambda_\mathrm{s}=0.1$. Since images for the task of urban scene segmentation usually contain multiple classes, which may also have different styles, we follow \citealp{CACE} in this case and use class-specific moments to calculate Eq. \ref{eq:adain} and Eq. \ref{eq:adain_loss}. These moments are extracted using a resized version of the source segmentation mask for the content feature map and pseudo-labels for the style feature map. The target moments are stored in style memory $Q$, allowing to transfer source images into previous styles. 

\subsection{Self-training}
Self-training first converts the $N_t$ softmax outputs $\{p_{ti}^T\}_{i=1}^{N_t}$ of the model for the current test images $\{x_{ti}^T\}_{i=1}^{N_t}$ at time step $t$ into pseudo-labels $\hat{y}_{ti}^T = \mathrm{argmax}(p_{ti}^T)$. These pseudo-labels are subsequently used to minimize the following cross-entropy loss
\begin{equation}
    \mathcal{L}_{\mathrm{CE}}^T(\hat{y}_{t}^T, p_{t}^T) = - \frac{1}{N_t} \sum_{i=1}^{N_t}\sum_{c=1}^C \hat{y}_{ti}^T \mathrm{log}(p_{tic}^T),
\end{equation}
where $C$ denotes the total number of classes. Clearly, most problems in reality are not as simple as depicted in Figure \ref{fig:gradual_motivation}, and there can already be erroneous predictions within the training domain. Since the amount of incorrect predictions can increase, especially when a domain change occurs, it is important to prevent the accumulation of the initial and subsequent errors. A factor that amplifies error accumulation when conducting self-training with pseudo-labels is that the cross-entropy loss has large gradient magnitudes for uncertain predictions \shortcite{mummadi2021test}. Since it is mostly the uncertain predictions that tend to be incorrect, their incorporation into the training process will prevent a successful adaption to the current target domain. This problem can be mitigated by using a threshold which filters out all pseudo-labels below a certain (softmax) confidence level. Although defining a fixed threshold can work well when adapting the model to a single target domain, it is insufficient for a test sequence containing multiple domains. In addition, different models and problems tend to have different confidences: Over-confident networks naturally have high confidences, while datasets with many classes tend to be less confident. Therefore, we introduce an adaptively smoothed threshold with momentum $\alpha_\text{th}=0.1$ that leverages the current softmax probabilities as follows:
\begin{equation}
    \gamma_t = (1-\alpha_\text{th})\, \gamma_{t-1} + \alpha_\text{th} \sqrt{\frac{1}{N_t} \sum_{i=1}^{N_t} \underset{c}{\mathrm{max}}(p_{tic}^T)}.
\end{equation}

\begin{figure*}
  \centering
  \setlength{\tabcolsep}{1pt}
  \begin{tabular}{ccc}
    clear & day2night & clear2fog \\
    \setlength{\tabcolsep}{0pt}
    \begin{tabular}{cc}
    \includegraphics[width=2.2cm]{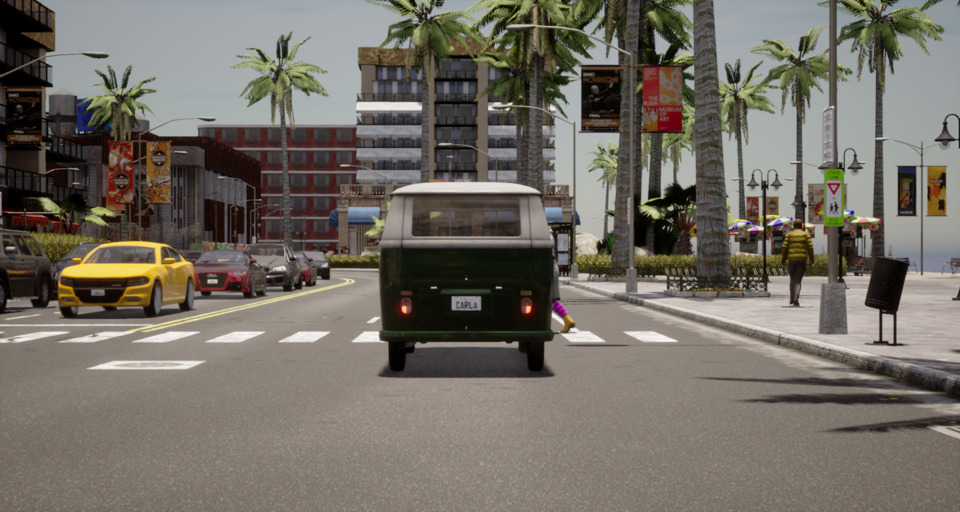} & \includegraphics[width=2.2cm]{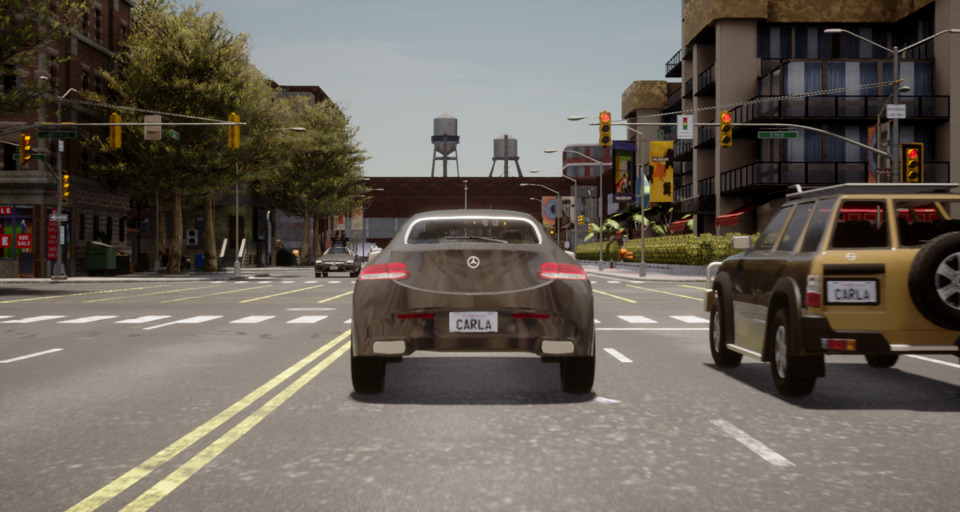} \\[-4pt]
    \includegraphics[width=2.2cm]{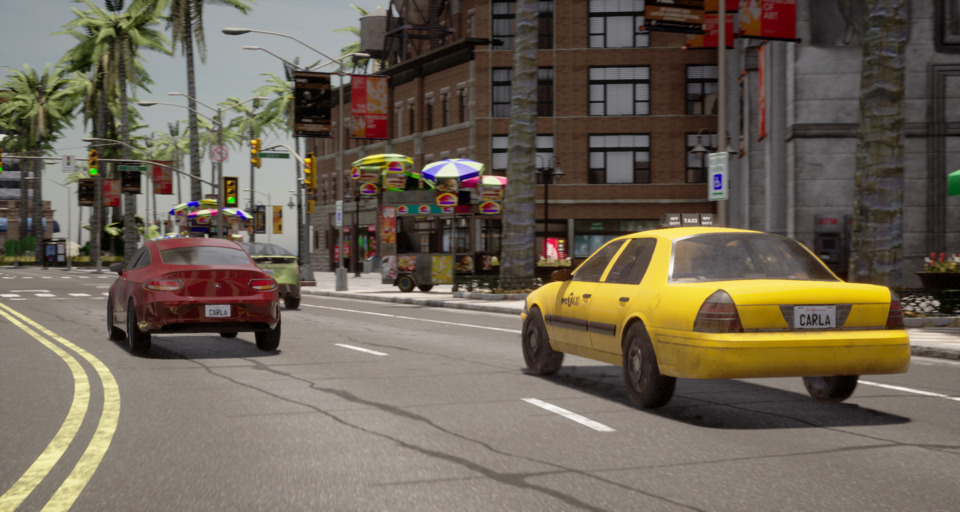} & \includegraphics[width=2.2cm]{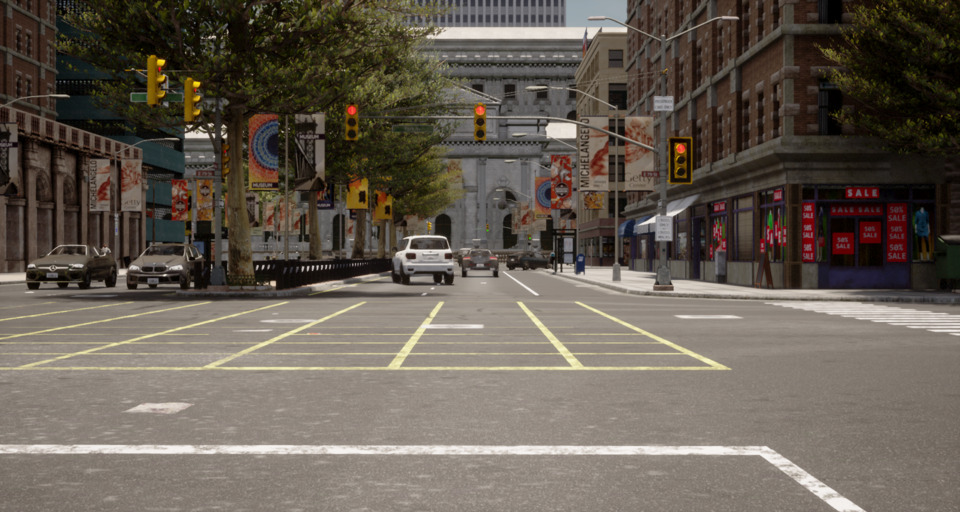} 
    \end{tabular}
    & 
    \setlength{\tabcolsep}{0pt}
    \begin{tabular}{cc}
    \includegraphics[width=2.2cm]{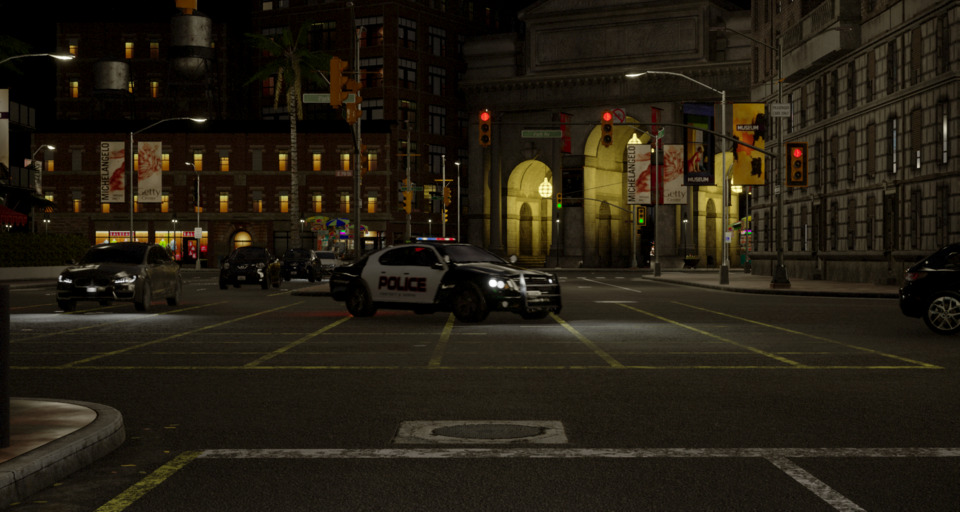} & \includegraphics[width=2.2cm]{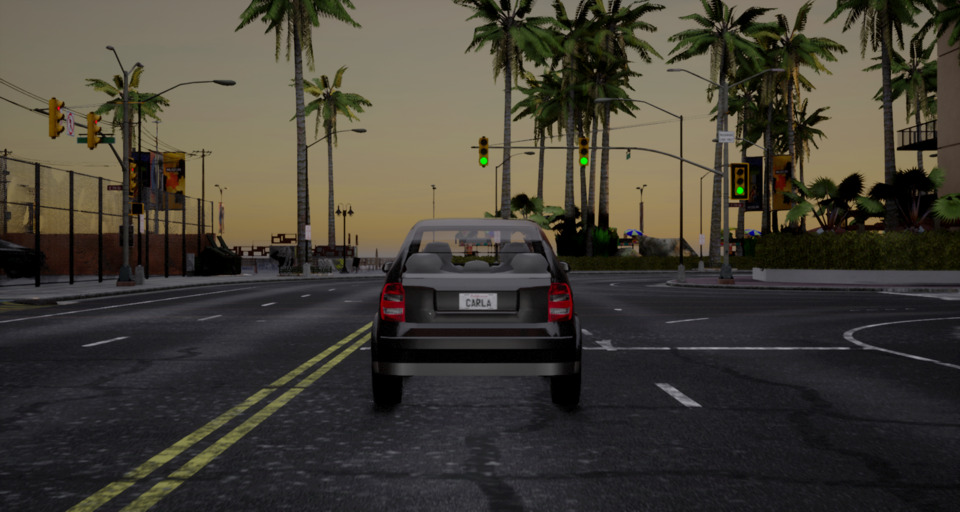} \\[-4pt]
    \includegraphics[width=2.2cm]{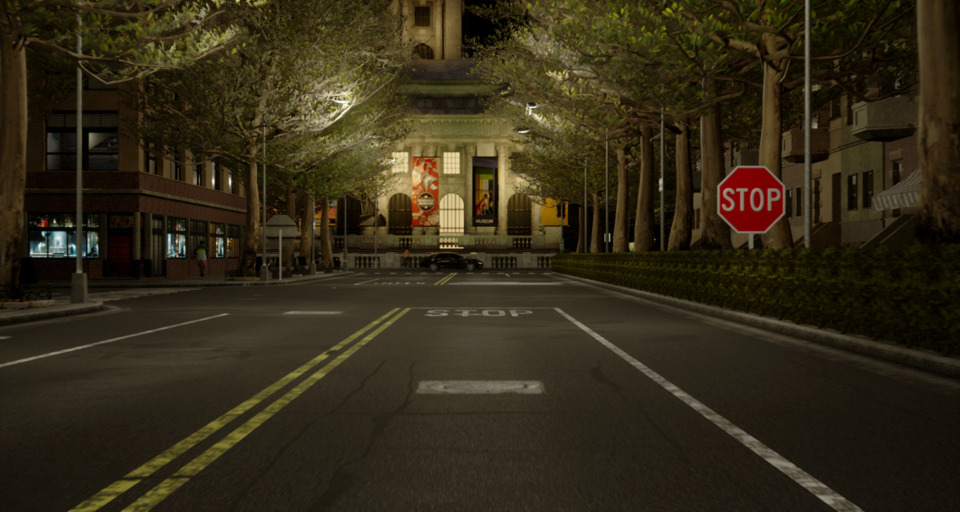} & \includegraphics[width=2.2cm]{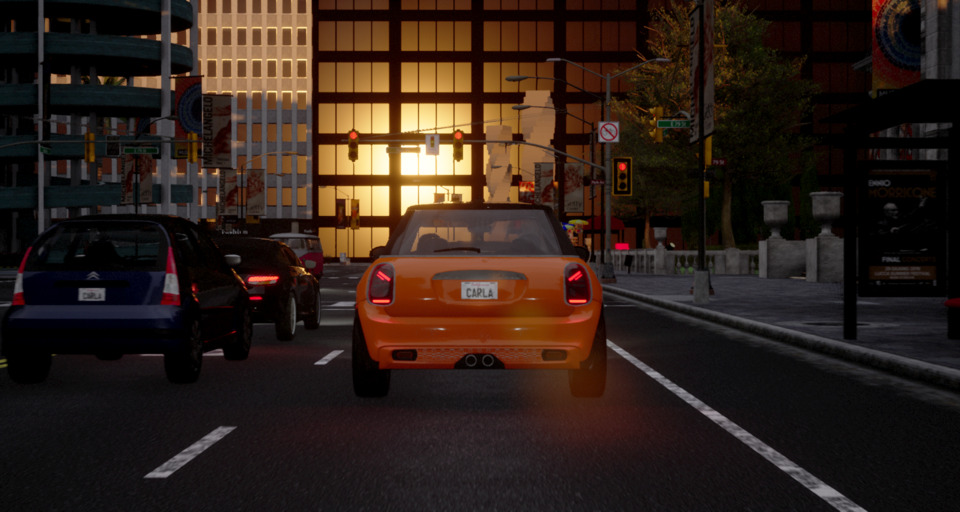} 
    \end{tabular}
    & 
    \setlength{\tabcolsep}{0pt}
    \begin{tabular}{cc}
    \includegraphics[width=2.2cm]{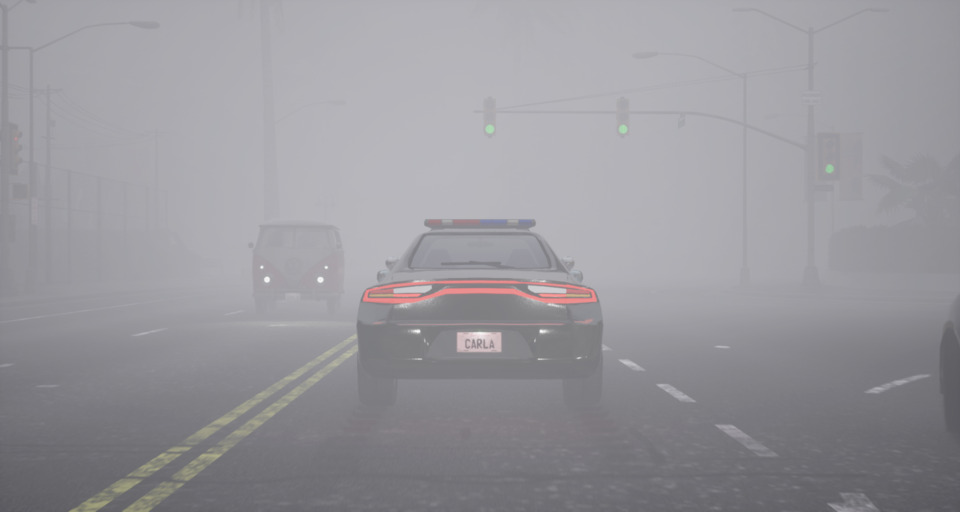} & \includegraphics[width=2.2cm]{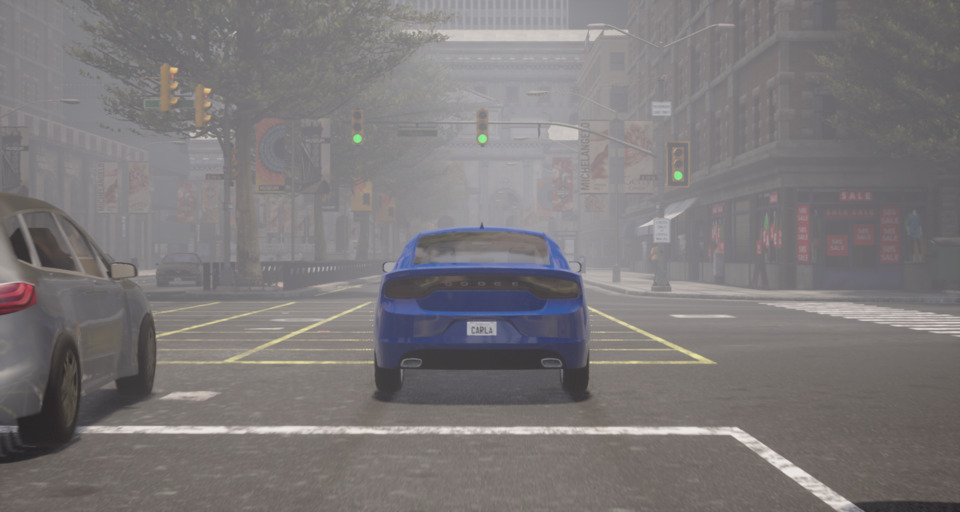} \\[-4pt]
    \includegraphics[width=2.2cm]{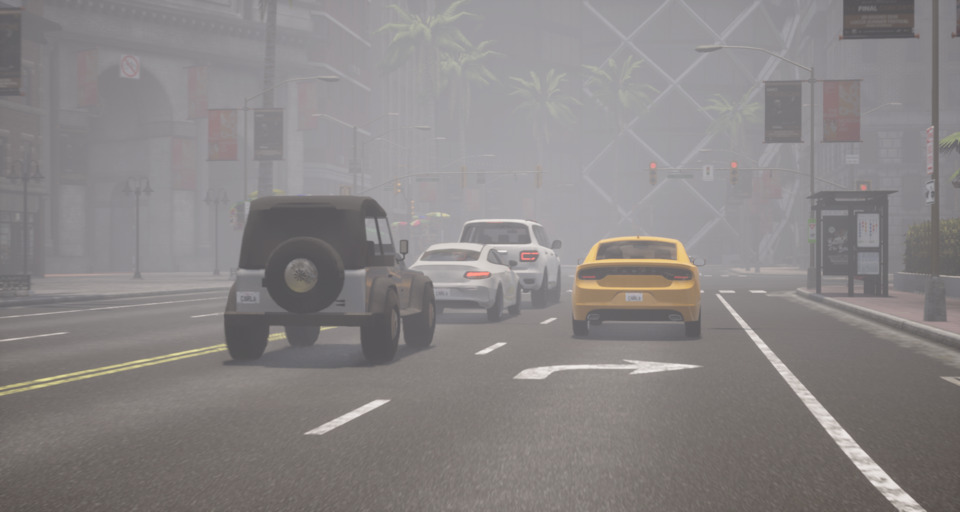} & \includegraphics[width=2.2cm]{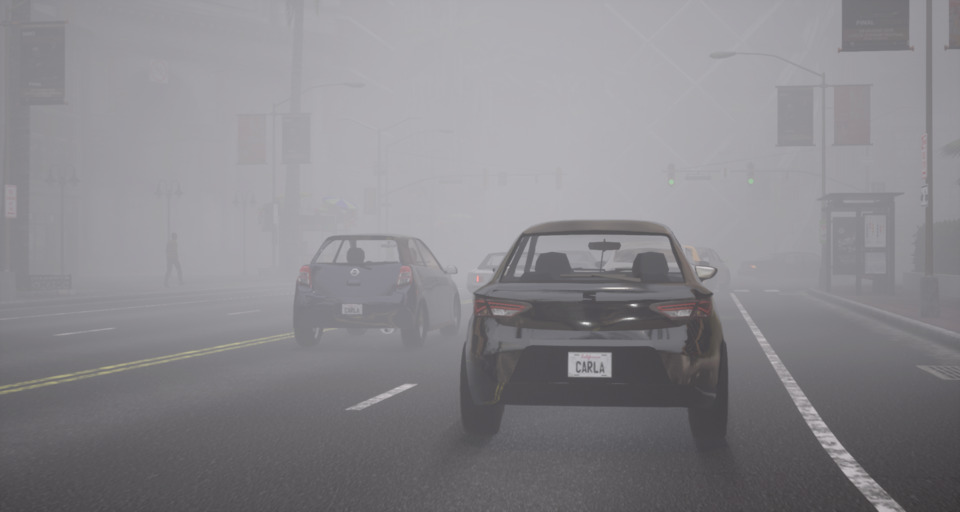} 
    \end{tabular}
    \\
    \setlength{\tabcolsep}{0pt}
    \begin{tabular}{cc}
    \includegraphics[width=2.2cm]{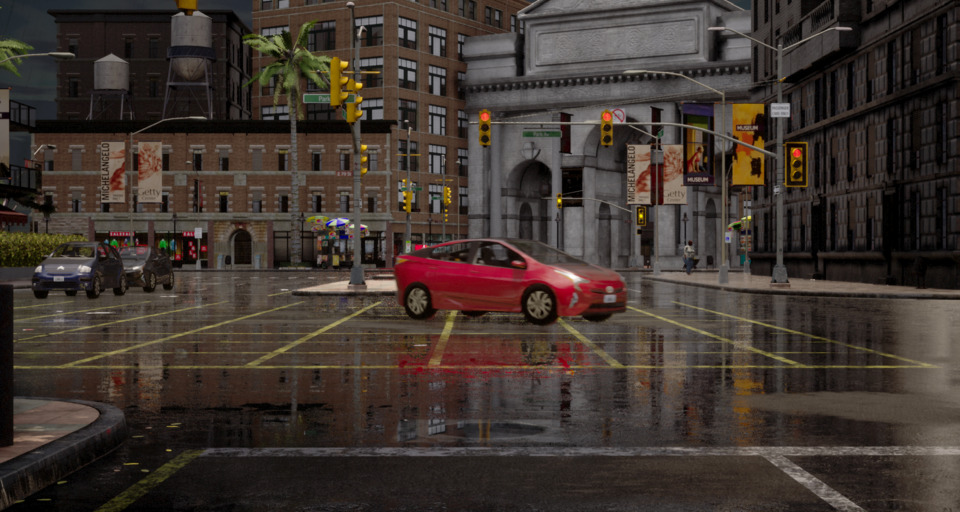} & \includegraphics[width=2.2cm]{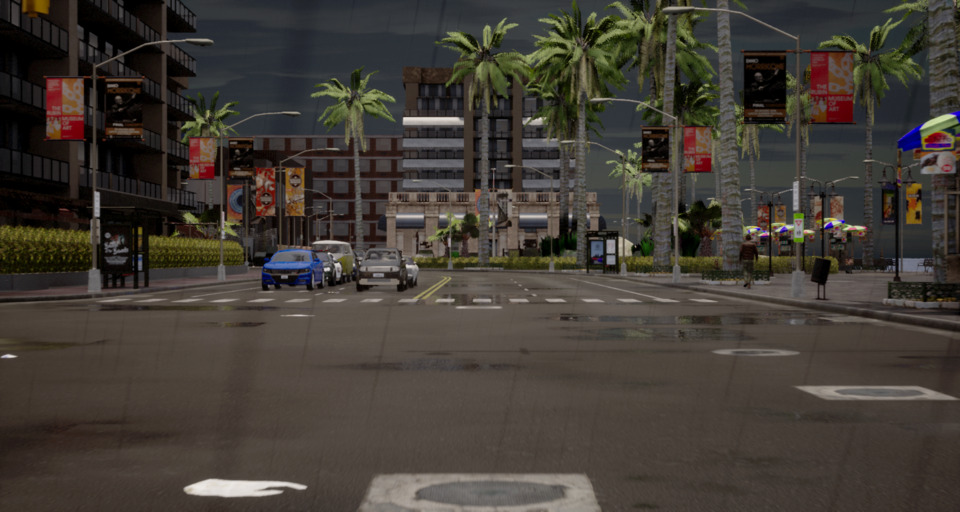} \\[-4pt]
    \includegraphics[width=2.2cm]{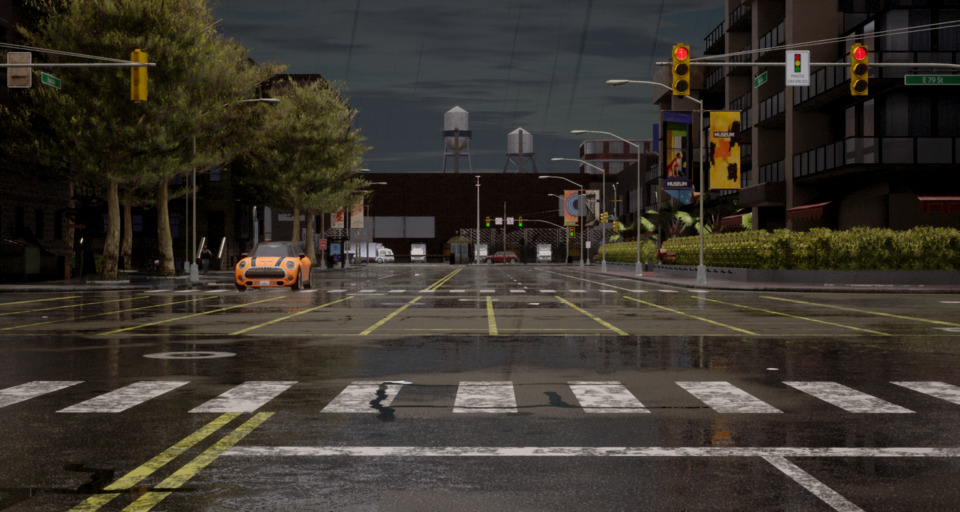} & \includegraphics[width=2.2cm]{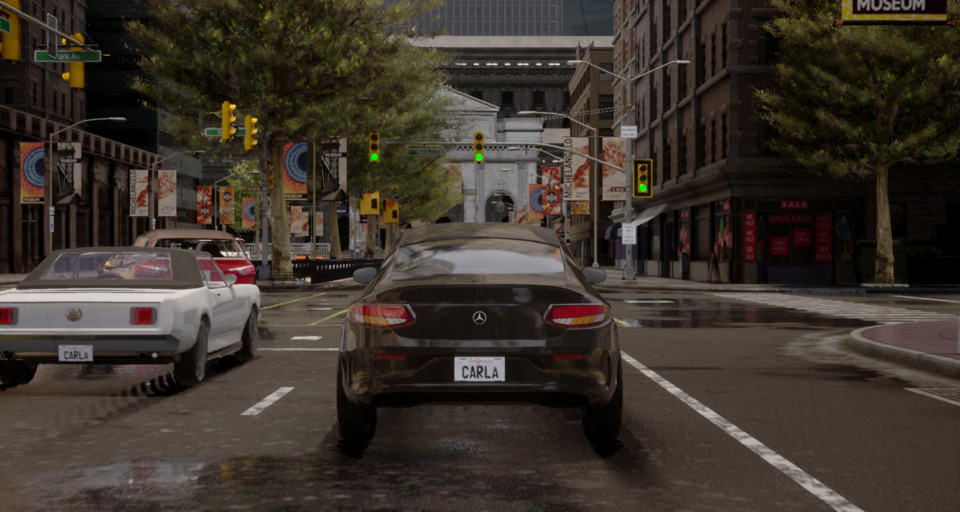} 
    \end{tabular}
    & 
    \setlength{\tabcolsep}{0pt}
    \begin{tabular}{cc}
    \includegraphics[width=2.2cm]{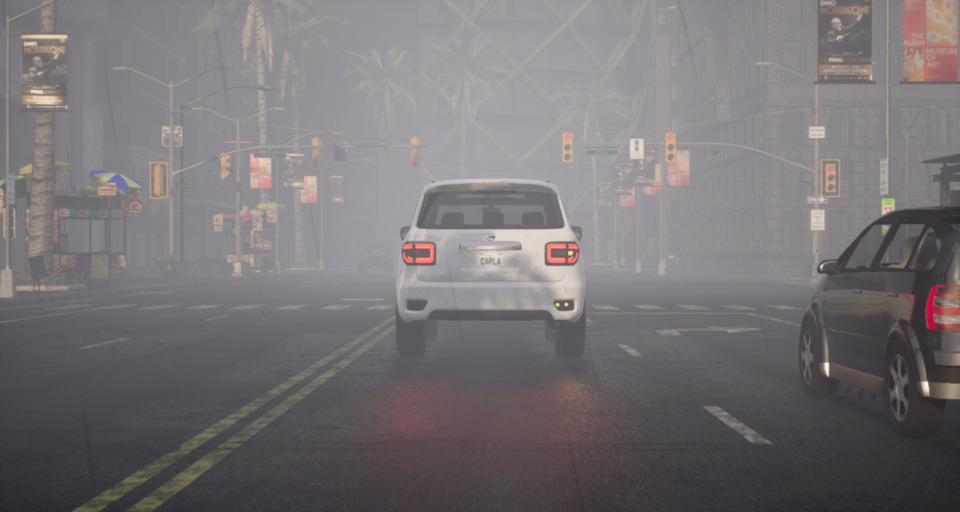} & \includegraphics[width=2.2cm]{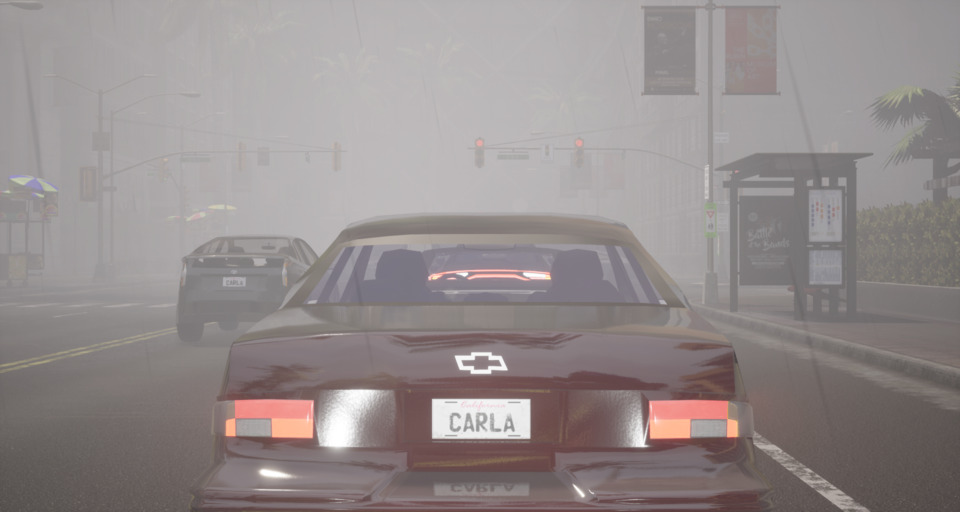} \\[-4pt]
    \includegraphics[width=2.2cm]{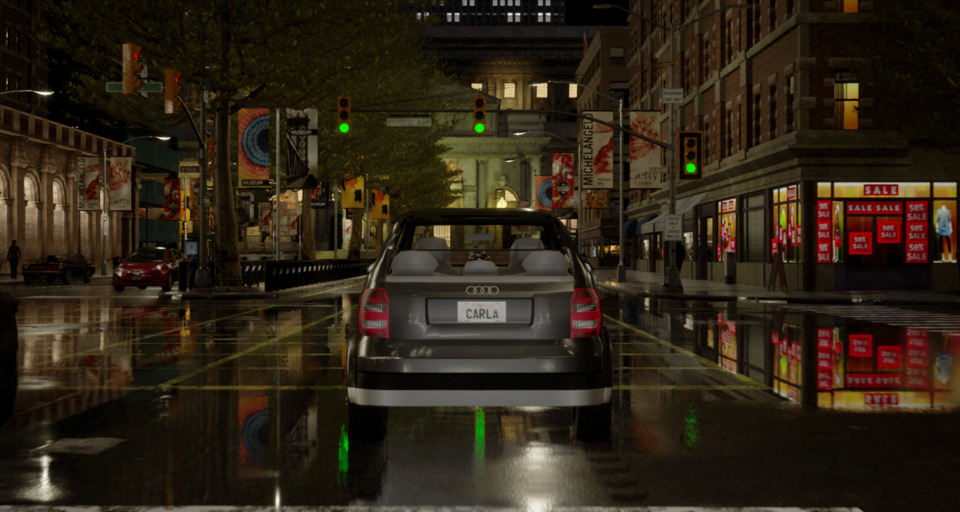} & \includegraphics[width=2.2cm]{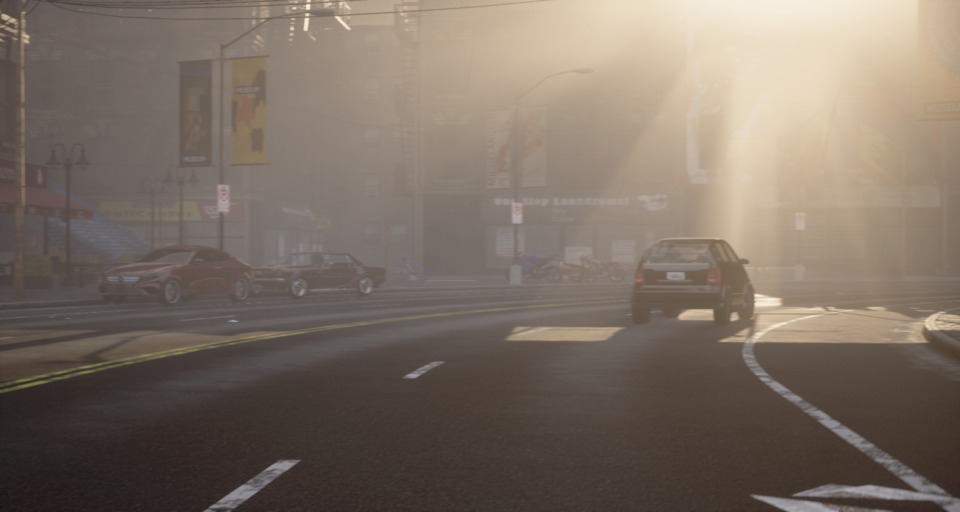} 
    \end{tabular}
    & 
    \setlength{\tabcolsep}{0pt}
    \begin{tabular}{cc}
    \includegraphics[width=2.2cm]{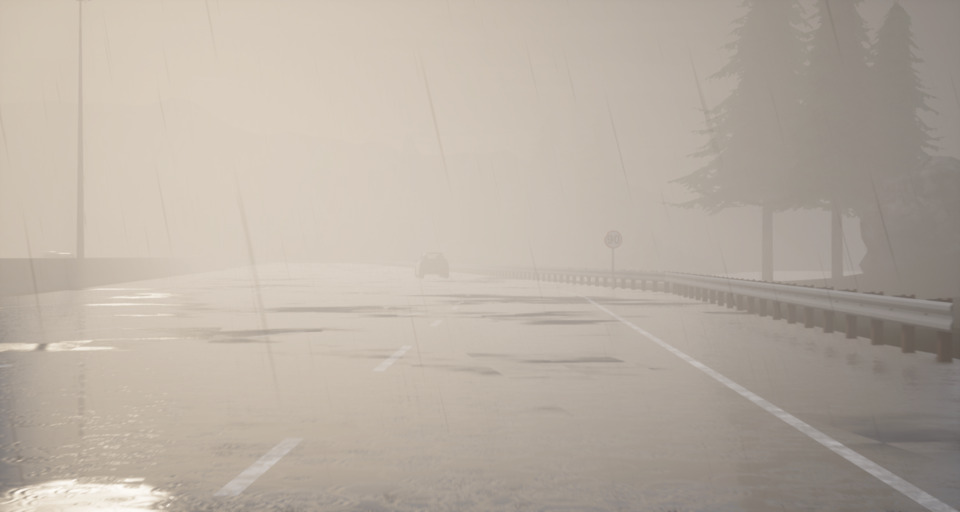} & \includegraphics[width=2.2cm]{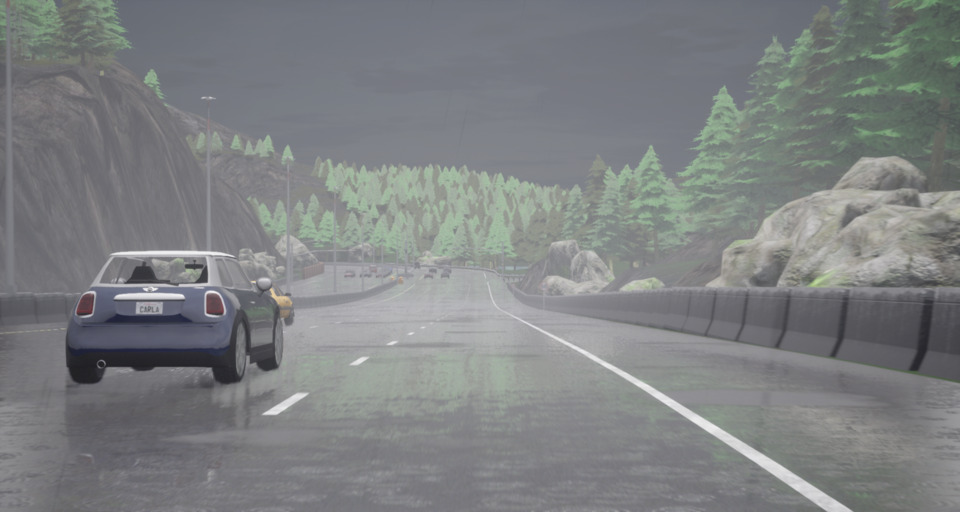} \\[-4pt]
    \includegraphics[width=2.2cm]{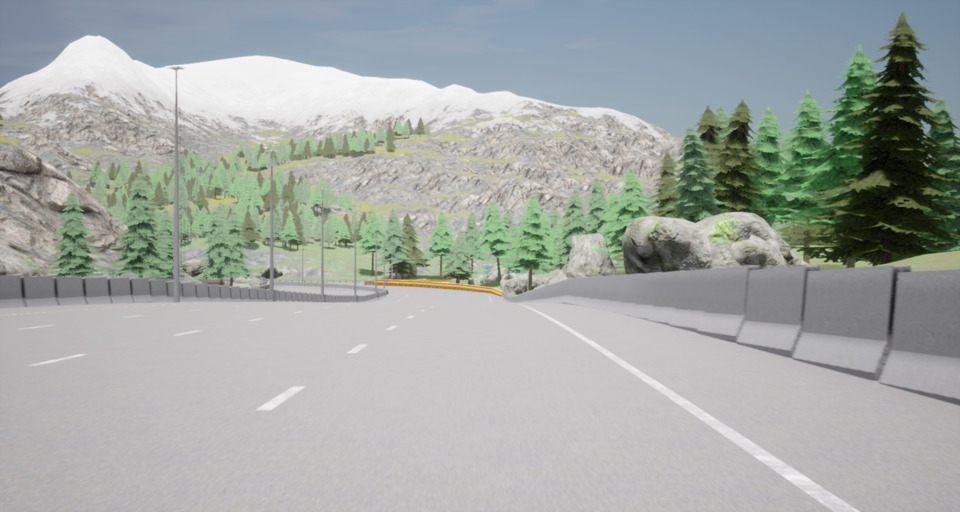} & \includegraphics[width=2.2cm]{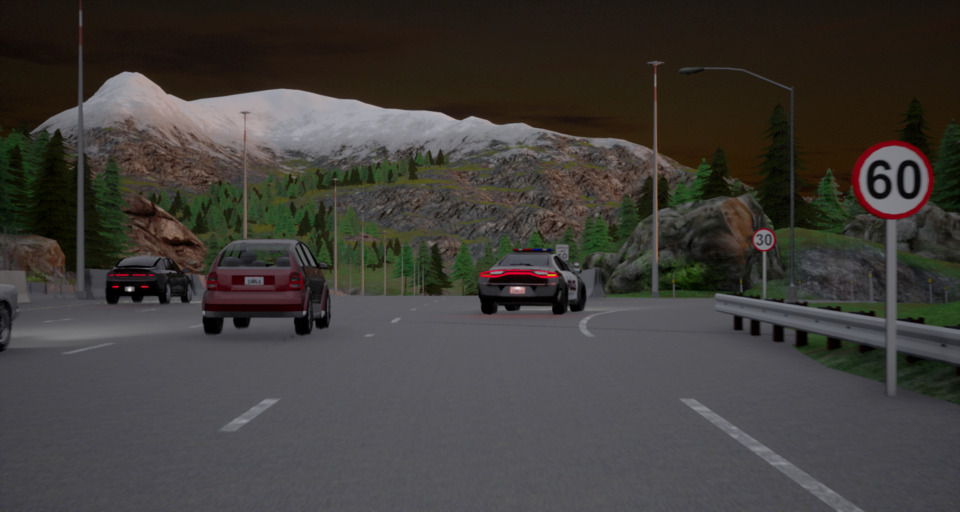} 
    \end{tabular}
    \\
    clear2rain & dynamic & highway
  \end{tabular}
  \caption{CarlaTTA: A synthetic driving dataset to explore gradual domain shifts in urban scenes.}
  \label{fig:carla_dataset}
\end{figure*}

\section{Dataset: Gradual Domain Changes for Urban Scene Segmentation}
Currently, there are not many datasets that are suited for investigating gradual test-time adaptation. Even though there already exist various real-world and synthetic driving datasets that contain different domains, such as Cityscapes \shortcite{cordts2016cityscapes}, ACDC \shortcite{sakaridis2021acdc}, Waymo \shortcite{sun2020scalability}, BDD100K \shortcite{yu2020bdd100k}, SYNTHIA \shortcite{ros2016synthia}, and GTA5 \shortcite{Richter_2016_ECCV}, they all involve only stationary domains and no sequences with gradual changes. To close this gap we introduce CarlaTTA: a dataset that enables the exploration of gradual test-time adaptation for urban scene segmentation. It is based on CARLA \shortcite{osovitskiy17carla}, an open-source simulator for autonomous driving research. We create five gradual test-sequences, all evolving from the stationary source domain \textit{clear} which is recorded at noon in clear weather. \textit{day2night} depicts one complete day-night cycle by varying the sun altitude and sun azimuth angle. Different weather changes are addressed by the sequences \textit{clear2fog} and \textit{clear2rain}, where \textit{clear2fog} changes cloudiness and fog density, while \textit{clear2rain} varies cloudiness, precipitation, puddles, and wetness. \textit{dynamic} combines the domain changes \textit{day2night}, \textit{clear2fog}, and \textit{clear2rain}. Not only does it result in overlapping domain shifts, but also introduces new shifts, such as, reflecting lights during a rainy night. To also investigate long-term behavior, \textit{dynamic} contains multiple day-night and weather cycles resulting in a five times longer sequence. \textit{highway} builds on top of the dynamic weather setting. In contrast to the previous datasets which mainly introduce covariate shifts, \textit{highway} also introduces label distribution shifts, since the vehicle drives from the city onto the highway. Example images are shown in Figure \ref{fig:carla_dataset}. Further visualizations and insights into our dataset, including a detailed illustration of the weather parameters, are presented in Appendix \ref{appendix:dataset}.


\section{Experiments}

\textbf{Baselines} Since BN has proven to be very effective during test-time \shortcite{schneider2020improving}, we consider several variations that can be derived from the following equations:
\begin{align}
\begin{split}
    \mu_{tm} = (1 - \alpha) \hat{\mu}_m^S + \alpha \mu_{tm}^T
    \\
    \sigma_{tm} = (1 - \alpha) \hat{\sigma}_m^S + \alpha \sigma_{tm}^T. 
    \label{eq:bn}
\end{split}
\end{align}
While $(\hat{\mu}_m^S, \hat{\sigma}_m^S)$ denote the running mean and standard deviation of channel $m$ estimated during source training, $(\mu_{tm}^T, \sigma_{tm}^T)$ are the corresponding moments extracted from the current test batch at time step $t$. By using Eq. \ref{eq:bn}, the notation of BN related baselines can be harmonized: $\alpha=0$ refers to the commonly known \textit{source} baseline (BN--0), $\alpha=1$ only exploits the current test statistics (BN--1), and $\alpha=0.1$ leverages the source statistics as a prior (BN--0.1). However, none of them exploits gradual domain shifts, since Eq. \ref{eq:bn} is instantaneous at time step $t$. Therefore, we introduce BN--EMA, which incorporates previous domain shifts by performing an exponential moving average using the running statistics from time step $(t-1)$:
\begin{align}
\begin{split}
    \hat{\mu}_{tm} = (1 - \alpha) \hat{\mu}_{(t-1)m} + \alpha \mu_{tm}^T 
    \\ 
    \hat{\sigma}_{tm} = (1 - \alpha) \hat{\sigma}_{(t-1)m} + \alpha \sigma_{tm}^T.
\end{split}
\end{align}
To further evaluate our method, we compare to several approaches from related fields: TENT \shortcite{TENT} uses BN--1 in combination with an entropy minimization strategy with respect to the BN parameters. CoTTA \shortcite{wang2022continual} utilizes BN--1 and a mean teacher with test-time augmentation to perform entropy minimization. Further it introduces stochastic restore, where source pre-trained weights are restored with a certain probability. AdaContrast \shortcite{chen2022contrastive} uses pseudo-label refinement for self-training and contrastive learning.

For segmentation, we additionally consider MEMO \shortcite{MEMO}, which combines test-time augmentation and entropy minimization and two methods from one-shot UDA. While ASM \shortcite{ASM} uses an AdaIN based style transfer model to explore the style space, SM-PPM \shortcite{SM-PPM} integrates style mixing into the segmentation network and combines it with patch-wise prototypical matching.

\subsection{Adapting to shifts caused by corruptions}

\begin{table*}[!h]
\renewcommand{\arraystretch}{1.2}
\centering
\caption{Classification error rate~(\%) on the corruption benchmark for the online continual test-time adaptation task on the highest corruption severity level 5. We report the performance of our method averaged over 5 runs.} \label{tab:continual-corruptions}
\scalebox{0.93}{
\tabcolsep3pt
\begin{tabular}{l|l|c|c|  ccccccccccccccc|c}\hline
& \multicolumn{3}{l|}{Time} & \multicolumn{15}{l|}{$t\xrightarrow{\hspace*{12.2cm}}$}& \\ \hline
& Method & \rotatebox[origin=c]{90}{ source-free } & \rotatebox[origin=c]{90}{updates} & \rotatebox[origin=c]{70}{Gaussian} & \rotatebox[origin=c]{70}{shot} & \rotatebox[origin=c]{70}{impulse} & \rotatebox[origin=c]{70}{defocus} & \rotatebox[origin=c]{70}{glass} & \rotatebox[origin=c]{70}{motion} & \rotatebox[origin=c]{70}{zoom} & \rotatebox[origin=c]{70}{snow} & \rotatebox[origin=c]{70}{frost} & \rotatebox[origin=c]{70}{fog}  & \rotatebox[origin=c]{70}{brightness} & \rotatebox[origin=c]{70}{contrast} & \rotatebox[origin=c]{70}{elastic\_trans} & \rotatebox[origin=c]{70}{pixelate} & \rotatebox[origin=c]{70}{jpeg} & Mean \\
\hline
\multirow{7}{*}{\rotatebox[origin=c]{90}{CIFAR10C}}& BN--0 (src.) & \ding{51} & - & 72.3 & 65.7 & 72.9 & 46.9 & 54.3 & 34.8 & 42.0 & 25.1 & 41.3 & 26.0 & 9.3 & 46.7 & 26.6 & 58.5 & 30.3 & 43.5 \\
& BN--1     & \ding{51} & - & 28.1 & 26.1 & 36.3 & 12.8 & 35.3 & 14.2 & 12.1 & 17.3 & 17.4 & 15.3 & 8.4 & 12.6 & 23.8 & 19.7 & 27.3 & 20.4 \\
& TENT-cont.& \ding{51} & 1 & 24.8 & 20.6 & 28.6 &	14.4 & 31.1 & 16.5 & 14.1 & 19.1 & 18.6 & 18.6 & 12.2 & 20.3 & 25.7 & 20.8 & 24.9 & 20.7 \\
& AdaContrast & \ding{51} & 1 & 29.1 & 22.5 & 30.0 & 14.0 & 32.7 & 14.1 & 12.0 & 16.6 & 14.9 & 14.4 & 8.1 & 10.0 & 21.9 & 17.7 & 20.0 & 18.5\\
& CoTTA     & \ding{51} & 1 & 24.3 & 21.3 & 26.6 & 11.6 & 27.6 & 12.2 & 10.3 & 14.8 & 14.1 & 12.4 & 7.5 & 10.6 & \textbf{18.3} & \textbf{13.4} & \textbf{17.3} & 16.2 \\
& GTTA-MIX & \ding{55} & 1 & 26.0 & 21.5 & 29.7 & 11.1 & 30.0 & 12.2 & 10.5 & 15.1 & 14.1 & 12.3 & 7.5 & 10.0 & 20.4 & 15.8 & 21.4 & 17.2$\pm$0.06\\
& GTTA-MIX & \ding{55} & 4 & \textbf{23.4} & \textbf{18.3} & \textbf{25.5} & \textbf{10.1} & \textbf{27.3} & \textbf{11.6} & \textbf{10.1} & \textbf{14.1} & \textbf{13.0} & \textbf{10.9} & \textbf{7.4} & \textbf{9.0} & 19.4 & 14.5 & 19.8 & \textbf{15.6}$\pm$0.04\\
\hline
\multirow{7}{*}{\rotatebox[origin=c]{90}{CIFAR100C}}& BN--0 (src.) & \ding{51} & - & 73.0&	68.0&	39.4&	29.3&	54.1&	30.8&	28.8&	39.5&	45.8&	50.3&	29.5&	55.1&	37.2	&74.7&	41.2&	46.4\\
& BN--1     & \ding{51} & - & 42.1 & 40.7 & 42.7 & 27.6 & 41.9 & 29.7 & 27.9 & 34.9 & 35.0 & 41.5 & 26.5 & 30.3 & 35.7 & 32.9 & 41.2 & 35.4 \\
& TENT-cont.& \ding{51} & 1 & 37.2 & 35.8 & 41.7 & 37.9 & 51.2 & 48.3 & 48.5 & 58.4 & 63.7 & 71.1 & 70.4 & 82.3 & 88.0 & 88.5 & 90.4 & 60.9 \\
& AdaContrast & \ding{51} & 1 & 42.3 & 36.8 & 38.6 & 27.7 & 40.1 & 29.1 & 27.5 & 32.9 & 30.7 & 38.2 & 25.9 & 28.3 & 33.9 & 33.3 & 36.2 & 33.4 \\
& CoTTA     & \ding{51} & 1 & 40.1 & 37.7 & 39.7 & 26.9 & 38.0 & 27.9 & 26.4 & 32.8 & 31.8 & 40.3 & 24.7 & 26.9 & 32.5 & 28.3 & 33.5 & 32.5 \\
& GTTA-MIX & \ding{55} & 1 & 39.4 & 34.4 & 36.6 & 24.7 & 36.8 & 26.6 & 24.3 & 30.1 & 28.9 & 34.6 & 22.8 & 25.1 & 30.7 & 26.9
& 34.7 & 30.4$\pm$0.01\\
& GTTA-MIX & \ding{55} & 4 & \textbf{36.4} & \textbf{32.1} & \textbf{34.0} & \textbf{24.4} & \textbf{35.2} & \textbf{25.9} & \textbf{23.9} & \textbf{28.9} & \textbf{27.5} & \textbf{30.9} & \textbf{22.6} & \textbf{23.4} & \textbf{29.4} & \textbf{25.5}
& \textbf{33.3} & \textbf{28.9}$\pm$0.02 \\
\hline
\multirow{9}{*}{\rotatebox[origin=c]{90}{ImageNetC}} & BN--0 (src.) & \ding{51} & - & 97.8 & 97.1 & 98.2 & 81.7 & 89.8 & 85.2 & 78.0 & 83.5 & 77.1 & 75.9 & 41.3 & 94.5 & 82.5 & 79.3 & 68.6 & 82.0 \\
 & BN--1     & \ding{51} & - & 85.0 & 83.7 & 85.0 & 84.7 & 84.3 & 73.7 & 61.2 & 66.0 & 68.2 & 52.1 & 34.9 & 82.7 & 55.9 & 51.3 & 59.8 & 68.6 \\
 & TENT-cont.& \ding{51} & 1 & 81.6 & 74.6 & 72.7 & 77.6 & 73.8 & 65.5 & 55.3 & 61.6 & 63.0 & 51.7 & 38.2 & 72.1 & 50.8 & 47.4 & 53.3 & 62.6 \\
 & AdaContrast & \ding{51} & 1 & 82.9 & 80.9 & 78.4 & 81.4 & 78.7 & 72.9 & 64.0 & 63.5 & 64.5 & 53.5 & 38.4 & 66.7 & 54.6 & 49.4 & 53.0 & 65.5 \\
 & CoTTA     & \ding{51} & 1 & 84.7 & 82.1 & 80.6 & 81.3 & 79.0 & 68.6 & 57.5 & 60.3 & 60.5 & 48.3 & 36.6 & 66.1 & \textbf{47.2} & \textbf{41.2} & \textbf{46.0} & 62.7 \\
 & GTTA-MIX & \ding{55} & 1 & 80.5 & 74.7 & 72.4 & 77.8 & 75.7 & 64.3 & 54.0 & 57.0 & 58.6 & 44.6 & 33.9 & 67.5 & 49.4 & 44.7 & 49.3 & 60.3$\pm$0.07 \\
 & GTTA-MIX & \ding{55} & 4 & \textbf{75.2} & \textbf{67.4} & \textbf{64.6} & 73.3 & 72.5 & 61.8 & 52.7 & \textbf{53.0} & \textbf{54.9} & \textbf{42.6} & 33.8 & 63.9 & 48.9 & 44.4 & 47.0 & 57.1$\pm$0.14 \\
 & GTTA-ST & \ding{55} & 1 & 80.6 & 74.1 & 74.3 & 76.8 & 74.9 & 62.3 & 53.9 & 56.4 & 58.0 & 44.1 & \textbf{33.4} & 62.2 & 48.6 & 44.9 & 50.4 & 59.7$\pm$0.08 \\
 & GTTA-ST & \ding{55} & 4 & 76.7 & 69.0 & 69.3 & \textbf{73.1} & \textbf{72.1} & \textbf{59.6} & \textbf{51.6} & 53.4 & 56.7 & 42.9 & 33.7 & \textbf{57.2} & 47.9 & 43.4 & 47.9 & \textbf{57.0}$\pm$0.06 \\
\hline
\end{tabular}}
\end{table*}
\textbf{Corruption benchmarks}
\label{sec:corruption_benchmark}
CIFAR10C, CIFAR100C, and ImageNet-C were originally published to evaluate robustness of neural networks \shortcite{hendrycks2019benchmarking}. The benchmark comprises of 15 corruptions with 5 severity levels, which were applied to the validation images of ImageNet \shortcite{imagenet_cvpr09} and the test images of CIFAR10 and CIFAR100 \shortcite{krizhevsky2009learning}, respectively. In accordance with the RobustBench benchmark \shortcite{croce2020robustbench}, a pre-trained WideResNet-28 is used for CIFAR10-to-CIFAR10C, ResNeXt-29 for CIFAR100-to-CIFAR100C, and ResNet-50 for ImageNet-to-ImageNet-C. Following the implementation and hyperparameters of \citealp{wang2022continual}, a batch size of 200 is utilized for CIFAR and a batch size of 64 for ImageNet. Note that we also investigate single sample test-time adaptation in Appendix \ref{appendix:single_sample}. We use Adam \shortcite{kingma2014adam} as an optimizer with a fixed learning rate of 1e-5 for all experiments. Due to the low-resolution images of CIFAR, we only consider the mixup variant GTTA-MIX for CIFAR10C and CIFAR100C. A mixup strength of $\lambda_\text{mix}=\frac{1}{3}$ is used for all experiments. For ImageNet-C, we additionally compare to the style transfer variant GTTA-ST. The style transfer network consists of the same VGG19 based encoder-decoder architecture as used in \cite{CACE} and is pre-trained for 20k iterations on the source domain using Adam with learning rate $1\times10^{-4}$.

\begin{table*}[!h]
\renewcommand{\arraystretch}{1.1}
\centering
\caption{Average classification error rate~(\%) for the gradual corruption benchmark across all 15 corruptions. We separately report the performance averaged over all severity levels (level 1--5) and averaged only over the highest severity level 5 (level 5). The number in brackets denotes the difference compared to the continual benchmark.}
\label{tab:gradual-corruptions}
\scalebox{0.94}{
\tabcolsep3pt
\begin{tabular}{l|l|ccccccccc}\hline
 & & BN--0 & BN--1 & TENT-cont. & AdaCont. & CoTTA & GTTA-MIX & GTTA-MIX & GTTA-ST & GTTA-ST\\ \hline
 & src.-free & \ding{51} & \ding{51} & \ding{51} & \ding{51} & \ding{51} & \ding{55} & \ding{55} & \ding{55} & \ding{55} \\
 & updates & - & - & 1 & 1 & 1 & 1 & 4 & 1 & 4 \\ \hline
\multirow{2}{*}{CIFAR10C} & level 1--5 & 24.7 & 13.7 & 20.4 & 12.1 & 10.9 & 10.5 & \textbf{9.5} & - & - \\
& level 5 & 43.5 & 20.4 & 25.1 \small{(+4.4)} & 15.8 \small{(-2.7)} & 14.2 \small{(-2.0)} & 15.0 \small{(-2.2)} & \textbf{13.0} \small{(-2.6)} & - & - \\ \hline
\multirow{2}{*}{CIFAR100C} & level 1--5 & 33.6 & 29.9 & 74.8 & 33.0 & 26.3 & 24.3 & \textbf{23.9} & - & - \\
& level 5 & 46.4 & 35.4 & 75.9 \small{(+15.0)} & 35.9 \small{(+2.5)} & 28.3 \small{(-4.2)} & 27.6 \small{(-2.8)} & \textbf{26.1} \small{(-2.8)} & - & - \\ \hline
\multirow{2}{*}{ImageNetC} & level 1--5 & 58.4 & 48.3 & 46.4 & 66.3 & 38.8 & 39.3 & \textbf{37.7} & 39.8 & 38.7 \\
& level 5 & 82.0 & 68.6 & 58.9 \small{(-3.7)} & 72.6 \small{(+7.1)} & \textbf{43.1} \small{(-19.6)} & 51.8 \small{(-8.5)} & 47.7 \small{(-9.4)} & 51.9 \small{(-7.8)} & 48.3 \small{(-8.7)} \\ \hline
\end{tabular}}
\end{table*}

\textbf{Continual corruption benchmarks}
We first consider the continual TTA setting, as proposed in \citealp{wang2022continual}. Starting with a network pre-trained on source data, the model is adapted during test-time in an online fashion. Unlike the standard setting where the model is reset before being adapted to a new corruption type, the continual setting does not assume to have any knowledge about the current domain or shift. Test-time adaptation is performed under the highest corruption severity level 5.

The results are reported in Table \ref{tab:continual-corruptions}. Simply evaluating the pre-trained source model yields an average error of 43.5\% for CIFAR10C, 46.4\% for CIFAR100C, and 82.0\% for ImageNet-C. Using the current test batch to adapt the batch statistics (BN--1) already drastically decreases the error for all datasets. As already pointed out by \citealp{wang2022continual}, TENT-continual outperforms BN--1 in early stages, but quickly deteriorates after a few corruptions. This becomes particularly evident for CIFAR100C, where TENT achieves an error of 90.4\% for the last corruption. To avoid error accumulation, one can use TENT-episodic instead. However, in the episodic setup, knowledge from previous examples cannot be leveraged, resulting in a performance on par with BN--1. Another option to stabilize the training which we investigate in Appendix \ref{appendix:component_analysis} is source replay. This has the restriction of requiring access to source data, but stabilizes self-training and improves the performance, e.g., for TENT on all datasets. CoTTA shows its strong suits for CIFAR10C outperforming BN--1 by 4.2\% and performs comparably to TENT on ImageNet-C. AdaContrast outperforms BN--1, but lacks behind CoTTA on all datasets. Our method GTTA successfully shows on all datasets that generating intermediate domains by mixup or style transfer allows a better adaptation via self-training. Performing a single update per test batch leads to state-of-the-art results on CIFAR100C and ImageNet-C. Performing four updates results in a further improvement on all datasets and also sets state-of-the-art results on CIFAR10C. Note that utilizing more update steps for source-free methods like CoTTA does not improve the performance, on the contrary. A more in-depth analysis about when multiple update steps are beneficial is discussed in Appendix \ref{appendix:component_analysis}.

\textbf{Gradual corruption benchmarks}
We now investigate a setting, where the test domain changes gradually. Starting from the lowest severity level 1, the severity level is incremented as follows: $1 \rightarrow 2 \rightarrow \dots \rightarrow 5 \rightarrow \dots \rightarrow 2 \rightarrow 1$. After a cycle, we switch to the next corruption type and sequentially repeat the same procedure for all corruptions.

The results for the gradual setting are reported in Table \ref{tab:gradual-corruptions}. For a direct comparison between the results of the continual and gradual setting, we report the average error at the highest severity level 5 in addition to the average over all severities. The effect that TENT-continual's performance deteriorates over time is even more prominent in the gradual setup due to longer sequences. TENT and AdaContrast show both for some of the datasets a worse performance at level 5 compared to the continual setup. CoTTA and GTTA both benefit from the gradually changing domains. GTTA-MIX shows the best performance on all datasets, except for ImageNet-C at level 5 where CoTTA takes the lead. GTTA-ST performs slightly worse than GTTA-MIX.

\subsection{Adapting to real-world distribution shifts}
\textbf{ImageNet-R}
To investigate the performance in the presence of distribution shifts not caused by corruptions, we also analyze ImageNet-R \shortcite{hendrycks2021many} using the same setting as for ImageNet-C. ImageNet-R consists of 30,000 samples depicting several renditions of 200 ImageNet classes. The results are shown in Table \ref{tab:imagenet-r}. While GTTA-MIX again outperforms previous methods on this benchmark, GTTA-ST shows a tremendous improvement.
\begin{table}[t]
\renewcommand{\arraystretch}{1.2}
\centering
\caption{Classification error rate~(\%) for ImageNet-R averaged over 5 runs using a ResNet-50.} \label{tab:imagenet-r}
\scalebox{0.97}{
\tabcolsep3pt
\begin{tabular}{l|c|c|c}\hline
Method & source-free & updates & error rate \\
\hline
BN--0 (source) & \ding{51} & - & 63.8 \\
BN--1       & \ding{51} & - & 60.4 \\
TENT-cont.  & \ding{51} & 1 & 57.6 \\
AdaContrast & \ding{51} & 1 & 59.1 \\
CoTTA       & \ding{51} & 1 & 57.4 \\
GTTA--MIX & \ding{55} & 1 & 56.4$\pm$0.23 \\
GTTA--MIX & \ding{55} & 4 & 56.6$\pm$0.76 \\
GTTA--ST & \ding{55} & 1 & 53.8$\pm$0.19 \\
GTTA--ST & \ding{55} & 4 & \textbf{52.5}$\pm$0.24 \\
\hline
\end{tabular}}
\end{table}

\textbf{Comparing GTTA-MIX and GTTA-ST}
We find that mixup is especially suited for compensating domain gaps covered by the corruption benchmark and not necessarily for real-world distribution shifts where style transfer demonstrates its advantages. Since mixup in our case is a linear combination of source and test images, it is intuitive, that GTTA-MIX particularly performs well on corruptions that are additive. Examples are, Gaussian noise, snow, frost, and fog. When it comes to natural distribution shifts, such as introduced by ImageNet-R, mixup has its limitations. In contrast, style transfer based on adaptive instance normalization can perform arbitrary style transfer, as shown by the original work \shortcite{AdaIN}. Even though GTTA-ST can cope with various domain shifts, as established by the results on ImageNet-C and ImageNet-R, it has a slight memory and computational overhead due to the additional style transfer network. 

\subsection{Experiments on CarlaTTA}
\textbf{Setup} To demonstrate the effectiveness of our approach for natural shifts in the context of autonomous driving, we consider the CarlaTTA benchmark below. We report the mean intersection-over-union (mIoU) over the entire test sequence. All methods use the same pre-trained source model trained for 100k iterations on the stationary source domain \textit{clear}. To prevent overfitting to the source domain, we apply random horizontal flipping, Gaussian blur, color jittering, as well as random scaling in the range [0.75, 2] before the image is cropped to a size of $1024 \times 512$.  Following the standard framework in UDA for semantic segmentation \shortcite{AdaptSegNet}, we use the DeepLab-V2 \shortcite{Deeplabv2} architecture with a ResNet-101 backbone. The style transfer network consists of the same architecture as described in Section \ref{sec:corruption_benchmark}, with the only difference that now class-conditional AdaIN layers are used.

\textbf{Implementation details} The segmentation model is trained with SGD using a constant learning of $2.5\times10^{-4}$, momentum of $0.9$, and weight decay of $5\times10^{-4}$. During test-time adaptation, we use batches consisting of two source samples and two crops of the current test sample. While one of the source samples is transferred into the current test style, the other is transferred into a previously seen style, as domain shifts may reoccur. During the online adaptation, both networks are updated once for each new test sample.

\subsubsection{Results for CarlaTTA}
\begin{table}[t] 
\renewcommand{\arraystretch}{1.1}
\centering
\caption{mIoU (\%) for CarlaTTA. The source model achieves 78.4\% mIoU on the \textit{clear} test split.}
\scalebox{0.92}{\tabcolsep5pt
\label{tab:carla-main}
\begin{tabular}{l|c|ccccc}
    \hline
  Method & \rotatebox[origin=c]{90}{src.-free} & \rotatebox[origin=c]{70}{day2night}  & \rotatebox[origin=c]{70}{clear2fog}  & \rotatebox[origin=c]{70}{clear2rain} & \rotatebox[origin=c]{70}{dynamic} & \rotatebox[origin=c]{70}{highway}  \\
    \hline
    BN--0 (source) & \ding{51} & 58.4 & 52.8 & 71.8 & 46.6 & 28.7 \\  
    BN--0.1 & \ding{51} & 62.7 & 56.5 & 72.8 & 52.1 & 37.2 \\  
    BN--1 & \ding{51} & 62.0 & 56.8 & 71.4 & 52.6 & 32.8 \\  
    BN--EMA & \ding{51} & 63.4 & 58.3 & 73.4 & 53.9 & 31.9 \\  
    \hline
    MEMO & \ding{51} & 61.0 & 55.1 & 71.6 & 50.3 & 35.2 \\
    TENT-cont. & \ding{51} & 61.5 & 56.0 & 70.9 & 50.3 & 32.0 \\
    TENT-ep. & \ding{51} & 61.9 & 56.8 & 71.4 & 52.6 & 32.8 \\
    CoTTA & \ding{51} & 61.4 & 56.8 & 70.7 & 46.4 & 33.8 \\  
    \hline
    ASM & \ding{55} & 58.5 & 53.0 & 69.2 & 50.2 & 39.4 \\ 
    SM-PPM & \ding{55} & 63.1 & 56.7 & 72.7 & 53.2 & 33.4 \\  
    \hline
    self-training & \ding{55} & 63.2 &	54.1 & 74.4 & 50.3 & 33.2 \\
    style-transfer & \ding{55} & 66.0 & \textbf{62.2} & 74.6 & 59.1 & 41.9 \\
    GTTA-ST & \ding{55} & \textbf{66.7} & 61.6 & \textbf{74.7} & \textbf{60.3} & \textbf{44.8} \\ \hline
\end{tabular}}
\end{table}
\begin{figure}[t]
    \centering
    \includegraphics[scale=0.70]{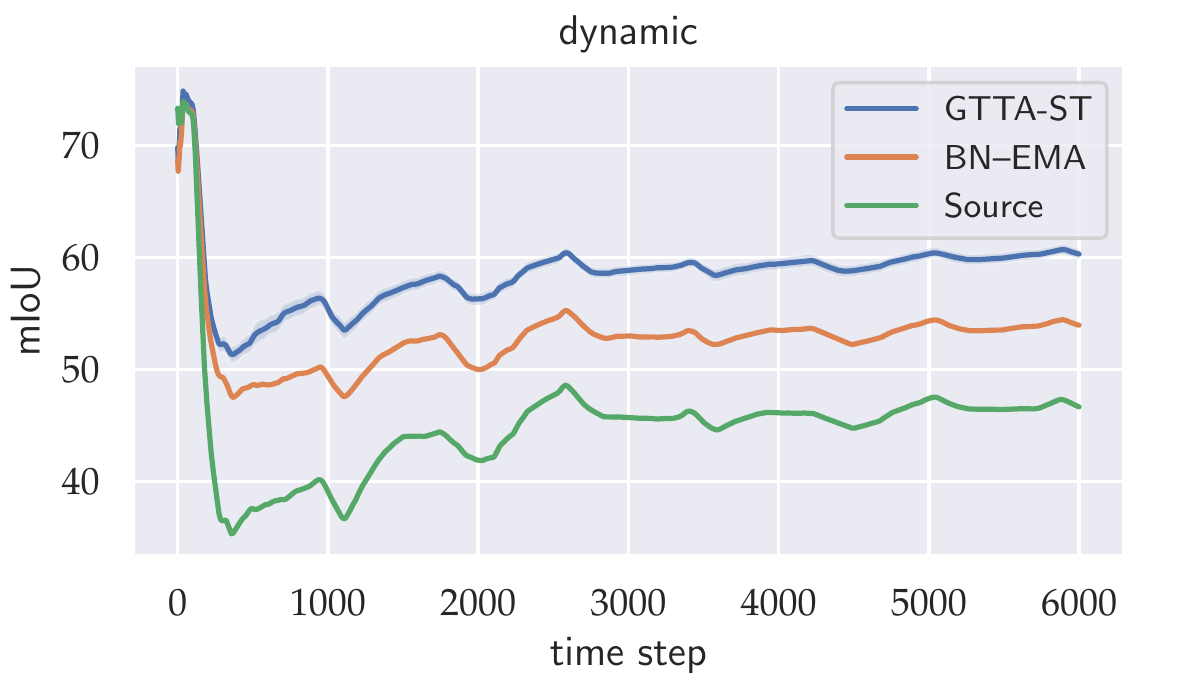}
    \caption{mIoU up to time step $t$ for the \textit{dynamic} sequence.}
    \label{fig:dynamic_sequence}
\end{figure}
Our results are summarized in Table \ref{tab:carla-main}. As expected, BN--0 (source) performs the worst by a large margin. While BN--EMA outperforms for all scenarios except \textit{highway}, BN--0.1 is absolutely $4.4\%$ better than the second best (BN--1) on the \textit{highway} split. Regarding TENT, we find that the episodic setting performs better than the continual setting. Nevertheless, both variants cannot surpass BN--1. Following \citealp{SM-PPM}, we evaluate ASM without the attention module and use 4 updates per test sample. For SM-PPM, we get the best results using 8 adaptation steps. SM-PPM performs better than TENT or ASM, however, it is still slightly worse on average compared to BN--EMA. CoTTA does not perform better than BN--1 and performance significantly drops for the longer \textit{dynamic} sequence. We attribute this to the circumstances that the mean teacher always lags behind the current test domain.

In contrast, our approach GTTA-ST substantially outperforms all baselines by a large margin. Compared to the source model, the mIoU increases by more than $8\%$ in four out of five cases. While self-training alone only provides a clear advantage in two cases, it cannot effectively exploit the gradual domain shift in this setting and even suffers from error accumulation. Style transfer, on the other hand, has a clear advantage in all evaluation settings, since it does not introduce any error accumulation due to label-noise. The combination of both methods now increases the performance on \textit{day2night}, \textit{dynamic}, and \textit{highway} as through the intermediate domain introduced by style transfer, self-training benefits from more reliable pseudo-labels.

In Figure \ref{fig:dynamic_sequence}, we illustrate the mIoU up to time step $t$ for the \textit{dynamic} sequence. While the performance of the source model suffers heavily from the domain shift, GTTA-ST is able to maintain good performance throughout the test sequence. Further visualizations and ablation studies are located in Appendix \ref{appendix:ablation_carla}.


\section{Conclusion}
In this work we addressed current challenges in online continual and gradual test-time adaptation. Through the creation of intermediate domains by mixup or style transfer, successful self-training for arbitrary domain shifts can be performed. This is supported by experiments for the various gradual changes covered by CarlaTTA and the continual and gradual corruption benchmarks. On all presented benchmarks, we outperform existing methods by a large margin. We are certain that CarlaTTA will give other researchers the opportunity to further investigate the setting of gradual test-time adaptation.


\section{Acknowledgments}
This publication was created as part of the research project "KI Delta Learning" (project number: 19A19013R) funded by the Federal Ministry for Economic Affairs and Energy (BMWi) on the basis of a decision by the German Bundestag.


\bibliographystyle{apacite}
\bibliography{egbib}

\begin{thebibliography}{}

\bibitem [\protect \citeauthoryear {%
Bartler%
, B{\"u}hler%
, Wiewel%
, D{\"o}bler%
\BCBL {}\ \BBA {} Yang%
}{%
Bartler%
\ \protect \BOthers {.}}{%
{\protect \APACyear {2022}}%
}]{%
MT3}
\APACinsertmetastar {%
MT3}%
\begin{APACrefauthors}%
Bartler, A.%
, B{\"u}hler, A.%
, Wiewel, F.%
, D{\"o}bler, M.%
\BCBL {}\ \BBA {} Yang, B.%
\end{APACrefauthors}%
\unskip\
\newblock
\APACrefYearMonthDay{2022}{}{}.
\newblock
{\BBOQ}\APACrefatitle {Mt3: Meta test-time training for self-supervised
  test-time adaption} {Mt3: Meta test-time training for self-supervised
  test-time adaption}.{\BBCQ}
\newblock
\BIn{} \APACrefbtitle {International Conference on Artificial Intelligence and
  Statistics} {International conference on artificial intelligence and
  statistics}\ (\BPGS\ 3080--3090).
\PrintBackRefs{\CurrentBib}

\bibitem [\protect \citeauthoryear {%
Bobu%
, Tzeng%
, Hoffman%
\BCBL {}\ \BBA {} Darrell%
}{%
Bobu%
\ \protect \BOthers {.}}{%
{\protect \APACyear {2018}}%
}]{%
bobu2018adapting}
\APACinsertmetastar {%
bobu2018adapting}%
\begin{APACrefauthors}%
Bobu, A.%
, Tzeng, E.%
, Hoffman, J.%
\BCBL {}\ \BBA {} Darrell, T.%
\end{APACrefauthors}%
\unskip\
\newblock
\APACrefYearMonthDay{2018}{}{}.
\newblock
{\BBOQ}\APACrefatitle {Adapting to continuously shifting domains} {Adapting to
  continuously shifting domains}.{\BBCQ}
\newblock

\PrintBackRefs{\CurrentBib}

\bibitem [\protect \citeauthoryear {%
D.~Chen%
, Wang%
, Darrell%
\BCBL {}\ \BBA {} Ebrahimi%
}{%
D.~Chen%
\ \protect \BOthers {.}}{%
{\protect \APACyear {2022}}%
}]{%
chen2022contrastive}
\APACinsertmetastar {%
chen2022contrastive}%
\begin{APACrefauthors}%
Chen, D.%
, Wang, D.%
, Darrell, T.%
\BCBL {}\ \BBA {} Ebrahimi, S.%
\end{APACrefauthors}%
\unskip\
\newblock
\APACrefYearMonthDay{2022}{}{}.
\newblock
{\BBOQ}\APACrefatitle {Contrastive Test-time Adaptation} {Contrastive test-time
  adaptation}.{\BBCQ}
\newblock
\BIn{} \APACrefbtitle {CVPR.} {Cvpr.}
\PrintBackRefs{\CurrentBib}

\bibitem [\protect \citeauthoryear {%
H\BHBI Y.~Chen%
\ \BBA {} Chao%
}{%
H\BHBI Y.~Chen%
\ \BBA {} Chao%
}{%
{\protect \APACyear {2021}}%
}]{%
IDOL}
\APACinsertmetastar {%
IDOL}%
\begin{APACrefauthors}%
Chen, H\BHBI Y.%
\BCBT {}\ \BBA {} Chao, W\BHBI L.%
\end{APACrefauthors}%
\unskip\
\newblock
\APACrefYearMonthDay{2021}{}{}.
\newblock
{\BBOQ}\APACrefatitle {Gradual Domain Adaptation without Indexed Intermediate
  Domains} {Gradual domain adaptation without indexed intermediate
  domains}.{\BBCQ}
\newblock
\APACjournalVolNumPages{Advances in Neural Information Processing
  Systems}{34}{}{}.
\PrintBackRefs{\CurrentBib}

\bibitem [\protect \citeauthoryear {%
L\BHBI C.~Chen%
, Papandreou%
, Kokkinos%
, Murphy%
\BCBL {}\ \BBA {} Yuille%
}{%
L\BHBI C.~Chen%
\ \protect \BOthers {.}}{%
{\protect \APACyear {2017}}%
}]{%
Deeplabv2}
\APACinsertmetastar {%
Deeplabv2}%
\begin{APACrefauthors}%
Chen, L\BHBI C.%
, Papandreou, G.%
, Kokkinos, I.%
, Murphy, K.%
\BCBL {}\ \BBA {} Yuille, A\BPBI L.%
\end{APACrefauthors}%
\unskip\
\newblock
\APACrefYearMonthDay{2017}{}{}.
\newblock
{\BBOQ}\APACrefatitle {Deeplab: Semantic image segmentation with deep
  convolutional nets, atrous convolution, and fully connected crfs} {Deeplab:
  Semantic image segmentation with deep convolutional nets, atrous convolution,
  and fully connected crfs}.{\BBCQ}
\newblock
\APACjournalVolNumPages{IEEE transactions on pattern analysis and machine
  intelligence}{40}{4}{834--848}.
\PrintBackRefs{\CurrentBib}

\bibitem [\protect \citeauthoryear {%
Cordts%
\ \protect \BOthers {.}}{%
Cordts%
\ \protect \BOthers {.}}{%
{\protect \APACyear {2016}}%
}]{%
cordts2016cityscapes}
\APACinsertmetastar {%
cordts2016cityscapes}%
\begin{APACrefauthors}%
Cordts, M.%
, Omran, M.%
, Ramos, S.%
, Rehfeld, T.%
, Enzweiler, M.%
, Benenson, R.%
\BDBL {}Schiele, B.%
\end{APACrefauthors}%
\unskip\
\newblock
\APACrefYearMonthDay{2016}{}{}.
\newblock
{\BBOQ}\APACrefatitle {The cityscapes dataset for semantic urban scene
  understanding} {The cityscapes dataset for semantic urban scene
  understanding}.{\BBCQ}
\newblock
\BIn{} \APACrefbtitle {Proceedings of the IEEE conference on computer vision
  and pattern recognition} {Proceedings of the ieee conference on computer
  vision and pattern recognition}\ (\BPGS\ 3213--3223).
\PrintBackRefs{\CurrentBib}

\bibitem [\protect \citeauthoryear {%
Croce%
\ \protect \BOthers {.}}{%
Croce%
\ \protect \BOthers {.}}{%
{\protect \APACyear {2020}}%
}]{%
croce2020robustbench}
\APACinsertmetastar {%
croce2020robustbench}%
\begin{APACrefauthors}%
Croce, F.%
, Andriushchenko, M.%
, Sehwag, V.%
, Debenedetti, E.%
, Flammarion, N.%
, Chiang, M.%
\BDBL {}Hein, M.%
\end{APACrefauthors}%
\unskip\
\newblock
\APACrefYearMonthDay{2020}{}{}.
\newblock
{\BBOQ}\APACrefatitle {Robustbench: a standardized adversarial robustness
  benchmark} {Robustbench: a standardized adversarial robustness
  benchmark}.{\BBCQ}
\newblock
\APACjournalVolNumPages{arXiv preprint arXiv:2010.09670}{}{}{}.
\PrintBackRefs{\CurrentBib}

\bibitem [\protect \citeauthoryear {%
Deng%
\ \protect \BOthers {.}}{%
Deng%
\ \protect \BOthers {.}}{%
{\protect \APACyear {2009}}%
}]{%
imagenet_cvpr09}
\APACinsertmetastar {%
imagenet_cvpr09}%
\begin{APACrefauthors}%
Deng, J.%
, Dong, W.%
, Socher, R.%
, Li, L\BHBI J.%
, Li, K.%
\BCBL {}\ \BBA {} Fei-Fei, L.%
\end{APACrefauthors}%
\unskip\
\newblock
\APACrefYearMonthDay{2009}{}{}.
\newblock
{\BBOQ}\APACrefatitle {{ImageNet: A Large-Scale Hierarchical Image Database}}
  {{ImageNet: A Large-Scale Hierarchical Image Database}}.{\BBCQ}
\newblock
\BIn{} \APACrefbtitle {CVPR09.} {Cvpr09.}
\PrintBackRefs{\CurrentBib}

\bibitem [\protect \citeauthoryear {%
Dosovitskiy%
, Ros%
, Codevilla%
, Lopez%
\BCBL {}\ \BBA {} Koltun%
}{%
Dosovitskiy%
\ \protect \BOthers {.}}{%
{\protect \APACyear {2017}}%
}]{%
osovitskiy17carla}
\APACinsertmetastar {%
osovitskiy17carla}%
\begin{APACrefauthors}%
Dosovitskiy, A.%
, Ros, G.%
, Codevilla, F.%
, Lopez, A.%
\BCBL {}\ \BBA {} Koltun, V.%
\end{APACrefauthors}%
\unskip\
\newblock
\APACrefYearMonthDay{2017}{}{}.
\newblock
{\BBOQ}\APACrefatitle {{CARLA}: {An} Open Urban Driving Simulator} {{CARLA}:
  {An} open urban driving simulator}.{\BBCQ}
\newblock
\BIn{} \APACrefbtitle {Proceedings of the 1st Annual Conference on Robot
  Learning} {Proceedings of the 1st annual conference on robot learning}\
  (\BPGS\ 1--16).
\PrintBackRefs{\CurrentBib}

\bibitem [\protect \citeauthoryear {%
Ganin%
\ \BBA {} Lempitsky%
}{%
Ganin%
\ \BBA {} Lempitsky%
}{%
{\protect \APACyear {2015}}%
}]{%
DANN}
\APACinsertmetastar {%
DANN}%
\begin{APACrefauthors}%
Ganin, Y.%
\BCBT {}\ \BBA {} Lempitsky, V.%
\end{APACrefauthors}%
\unskip\
\newblock
\APACrefYearMonthDay{2015}{}{}.
\newblock
{\BBOQ}\APACrefatitle {Unsupervised domain adaptation by backpropagation}
  {Unsupervised domain adaptation by backpropagation}.{\BBCQ}
\newblock
\BIn{} \APACrefbtitle {International conference on machine learning}
  {International conference on machine learning}\ (\BPGS\ 1180--1189).
\PrintBackRefs{\CurrentBib}

\bibitem [\protect \citeauthoryear {%
Hendrycks%
\ \protect \BOthers {.}}{%
Hendrycks%
\ \protect \BOthers {.}}{%
{\protect \APACyear {2021}}%
}]{%
hendrycks2021many}
\APACinsertmetastar {%
hendrycks2021many}%
\begin{APACrefauthors}%
Hendrycks, D.%
, Basart, S.%
, Mu, N.%
, Kadavath, S.%
, Wang, F.%
, Dorundo, E.%
\BDBL {}others%
\end{APACrefauthors}%
\unskip\
\newblock
\APACrefYearMonthDay{2021}{}{}.
\newblock
{\BBOQ}\APACrefatitle {The many faces of robustness: A critical analysis of
  out-of-distribution generalization} {The many faces of robustness: A critical
  analysis of out-of-distribution generalization}.{\BBCQ}
\newblock
\BIn{} \APACrefbtitle {Proceedings of the IEEE/CVF International Conference on
  Computer Vision} {Proceedings of the ieee/cvf international conference on
  computer vision}\ (\BPGS\ 8340--8349).
\PrintBackRefs{\CurrentBib}

\bibitem [\protect \citeauthoryear {%
Hendrycks%
\ \BBA {} Dietterich%
}{%
Hendrycks%
\ \BBA {} Dietterich%
}{%
{\protect \APACyear {2019}}%
}]{%
hendrycks2019benchmarking}
\APACinsertmetastar {%
hendrycks2019benchmarking}%
\begin{APACrefauthors}%
Hendrycks, D.%
\BCBT {}\ \BBA {} Dietterich, T.%
\end{APACrefauthors}%
\unskip\
\newblock
\APACrefYearMonthDay{2019}{}{}.
\newblock
{\BBOQ}\APACrefatitle {Benchmarking neural network robustness to common
  corruptions and perturbations} {Benchmarking neural network robustness to
  common corruptions and perturbations}.{\BBCQ}
\newblock
\APACjournalVolNumPages{arXiv preprint arXiv:1903.12261}{}{}{}.
\PrintBackRefs{\CurrentBib}

\bibitem [\protect \citeauthoryear {%
Hendrycks%
\ \protect \BOthers {.}}{%
Hendrycks%
\ \protect \BOthers {.}}{%
{\protect \APACyear {2019}}%
}]{%
hendrycks2019augmix}
\APACinsertmetastar {%
hendrycks2019augmix}%
\begin{APACrefauthors}%
Hendrycks, D.%
, Mu, N.%
, Cubuk, E\BPBI D.%
, Zoph, B.%
, Gilmer, J.%
\BCBL {}\ \BBA {} Lakshminarayanan, B.%
\end{APACrefauthors}%
\unskip\
\newblock
\APACrefYearMonthDay{2019}{}{}.
\newblock
{\BBOQ}\APACrefatitle {Augmix: A simple data processing method to improve
  robustness and uncertainty} {Augmix: A simple data processing method to
  improve robustness and uncertainty}.{\BBCQ}
\newblock
\APACjournalVolNumPages{arXiv preprint arXiv:1912.02781}{}{}{}.
\PrintBackRefs{\CurrentBib}

\bibitem [\protect \citeauthoryear {%
Hoffman%
, Darrell%
\BCBL {}\ \BBA {} Saenko%
}{%
Hoffman%
\ \protect \BOthers {.}}{%
{\protect \APACyear {2014}}%
}]{%
hoffman14}
\APACinsertmetastar {%
hoffman14}%
\begin{APACrefauthors}%
Hoffman, J.%
, Darrell, T.%
\BCBL {}\ \BBA {} Saenko, K.%
\end{APACrefauthors}%
\unskip\
\newblock
\APACrefYearMonthDay{2014}{}{}.
\newblock
{\BBOQ}\APACrefatitle {Continuous manifold based adaptation for evolving visual
  domains} {Continuous manifold based adaptation for evolving visual
  domains}.{\BBCQ}
\newblock
\BIn{} \APACrefbtitle {Proceedings of the IEEE Conference on Computer Vision
  and Pattern Recognition} {Proceedings of the ieee conference on computer
  vision and pattern recognition}\ (\BPGS\ 867--874).
\PrintBackRefs{\CurrentBib}

\bibitem [\protect \citeauthoryear {%
Hoffman%
\ \protect \BOthers {.}}{%
Hoffman%
\ \protect \BOthers {.}}{%
{\protect \APACyear {2018}}%
}]{%
CYCADA}
\APACinsertmetastar {%
CYCADA}%
\begin{APACrefauthors}%
Hoffman, J.%
, Tzeng, E.%
, Park, T.%
, Zhu, J\BHBI Y.%
, Isola, P.%
, Saenko, K.%
\BDBL {}Darrell, T.%
\end{APACrefauthors}%
\unskip\
\newblock
\APACrefYearMonthDay{2018}{}{}.
\newblock
{\BBOQ}\APACrefatitle {Cycada: Cycle-consistent adversarial domain adaptation}
  {Cycada: Cycle-consistent adversarial domain adaptation}.{\BBCQ}
\newblock
\BIn{} \APACrefbtitle {International conference on machine learning}
  {International conference on machine learning}\ (\BPGS\ 1989--1998).
\PrintBackRefs{\CurrentBib}

\bibitem [\protect \citeauthoryear {%
Hoyer%
, Dai%
\BCBL {}\ \BBA {} Van~Gool%
}{%
Hoyer%
\ \protect \BOthers {.}}{%
{\protect \APACyear {2022}}%
}]{%
hoyer2022daformer}
\APACinsertmetastar {%
hoyer2022daformer}%
\begin{APACrefauthors}%
Hoyer, L.%
, Dai, D.%
\BCBL {}\ \BBA {} Van~Gool, L.%
\end{APACrefauthors}%
\unskip\
\newblock
\APACrefYearMonthDay{2022}{}{}.
\newblock
{\BBOQ}\APACrefatitle {Daformer: Improving network architectures and training
  strategies for domain-adaptive semantic segmentation} {Daformer: Improving
  network architectures and training strategies for domain-adaptive semantic
  segmentation}.{\BBCQ}
\newblock
\BIn{} \APACrefbtitle {Proceedings of the IEEE/CVF Conference on Computer
  Vision and Pattern Recognition} {Proceedings of the ieee/cvf conference on
  computer vision and pattern recognition}\ (\BPGS\ 9924--9935).
\PrintBackRefs{\CurrentBib}

\bibitem [\protect \citeauthoryear {%
Huang%
\ \BBA {} Belongie%
}{%
Huang%
\ \BBA {} Belongie%
}{%
{\protect \APACyear {2017}}%
}]{%
AdaIN}
\APACinsertmetastar {%
AdaIN}%
\begin{APACrefauthors}%
Huang, X.%
\BCBT {}\ \BBA {} Belongie, S.%
\end{APACrefauthors}%
\unskip\
\newblock
\APACrefYearMonthDay{2017}{}{}.
\newblock
{\BBOQ}\APACrefatitle {Arbitrary style transfer in real-time with adaptive
  instance normalization} {Arbitrary style transfer in real-time with adaptive
  instance normalization}.{\BBCQ}
\newblock
\BIn{} \APACrefbtitle {Proceedings of the IEEE International Conference on
  Computer Vision} {Proceedings of the ieee international conference on
  computer vision}\ (\BPGS\ 1501--1510).
\PrintBackRefs{\CurrentBib}

\bibitem [\protect \citeauthoryear {%
Isola%
, Zhu%
, Zhou%
\BCBL {}\ \BBA {} Efros%
}{%
Isola%
\ \protect \BOthers {.}}{%
{\protect \APACyear {2017}}%
}]{%
isola2017image}
\APACinsertmetastar {%
isola2017image}%
\begin{APACrefauthors}%
Isola, P.%
, Zhu, J\BHBI Y.%
, Zhou, T.%
\BCBL {}\ \BBA {} Efros, A\BPBI A.%
\end{APACrefauthors}%
\unskip\
\newblock
\APACrefYearMonthDay{2017}{}{}.
\newblock
{\BBOQ}\APACrefatitle {Image-to-image translation with conditional adversarial
  networks} {Image-to-image translation with conditional adversarial
  networks}.{\BBCQ}
\newblock
\BIn{} \APACrefbtitle {Proceedings of the IEEE conference on computer vision
  and pattern recognition} {Proceedings of the ieee conference on computer
  vision and pattern recognition}\ (\BPGS\ 1125--1134).
\PrintBackRefs{\CurrentBib}

\bibitem [\protect \citeauthoryear {%
Kim%
, Kim%
\BCBL {}\ \BBA {} Kim%
}{%
Kim%
\ \protect \BOthers {.}}{%
{\protect \APACyear {2021}}%
}]{%
kim2021deep}
\APACinsertmetastar {%
kim2021deep}%
\begin{APACrefauthors}%
Kim, S.%
, Kim, S.%
\BCBL {}\ \BBA {} Kim, S.%
\end{APACrefauthors}%
\unskip\
\newblock
\APACrefYearMonthDay{2021}{}{}.
\newblock
{\BBOQ}\APACrefatitle {Deep Translation Prior: Test-time Training for
  Photorealistic Style Transfer} {Deep translation prior: Test-time training
  for photorealistic style transfer}.{\BBCQ}
\newblock
\APACjournalVolNumPages{arXiv preprint arXiv:2112.06150}{}{}{}.
\PrintBackRefs{\CurrentBib}

\bibitem [\protect \citeauthoryear {%
Kingma%
\ \BBA {} Ba%
}{%
Kingma%
\ \BBA {} Ba%
}{%
{\protect \APACyear {2014}}%
}]{%
kingma2014adam}
\APACinsertmetastar {%
kingma2014adam}%
\begin{APACrefauthors}%
Kingma, D\BPBI P.%
\BCBT {}\ \BBA {} Ba, J.%
\end{APACrefauthors}%
\unskip\
\newblock
\APACrefYearMonthDay{2014}{}{}.
\newblock
{\BBOQ}\APACrefatitle {Adam: A method for stochastic optimization} {Adam: A
  method for stochastic optimization}.{\BBCQ}
\newblock
\APACjournalVolNumPages{arXiv preprint arXiv:1412.6980}{}{}{}.
\PrintBackRefs{\CurrentBib}

\bibitem [\protect \citeauthoryear {%
Krizhevsky%
, Hinton%
\BCBL {}\ \protect \BOthers {.}}{%
Krizhevsky%
\ \protect \BOthers {.}}{%
{\protect \APACyear {2009}}%
}]{%
krizhevsky2009learning}
\APACinsertmetastar {%
krizhevsky2009learning}%
\begin{APACrefauthors}%
Krizhevsky, A.%
, Hinton, G.%
\BCBL {}\ \BOthersPeriod {.}\end{APACrefauthors}%
\unskip\
\newblock
\APACrefYearMonthDay{2009}{}{}.
\newblock
{\BBOQ}\APACrefatitle {Learning multiple layers of features from tiny images}
  {Learning multiple layers of features from tiny images}.{\BBCQ}
\newblock

\PrintBackRefs{\CurrentBib}

\bibitem [\protect \citeauthoryear {%
Kumar%
, Ma%
\BCBL {}\ \BBA {} Liang%
}{%
Kumar%
\ \protect \BOthers {.}}{%
{\protect \APACyear {2020}}%
}]{%
kumar2020understanding}
\APACinsertmetastar {%
kumar2020understanding}%
\begin{APACrefauthors}%
Kumar, A.%
, Ma, T.%
\BCBL {}\ \BBA {} Liang, P.%
\end{APACrefauthors}%
\unskip\
\newblock
\APACrefYearMonthDay{2020}{}{}.
\newblock
{\BBOQ}\APACrefatitle {Understanding self-training for gradual domain
  adaptation} {Understanding self-training for gradual domain
  adaptation}.{\BBCQ}
\newblock
\BIn{} \APACrefbtitle {International Conference on Machine Learning}
  {International conference on machine learning}\ (\BPGS\ 5468--5479).
\PrintBackRefs{\CurrentBib}

\bibitem [\protect \citeauthoryear {%
G.~Li%
, Kang%
, Liu%
, Wei%
\BCBL {}\ \BBA {} Yang%
}{%
G.~Li%
\ \protect \BOthers {.}}{%
{\protect \APACyear {2020}}%
}]{%
CCM}
\APACinsertmetastar {%
CCM}%
\begin{APACrefauthors}%
Li, G.%
, Kang, G.%
, Liu, W.%
, Wei, Y.%
\BCBL {}\ \BBA {} Yang, Y.%
\end{APACrefauthors}%
\unskip\
\newblock
\APACrefYearMonthDay{2020}{}{}.
\newblock
{\BBOQ}\APACrefatitle {Content-consistent matching for domain adaptive semantic
  segmentation} {Content-consistent matching for domain adaptive semantic
  segmentation}.{\BBCQ}
\newblock
\BIn{} \APACrefbtitle {European Conference on Computer Vision} {European
  conference on computer vision}\ (\BPGS\ 440--456).
\PrintBackRefs{\CurrentBib}

\bibitem [\protect \citeauthoryear {%
Y.~Li%
, Yuan%
\BCBL {}\ \BBA {} Vasconcelos%
}{%
Y.~Li%
\ \protect \BOthers {.}}{%
{\protect \APACyear {2019}}%
}]{%
Bidirectional}
\APACinsertmetastar {%
Bidirectional}%
\begin{APACrefauthors}%
Li, Y.%
, Yuan, L.%
\BCBL {}\ \BBA {} Vasconcelos, N.%
\end{APACrefauthors}%
\unskip\
\newblock
\APACrefYearMonthDay{2019}{}{}.
\newblock
{\BBOQ}\APACrefatitle {Bidirectional learning for domain adaptation of semantic
  segmentation} {Bidirectional learning for domain adaptation of semantic
  segmentation}.{\BBCQ}
\newblock
\BIn{} \APACrefbtitle {Proceedings of the IEEE Conference on Computer Vision
  and Pattern Recognition} {Proceedings of the ieee conference on computer
  vision and pattern recognition}\ (\BPGS\ 6936--6945).
\PrintBackRefs{\CurrentBib}

\bibitem [\protect \citeauthoryear {%
Liang%
, Hu%
\BCBL {}\ \BBA {} Feng%
}{%
Liang%
\ \protect \BOthers {.}}{%
{\protect \APACyear {2020}}%
}]{%
liang2020we}
\APACinsertmetastar {%
liang2020we}%
\begin{APACrefauthors}%
Liang, J.%
, Hu, D.%
\BCBL {}\ \BBA {} Feng, J.%
\end{APACrefauthors}%
\unskip\
\newblock
\APACrefYearMonthDay{2020}{}{}.
\newblock
{\BBOQ}\APACrefatitle {Do we really need to access the source data? source
  hypothesis transfer for unsupervised domain adaptation} {Do we really need to
  access the source data? source hypothesis transfer for unsupervised domain
  adaptation}.{\BBCQ}
\newblock
\BIn{} \APACrefbtitle {International Conference on Machine Learning}
  {International conference on machine learning}\ (\BPGS\ 6028--6039).
\PrintBackRefs{\CurrentBib}

\bibitem [\protect \citeauthoryear {%
Liu%
\ \protect \BOthers {.}}{%
Liu%
\ \protect \BOthers {.}}{%
{\protect \APACyear {2021}}%
}]{%
liu2021ttt++}
\APACinsertmetastar {%
liu2021ttt++}%
\begin{APACrefauthors}%
Liu, Y.%
, Kothari, P.%
, van Delft, B.%
, Bellot-Gurlet, B.%
, Mordan, T.%
\BCBL {}\ \BBA {} Alahi, A.%
\end{APACrefauthors}%
\unskip\
\newblock
\APACrefYearMonthDay{2021}{}{}.
\newblock
{\BBOQ}\APACrefatitle {TTT++: When Does Self-Supervised Test-Time Training Fail
  or Thrive?} {Ttt++: When does self-supervised test-time training fail or
  thrive?}{\BBCQ}
\newblock
\APACjournalVolNumPages{Advances in Neural Information Processing
  Systems}{34}{}{}.
\PrintBackRefs{\CurrentBib}

\bibitem [\protect \citeauthoryear {%
Luo%
, Liu%
, Guan%
, Yu%
\BCBL {}\ \BBA {} Yang%
}{%
Luo%
\ \protect \BOthers {.}}{%
{\protect \APACyear {2020}}%
}]{%
ASM}
\APACinsertmetastar {%
ASM}%
\begin{APACrefauthors}%
Luo, Y.%
, Liu, P.%
, Guan, T.%
, Yu, J.%
\BCBL {}\ \BBA {} Yang, Y.%
\end{APACrefauthors}%
\unskip\
\newblock
\APACrefYearMonthDay{2020}{}{}.
\newblock
{\BBOQ}\APACrefatitle {Adversarial style mining for one-shot unsupervised
  domain adaptation} {Adversarial style mining for one-shot unsupervised domain
  adaptation}.{\BBCQ}
\newblock
\APACjournalVolNumPages{Advances in Neural Information Processing
  Systems}{33}{}{20612--20623}.
\PrintBackRefs{\CurrentBib}

\bibitem [\protect \citeauthoryear {%
Marsden%
, Bartler%
, D{\"o}bler%
\BCBL {}\ \BBA {} Yang%
}{%
Marsden%
, Bartler%
\BCBL {}\ \protect \BOthers {.}}{%
{\protect \APACyear {2022}}%
}]{%
marsden2022contrastive}
\APACinsertmetastar {%
marsden2022contrastive}%
\begin{APACrefauthors}%
Marsden, R\BPBI A.%
, Bartler, A.%
, D{\"o}bler, M.%
\BCBL {}\ \BBA {} Yang, B.%
\end{APACrefauthors}%
\unskip\
\newblock
\APACrefYearMonthDay{2022}{}{}.
\newblock
{\BBOQ}\APACrefatitle {Contrastive learning and self-training for unsupervised
  domain adaptation in semantic segmentation} {Contrastive learning and
  self-training for unsupervised domain adaptation in semantic
  segmentation}.{\BBCQ}
\newblock
\BIn{} \APACrefbtitle {2022 International Joint Conference on Neural Networks
  (IJCNN)} {2022 international joint conference on neural networks (ijcnn)}\
  (\BPGS\ 1--8).
\PrintBackRefs{\CurrentBib}

\bibitem [\protect \citeauthoryear {%
Marsden%
, Wiewel%
, D{\"o}bler%
, Yang%
\BCBL {}\ \BBA {} Yang%
}{%
Marsden%
, Wiewel%
\BCBL {}\ \protect \BOthers {.}}{%
{\protect \APACyear {2022}}%
}]{%
CACE}
\APACinsertmetastar {%
CACE}%
\begin{APACrefauthors}%
Marsden, R\BPBI A.%
, Wiewel, F.%
, D{\"o}bler, M.%
, Yang, Y.%
\BCBL {}\ \BBA {} Yang, B.%
\end{APACrefauthors}%
\unskip\
\newblock
\APACrefYearMonthDay{2022}{}{}.
\newblock
{\BBOQ}\APACrefatitle {Continual Unsupervised Domain Adaptation for Semantic
  Segmentation using a Class-Specific Transfer} {Continual unsupervised domain
  adaptation for semantic segmentation using a class-specific transfer}.{\BBCQ}
\newblock
\BIn{} \APACrefbtitle {2022 International Joint Conference on Neural Networks
  (IJCNN)} {2022 international joint conference on neural networks (ijcnn)}\
  (\BPGS\ 1--8).
\PrintBackRefs{\CurrentBib}

\bibitem [\protect \citeauthoryear {%
McCloskey%
\ \BBA {} Cohen%
}{%
McCloskey%
\ \BBA {} Cohen%
}{%
{\protect \APACyear {1989}}%
}]{%
mccloskey1989catastrophic}
\APACinsertmetastar {%
mccloskey1989catastrophic}%
\begin{APACrefauthors}%
McCloskey, M.%
\BCBT {}\ \BBA {} Cohen, N\BPBI J.%
\end{APACrefauthors}%
\unskip\
\newblock
\APACrefYearMonthDay{1989}{}{}.
\newblock
{\BBOQ}\APACrefatitle {Catastrophic interference in connectionist networks: The
  sequential learning problem} {Catastrophic interference in connectionist
  networks: The sequential learning problem}.{\BBCQ}
\newblock
\BIn{} \APACrefbtitle {Psychology of learning and motivation} {Psychology of
  learning and motivation}\ (\BVOL~24, \BPGS\ 109--165).
\newblock
\APACaddressPublisher{}{Elsevier}.
\PrintBackRefs{\CurrentBib}

\bibitem [\protect \citeauthoryear {%
Mei%
, Zhu%
, Zou%
\BCBL {}\ \BBA {} Zhang%
}{%
Mei%
\ \protect \BOthers {.}}{%
{\protect \APACyear {2020}}%
}]{%
IAST}
\APACinsertmetastar {%
IAST}%
\begin{APACrefauthors}%
Mei, K.%
, Zhu, C.%
, Zou, J.%
\BCBL {}\ \BBA {} Zhang, S.%
\end{APACrefauthors}%
\unskip\
\newblock
\APACrefYearMonthDay{2020}{}{}.
\newblock
{\BBOQ}\APACrefatitle {Instance adaptive self-training for unsupervised domain
  adaptation} {Instance adaptive self-training for unsupervised domain
  adaptation}.{\BBCQ}
\newblock
\APACjournalVolNumPages{arXiv preprint arXiv:2008.12197}{}{}{}.
\PrintBackRefs{\CurrentBib}

\bibitem [\protect \citeauthoryear {%
Mintun%
, Kirillov%
\BCBL {}\ \BBA {} Xie%
}{%
Mintun%
\ \protect \BOthers {.}}{%
{\protect \APACyear {2021}}%
}]{%
mintun2021interaction}
\APACinsertmetastar {%
mintun2021interaction}%
\begin{APACrefauthors}%
Mintun, E.%
, Kirillov, A.%
\BCBL {}\ \BBA {} Xie, S.%
\end{APACrefauthors}%
\unskip\
\newblock
\APACrefYearMonthDay{2021}{}{}.
\newblock
{\BBOQ}\APACrefatitle {On interaction between augmentations and corruptions in
  natural corruption robustness} {On interaction between augmentations and
  corruptions in natural corruption robustness}.{\BBCQ}
\newblock
\APACjournalVolNumPages{Advances in Neural Information Processing
  Systems}{34}{}{}.
\PrintBackRefs{\CurrentBib}

\bibitem [\protect \citeauthoryear {%
Muandet%
, Balduzzi%
\BCBL {}\ \BBA {} Sch{\"o}lkopf%
}{%
Muandet%
\ \protect \BOthers {.}}{%
{\protect \APACyear {2013}}%
}]{%
muandet2013domain}
\APACinsertmetastar {%
muandet2013domain}%
\begin{APACrefauthors}%
Muandet, K.%
, Balduzzi, D.%
\BCBL {}\ \BBA {} Sch{\"o}lkopf, B.%
\end{APACrefauthors}%
\unskip\
\newblock
\APACrefYearMonthDay{2013}{}{}.
\newblock
{\BBOQ}\APACrefatitle {Domain generalization via invariant feature
  representation} {Domain generalization via invariant feature
  representation}.{\BBCQ}
\newblock
\BIn{} \APACrefbtitle {International Conference on Machine Learning}
  {International conference on machine learning}\ (\BPGS\ 10--18).
\PrintBackRefs{\CurrentBib}

\bibitem [\protect \citeauthoryear {%
Mummadi%
\ \protect \BOthers {.}}{%
Mummadi%
\ \protect \BOthers {.}}{%
{\protect \APACyear {2021}}%
}]{%
mummadi2021test}
\APACinsertmetastar {%
mummadi2021test}%
\begin{APACrefauthors}%
Mummadi, C\BPBI K.%
, Hutmacher, R.%
, Rambach, K.%
, Levinkov, E.%
, Brox, T.%
\BCBL {}\ \BBA {} Metzen, J\BPBI H.%
\end{APACrefauthors}%
\unskip\
\newblock
\APACrefYearMonthDay{2021}{}{}.
\newblock
{\BBOQ}\APACrefatitle {Test-time adaptation to distribution shift by confidence
  maximization and input transformation} {Test-time adaptation to distribution
  shift by confidence maximization and input transformation}.{\BBCQ}
\newblock
\APACjournalVolNumPages{arXiv preprint arXiv:2106.14999}{}{}{}.
\PrintBackRefs{\CurrentBib}

\bibitem [\protect \citeauthoryear {%
Qui{\~n}onero-Candela%
, Sugiyama%
, Schwaighofer%
\BCBL {}\ \BBA {} Lawrence%
}{%
Qui{\~n}onero-Candela%
\ \protect \BOthers {.}}{%
{\protect \APACyear {2008}}%
}]{%
quinonero2008dataset}
\APACinsertmetastar {%
quinonero2008dataset}%
\begin{APACrefauthors}%
Qui{\~n}onero-Candela, J.%
, Sugiyama, M.%
, Schwaighofer, A.%
\BCBL {}\ \BBA {} Lawrence, N\BPBI D.%
\end{APACrefauthors}%
\unskip\
\newblock
\APACrefYear{2008}.
\newblock
\APACrefbtitle {Dataset shift in machine learning} {Dataset shift in machine
  learning}.
\newblock
\APACaddressPublisher{}{Mit Press}.
\PrintBackRefs{\CurrentBib}

\bibitem [\protect \citeauthoryear {%
Richter%
, Vineet%
, Roth%
\BCBL {}\ \BBA {} Koltun%
}{%
Richter%
\ \protect \BOthers {.}}{%
{\protect \APACyear {2016}}%
}]{%
Richter_2016_ECCV}
\APACinsertmetastar {%
Richter_2016_ECCV}%
\begin{APACrefauthors}%
Richter, S\BPBI R.%
, Vineet, V.%
, Roth, S.%
\BCBL {}\ \BBA {} Koltun, V.%
\end{APACrefauthors}%
\unskip\
\newblock
\APACrefYearMonthDay{2016}{}{}.
\newblock
{\BBOQ}\APACrefatitle {Playing for Data: {G}round Truth from Computer Games}
  {Playing for data: {G}round truth from computer games}.{\BBCQ}
\newblock
\BIn{} B.~Leibe, J.~Matas, N.~Sebe\BCBL {}\ \BBA {} M.~Welling\ (\BEDS),
  \APACrefbtitle {European Conference on Computer Vision (ECCV)} {European
  conference on computer vision (eccv)}\ (\BVOL\ 9906, \BPGS\ 102--118).
\newblock
\APACaddressPublisher{}{Springer International Publishing}.
\PrintBackRefs{\CurrentBib}

\bibitem [\protect \citeauthoryear {%
Ros%
, Sellart%
, Materzynska%
, Vazquez%
\BCBL {}\ \BBA {} Lopez%
}{%
Ros%
\ \protect \BOthers {.}}{%
{\protect \APACyear {2016}}%
}]{%
ros2016synthia}
\APACinsertmetastar {%
ros2016synthia}%
\begin{APACrefauthors}%
Ros, G.%
, Sellart, L.%
, Materzynska, J.%
, Vazquez, D.%
\BCBL {}\ \BBA {} Lopez, A\BPBI M.%
\end{APACrefauthors}%
\unskip\
\newblock
\APACrefYearMonthDay{2016}{}{}.
\newblock
{\BBOQ}\APACrefatitle {The synthia dataset: A large collection of synthetic
  images for semantic segmentation of urban scenes} {The synthia dataset: A
  large collection of synthetic images for semantic segmentation of urban
  scenes}.{\BBCQ}
\newblock
\BIn{} \APACrefbtitle {Proceedings of the IEEE conference on computer vision
  and pattern recognition} {Proceedings of the ieee conference on computer
  vision and pattern recognition}\ (\BPGS\ 3234--3243).
\PrintBackRefs{\CurrentBib}

\bibitem [\protect \citeauthoryear {%
Sakaridis%
, Dai%
\BCBL {}\ \BBA {} Van~Gool%
}{%
Sakaridis%
\ \protect \BOthers {.}}{%
{\protect \APACyear {2021}}%
}]{%
sakaridis2021acdc}
\APACinsertmetastar {%
sakaridis2021acdc}%
\begin{APACrefauthors}%
Sakaridis, C.%
, Dai, D.%
\BCBL {}\ \BBA {} Van~Gool, L.%
\end{APACrefauthors}%
\unskip\
\newblock
\APACrefYearMonthDay{2021}{}{}.
\newblock
{\BBOQ}\APACrefatitle {ACDC: The adverse conditions dataset with
  correspondences for semantic driving scene understanding} {Acdc: The adverse
  conditions dataset with correspondences for semantic driving scene
  understanding}.{\BBCQ}
\newblock
\BIn{} \APACrefbtitle {Proceedings of the IEEE/CVF International Conference on
  Computer Vision} {Proceedings of the ieee/cvf international conference on
  computer vision}\ (\BPGS\ 10765--10775).
\PrintBackRefs{\CurrentBib}

\bibitem [\protect \citeauthoryear {%
Schneider%
\ \protect \BOthers {.}}{%
Schneider%
\ \protect \BOthers {.}}{%
{\protect \APACyear {2020}}%
}]{%
schneider2020improving}
\APACinsertmetastar {%
schneider2020improving}%
\begin{APACrefauthors}%
Schneider, S.%
, Rusak, E.%
, Eck, L.%
, Bringmann, O.%
, Brendel, W.%
\BCBL {}\ \BBA {} Bethge, M.%
\end{APACrefauthors}%
\unskip\
\newblock
\APACrefYearMonthDay{2020}{}{}.
\newblock
{\BBOQ}\APACrefatitle {Improving robustness against common corruptions by
  covariate shift adaptation} {Improving robustness against common corruptions
  by covariate shift adaptation}.{\BBCQ}
\newblock
\APACjournalVolNumPages{Advances in Neural Information Processing
  Systems}{33}{}{11539--11551}.
\PrintBackRefs{\CurrentBib}

\bibitem [\protect \citeauthoryear {%
P.~Sun%
\ \protect \BOthers {.}}{%
P.~Sun%
\ \protect \BOthers {.}}{%
{\protect \APACyear {2020}}%
}]{%
sun2020scalability}
\APACinsertmetastar {%
sun2020scalability}%
\begin{APACrefauthors}%
Sun, P.%
, Kretzschmar, H.%
, Dotiwalla, X.%
, Chouard, A.%
, Patnaik, V.%
, Tsui, P.%
\BDBL {}others%
\end{APACrefauthors}%
\unskip\
\newblock
\APACrefYearMonthDay{2020}{}{}.
\newblock
{\BBOQ}\APACrefatitle {Scalability in perception for autonomous driving: Waymo
  open dataset} {Scalability in perception for autonomous driving: Waymo open
  dataset}.{\BBCQ}
\newblock
\BIn{} \APACrefbtitle {Proceedings of the IEEE/CVF conference on computer
  vision and pattern recognition} {Proceedings of the ieee/cvf conference on
  computer vision and pattern recognition}\ (\BPGS\ 2446--2454).
\PrintBackRefs{\CurrentBib}

\bibitem [\protect \citeauthoryear {%
Y.~Sun%
\ \protect \BOthers {.}}{%
Y.~Sun%
\ \protect \BOthers {.}}{%
{\protect \APACyear {2020}}%
}]{%
TTT}
\APACinsertmetastar {%
TTT}%
\begin{APACrefauthors}%
Sun, Y.%
, Wang, X.%
, Liu, Z.%
, Miller, J.%
, Efros, A.%
\BCBL {}\ \BBA {} Hardt, M.%
\end{APACrefauthors}%
\unskip\
\newblock
\APACrefYearMonthDay{2020}{}{}.
\newblock
{\BBOQ}\APACrefatitle {Test-time training with self-supervision for
  generalization under distribution shifts} {Test-time training with
  self-supervision for generalization under distribution shifts}.{\BBCQ}
\newblock
\BIn{} \APACrefbtitle {International Conference on Machine Learning}
  {International conference on machine learning}\ (\BPGS\ 9229--9248).
\PrintBackRefs{\CurrentBib}

\bibitem [\protect \citeauthoryear {%
Tobin%
\ \protect \BOthers {.}}{%
Tobin%
\ \protect \BOthers {.}}{%
{\protect \APACyear {2017}}%
}]{%
tobin2017domain}
\APACinsertmetastar {%
tobin2017domain}%
\begin{APACrefauthors}%
Tobin, J.%
, Fong, R.%
, Ray, A.%
, Schneider, J.%
, Zaremba, W.%
\BCBL {}\ \BBA {} Abbeel, P.%
\end{APACrefauthors}%
\unskip\
\newblock
\APACrefYearMonthDay{2017}{}{}.
\newblock
{\BBOQ}\APACrefatitle {Domain randomization for transferring deep neural
  networks from simulation to the real world} {Domain randomization for
  transferring deep neural networks from simulation to the real world}.{\BBCQ}
\newblock
\BIn{} \APACrefbtitle {2017 IEEE/RSJ international conference on intelligent
  robots and systems (IROS)} {2017 ieee/rsj international conference on
  intelligent robots and systems (iros)}\ (\BPGS\ 23--30).
\PrintBackRefs{\CurrentBib}

\bibitem [\protect \citeauthoryear {%
Tranheden%
, Olsson%
, Pinto%
\BCBL {}\ \BBA {} Svensson%
}{%
Tranheden%
\ \protect \BOthers {.}}{%
{\protect \APACyear {2021}}%
}]{%
DACS}
\APACinsertmetastar {%
DACS}%
\begin{APACrefauthors}%
Tranheden, W.%
, Olsson, V.%
, Pinto, J.%
\BCBL {}\ \BBA {} Svensson, L.%
\end{APACrefauthors}%
\unskip\
\newblock
\APACrefYearMonthDay{2021}{}{}.
\newblock
{\BBOQ}\APACrefatitle {DACS: Domain Adaptation via Cross-domain Mixed Sampling}
  {Dacs: Domain adaptation via cross-domain mixed sampling}.{\BBCQ}
\newblock
\BIn{} \APACrefbtitle {Proceedings of the IEEE/CVF Winter Conference on
  Applications of Computer Vision} {Proceedings of the ieee/cvf winter
  conference on applications of computer vision}\ (\BPGS\ 1379--1389).
\PrintBackRefs{\CurrentBib}

\bibitem [\protect \citeauthoryear {%
Tremblay%
\ \protect \BOthers {.}}{%
Tremblay%
\ \protect \BOthers {.}}{%
{\protect \APACyear {2018}}%
}]{%
tremblay2018training}
\APACinsertmetastar {%
tremblay2018training}%
\begin{APACrefauthors}%
Tremblay, J.%
, Prakash, A.%
, Acuna, D.%
, Brophy, M.%
, Jampani, V.%
, Anil, C.%
\BDBL {}Birchfield, S.%
\end{APACrefauthors}%
\unskip\
\newblock
\APACrefYearMonthDay{2018}{}{}.
\newblock
{\BBOQ}\APACrefatitle {Training deep networks with synthetic data: Bridging the
  reality gap by domain randomization} {Training deep networks with synthetic
  data: Bridging the reality gap by domain randomization}.{\BBCQ}
\newblock
\BIn{} \APACrefbtitle {Proceedings of the IEEE conference on computer vision
  and pattern recognition workshops} {Proceedings of the ieee conference on
  computer vision and pattern recognition workshops}\ (\BPGS\ 969--977).
\PrintBackRefs{\CurrentBib}

\bibitem [\protect \citeauthoryear {%
Tsai%
\ \protect \BOthers {.}}{%
Tsai%
\ \protect \BOthers {.}}{%
{\protect \APACyear {2018}}%
}]{%
AdaptSegNet}
\APACinsertmetastar {%
AdaptSegNet}%
\begin{APACrefauthors}%
Tsai, Y\BHBI H.%
, Hung, W\BHBI C.%
, Schulter, S.%
, Sohn, K.%
, Yang, M\BHBI H.%
\BCBL {}\ \BBA {} Chandraker, M.%
\end{APACrefauthors}%
\unskip\
\newblock
\APACrefYearMonthDay{2018}{}{}.
\newblock
{\BBOQ}\APACrefatitle {Learning to adapt structured output space for semantic
  segmentation} {Learning to adapt structured output space for semantic
  segmentation}.{\BBCQ}
\newblock
\BIn{} \APACrefbtitle {Proceedings of the IEEE Conference on Computer Vision
  and Pattern Recognition} {Proceedings of the ieee conference on computer
  vision and pattern recognition}\ (\BPGS\ 7472--7481).
\PrintBackRefs{\CurrentBib}

\bibitem [\protect \citeauthoryear {%
Tsai%
, Sohn%
, Schulter%
\BCBL {}\ \BBA {} Chandraker%
}{%
Tsai%
\ \protect \BOthers {.}}{%
{\protect \APACyear {2019}}%
}]{%
PatchAlign}
\APACinsertmetastar {%
PatchAlign}%
\begin{APACrefauthors}%
Tsai, Y\BHBI H.%
, Sohn, K.%
, Schulter, S.%
\BCBL {}\ \BBA {} Chandraker, M.%
\end{APACrefauthors}%
\unskip\
\newblock
\APACrefYearMonthDay{2019}{}{}.
\newblock
{\BBOQ}\APACrefatitle {Domain adaptation for structured output via
  discriminative patch representations} {Domain adaptation for structured
  output via discriminative patch representations}.{\BBCQ}
\newblock
\BIn{} \APACrefbtitle {Proceedings of the IEEE International Conference on
  Computer Vision} {Proceedings of the ieee international conference on
  computer vision}\ (\BPGS\ 1456--1465).
\PrintBackRefs{\CurrentBib}

\bibitem [\protect \citeauthoryear {%
Vergara%
\ \protect \BOthers {.}}{%
Vergara%
\ \protect \BOthers {.}}{%
{\protect \APACyear {2012}}%
}]{%
vergara2012chemical}
\APACinsertmetastar {%
vergara2012chemical}%
\begin{APACrefauthors}%
Vergara, A.%
, Vembu, S.%
, Ayhan, T.%
, Ryan, M\BPBI A.%
, Homer, M\BPBI L.%
\BCBL {}\ \BBA {} Huerta, R.%
\end{APACrefauthors}%
\unskip\
\newblock
\APACrefYearMonthDay{2012}{}{}.
\newblock
{\BBOQ}\APACrefatitle {Chemical gas sensor drift compensation using classifier
  ensembles} {Chemical gas sensor drift compensation using classifier
  ensembles}.{\BBCQ}
\newblock
\APACjournalVolNumPages{Sensors and Actuators B: Chemical}{166}{}{320--329}.
\PrintBackRefs{\CurrentBib}

\bibitem [\protect \citeauthoryear {%
Vu%
, Jain%
, Bucher%
, Cord%
\BCBL {}\ \BBA {} P{\'e}rez%
}{%
Vu%
\ \protect \BOthers {.}}{%
{\protect \APACyear {2019}}%
}]{%
ADVENT}
\APACinsertmetastar {%
ADVENT}%
\begin{APACrefauthors}%
Vu, T\BHBI H.%
, Jain, H.%
, Bucher, M.%
, Cord, M.%
\BCBL {}\ \BBA {} P{\'e}rez, P.%
\end{APACrefauthors}%
\unskip\
\newblock
\APACrefYearMonthDay{2019}{}{}.
\newblock
{\BBOQ}\APACrefatitle {Advent: Adversarial entropy minimization for domain
  adaptation in semantic segmentation} {Advent: Adversarial entropy
  minimization for domain adaptation in semantic segmentation}.{\BBCQ}
\newblock
\BIn{} \APACrefbtitle {Proceedings of the IEEE conference on computer vision
  and pattern recognition} {Proceedings of the ieee conference on computer
  vision and pattern recognition}\ (\BPGS\ 2517--2526).
\PrintBackRefs{\CurrentBib}

\bibitem [\protect \citeauthoryear {%
D.~Wang%
, Shelhamer%
, Liu%
, Olshausen%
\BCBL {}\ \BBA {} Darrell%
}{%
D.~Wang%
\ \protect \BOthers {.}}{%
{\protect \APACyear {2020}}%
}]{%
TENT}
\APACinsertmetastar {%
TENT}%
\begin{APACrefauthors}%
Wang, D.%
, Shelhamer, E.%
, Liu, S.%
, Olshausen, B.%
\BCBL {}\ \BBA {} Darrell, T.%
\end{APACrefauthors}%
\unskip\
\newblock
\APACrefYearMonthDay{2020}{}{}.
\newblock
{\BBOQ}\APACrefatitle {Tent: Fully test-time adaptation by entropy
  minimization} {Tent: Fully test-time adaptation by entropy
  minimization}.{\BBCQ}
\newblock
\APACjournalVolNumPages{arXiv preprint arXiv:2006.10726}{}{}{}.
\PrintBackRefs{\CurrentBib}

\bibitem [\protect \citeauthoryear {%
Q.~Wang%
, Fink%
, Van~Gool%
\BCBL {}\ \BBA {} Dai%
}{%
Q.~Wang%
\ \protect \BOthers {.}}{%
{\protect \APACyear {2022}}%
}]{%
wang2022continual}
\APACinsertmetastar {%
wang2022continual}%
\begin{APACrefauthors}%
Wang, Q.%
, Fink, O.%
, Van~Gool, L.%
\BCBL {}\ \BBA {} Dai, D.%
\end{APACrefauthors}%
\unskip\
\newblock
\APACrefYearMonthDay{2022}{}{}.
\newblock
{\BBOQ}\APACrefatitle {Continual test-time domain adaptation} {Continual
  test-time domain adaptation}.{\BBCQ}
\newblock
\BIn{} \APACrefbtitle {Proceedings of the IEEE/CVF Conference on Computer
  Vision and Pattern Recognition} {Proceedings of the ieee/cvf conference on
  computer vision and pattern recognition}\ (\BPGS\ 7201--7211).
\PrintBackRefs{\CurrentBib}

\bibitem [\protect \citeauthoryear {%
X.~Wu%
, Wu%
, Lu%
, Ju%
\BCBL {}\ \BBA {} Wang%
}{%
X.~Wu%
\ \protect \BOthers {.}}{%
{\protect \APACyear {2021}}%
}]{%
SM-PPM}
\APACinsertmetastar {%
SM-PPM}%
\begin{APACrefauthors}%
Wu, X.%
, Wu, Z.%
, Lu, Y.%
, Ju, L.%
\BCBL {}\ \BBA {} Wang, S.%
\end{APACrefauthors}%
\unskip\
\newblock
\APACrefYearMonthDay{2021}{}{}.
\newblock
{\BBOQ}\APACrefatitle {Style Mixing and Patchwise Prototypical Matching for
  One-Shot Unsupervised Domain Adaptive Semantic Segmentation} {Style mixing
  and patchwise prototypical matching for one-shot unsupervised domain adaptive
  semantic segmentation}.{\BBCQ}
\newblock
\APACjournalVolNumPages{arXiv preprint arXiv:2112.04665}{}{}{}.
\PrintBackRefs{\CurrentBib}

\bibitem [\protect \citeauthoryear {%
Z.~Wu%
, Wang%
, Gonzalez%
, Goldstein%
\BCBL {}\ \BBA {} Davis%
}{%
Z.~Wu%
\ \protect \BOthers {.}}{%
{\protect \APACyear {2019}}%
}]{%
ACE}
\APACinsertmetastar {%
ACE}%
\begin{APACrefauthors}%
Wu, Z.%
, Wang, X.%
, Gonzalez, J\BPBI E.%
, Goldstein, T.%
\BCBL {}\ \BBA {} Davis, L\BPBI S.%
\end{APACrefauthors}%
\unskip\
\newblock
\APACrefYearMonthDay{2019}{}{}.
\newblock
{\BBOQ}\APACrefatitle {ACE: adapting to changing environments for semantic
  segmentation} {Ace: adapting to changing environments for semantic
  segmentation}.{\BBCQ}
\newblock
\BIn{} \APACrefbtitle {Proceedings of the IEEE International Conference on
  Computer Vision} {Proceedings of the ieee international conference on
  computer vision}\ (\BPGS\ 2121--2130).
\PrintBackRefs{\CurrentBib}

\bibitem [\protect \citeauthoryear {%
Wulfmeier%
, Bewley%
\BCBL {}\ \BBA {} Posner%
}{%
Wulfmeier%
\ \protect \BOthers {.}}{%
{\protect \APACyear {2018}}%
}]{%
wulfmeier}
\APACinsertmetastar {%
wulfmeier}%
\begin{APACrefauthors}%
Wulfmeier, M.%
, Bewley, A.%
\BCBL {}\ \BBA {} Posner, I.%
\end{APACrefauthors}%
\unskip\
\newblock
\APACrefYearMonthDay{2018}{}{}.
\newblock
{\BBOQ}\APACrefatitle {Incremental adversarial domain adaptation for
  continually changing environments} {Incremental adversarial domain adaptation
  for continually changing environments}.{\BBCQ}
\newblock
\BIn{} \APACrefbtitle {2018 IEEE International conference on robotics and
  automation (ICRA)} {2018 ieee international conference on robotics and
  automation (icra)}\ (\BPGS\ 4489--4495).
\PrintBackRefs{\CurrentBib}

\bibitem [\protect \citeauthoryear {%
Yang%
\ \BBA {} Soatto%
}{%
Yang%
\ \BBA {} Soatto%
}{%
{\protect \APACyear {2020}}%
}]{%
FDA}
\APACinsertmetastar {%
FDA}%
\begin{APACrefauthors}%
Yang, Y.%
\BCBT {}\ \BBA {} Soatto, S.%
\end{APACrefauthors}%
\unskip\
\newblock
\APACrefYearMonthDay{2020}{}{}.
\newblock
{\BBOQ}\APACrefatitle {Fda: Fourier domain adaptation for semantic
  segmentation} {Fda: Fourier domain adaptation for semantic
  segmentation}.{\BBCQ}
\newblock
\BIn{} \APACrefbtitle {Proceedings of the IEEE/CVF Conference on Computer
  Vision and Pattern Recognition} {Proceedings of the ieee/cvf conference on
  computer vision and pattern recognition}\ (\BPGS\ 4085--4095).
\PrintBackRefs{\CurrentBib}

\bibitem [\protect \citeauthoryear {%
Yu%
\ \protect \BOthers {.}}{%
Yu%
\ \protect \BOthers {.}}{%
{\protect \APACyear {2020}}%
}]{%
yu2020bdd100k}
\APACinsertmetastar {%
yu2020bdd100k}%
\begin{APACrefauthors}%
Yu, F.%
, Chen, H.%
, Wang, X.%
, Xian, W.%
, Chen, Y.%
, Liu, F.%
\BDBL {}Darrell, T.%
\end{APACrefauthors}%
\unskip\
\newblock
\APACrefYearMonthDay{2020}{}{}.
\newblock
{\BBOQ}\APACrefatitle {Bdd100k: A diverse driving dataset for heterogeneous
  multitask learning} {Bdd100k: A diverse driving dataset for heterogeneous
  multitask learning}.{\BBCQ}
\newblock
\BIn{} \APACrefbtitle {Proceedings of the IEEE/CVF conference on computer
  vision and pattern recognition} {Proceedings of the ieee/cvf conference on
  computer vision and pattern recognition}\ (\BPGS\ 2636--2645).
\PrintBackRefs{\CurrentBib}

\bibitem [\protect \citeauthoryear {%
H.~Zhang%
, Cisse%
, Dauphin%
\BCBL {}\ \BBA {} Lopez-Paz%
}{%
H.~Zhang%
\ \protect \BOthers {.}}{%
{\protect \APACyear {2017}}%
}]{%
zhang2017mixup}
\APACinsertmetastar {%
zhang2017mixup}%
\begin{APACrefauthors}%
Zhang, H.%
, Cisse, M.%
, Dauphin, Y\BPBI N.%
\BCBL {}\ \BBA {} Lopez-Paz, D.%
\end{APACrefauthors}%
\unskip\
\newblock
\APACrefYearMonthDay{2017}{}{}.
\newblock
{\BBOQ}\APACrefatitle {mixup: Beyond empirical risk minimization} {mixup:
  Beyond empirical risk minimization}.{\BBCQ}
\newblock
\APACjournalVolNumPages{arXiv preprint arXiv:1710.09412}{}{}{}.
\PrintBackRefs{\CurrentBib}

\bibitem [\protect \citeauthoryear {%
M.~Zhang%
, Levine%
\BCBL {}\ \BBA {} Finn%
}{%
M.~Zhang%
\ \protect \BOthers {.}}{%
{\protect \APACyear {2021}}%
}]{%
MEMO}
\APACinsertmetastar {%
MEMO}%
\begin{APACrefauthors}%
Zhang, M.%
, Levine, S.%
\BCBL {}\ \BBA {} Finn, C.%
\end{APACrefauthors}%
\unskip\
\newblock
\APACrefYearMonthDay{2021}{}{}.
\newblock
{\BBOQ}\APACrefatitle {MEMO: Test Time Robustness via Adaptation and
  Augmentation} {Memo: Test time robustness via adaptation and
  augmentation}.{\BBCQ}
\newblock
\APACjournalVolNumPages{arXiv preprint arXiv:2110.09506}{}{}{}.
\PrintBackRefs{\CurrentBib}

\bibitem [\protect \citeauthoryear {%
P.~Zhang%
\ \protect \BOthers {.}}{%
P.~Zhang%
\ \protect \BOthers {.}}{%
{\protect \APACyear {2021}}%
}]{%
ProDA}
\APACinsertmetastar {%
ProDA}%
\begin{APACrefauthors}%
Zhang, P.%
, Zhang, B.%
, Zhang, T.%
, Chen, D.%
, Wang, Y.%
\BCBL {}\ \BBA {} Wen, F.%
\end{APACrefauthors}%
\unskip\
\newblock
\APACrefYearMonthDay{2021}{}{}.
\newblock
{\BBOQ}\APACrefatitle {Prototypical pseudo label denoising and target structure
  learning for domain adaptive semantic segmentation} {Prototypical pseudo
  label denoising and target structure learning for domain adaptive semantic
  segmentation}.{\BBCQ}
\newblock
\BIn{} \APACrefbtitle {Proceedings of the IEEE/CVF Conference on Computer
  Vision and Pattern Recognition} {Proceedings of the ieee/cvf conference on
  computer vision and pattern recognition}\ (\BPGS\ 12414--12424).
\PrintBackRefs{\CurrentBib}

\bibitem [\protect \citeauthoryear {%
Q.~Zhang%
, Zhang%
, Liu%
\BCBL {}\ \BBA {} Tao%
}{%
Q.~Zhang%
\ \protect \BOthers {.}}{%
{\protect \APACyear {2019}}%
}]{%
CAG}
\APACinsertmetastar {%
CAG}%
\begin{APACrefauthors}%
Zhang, Q.%
, Zhang, J.%
, Liu, W.%
\BCBL {}\ \BBA {} Tao, D.%
\end{APACrefauthors}%
\unskip\
\newblock
\APACrefYearMonthDay{2019}{}{}.
\newblock
{\BBOQ}\APACrefatitle {Category anchor-guided unsupervised domain adaptation
  for semantic segmentation} {Category anchor-guided unsupervised domain
  adaptation for semantic segmentation}.{\BBCQ}
\newblock
\BIn{} \APACrefbtitle {Advances in Neural Information Processing Systems}
  {Advances in neural information processing systems}\ (\BPGS\ 435--445).
\PrintBackRefs{\CurrentBib}

\bibitem [\protect \citeauthoryear {%
Y.~Zhang%
, Deng%
, Jia%
\BCBL {}\ \BBA {} Zhang%
}{%
Y.~Zhang%
\ \protect \BOthers {.}}{%
{\protect \APACyear {2021}}%
}]{%
AuxSelfTrain}
\APACinsertmetastar {%
AuxSelfTrain}%
\begin{APACrefauthors}%
Zhang, Y.%
, Deng, B.%
, Jia, K.%
\BCBL {}\ \BBA {} Zhang, L.%
\end{APACrefauthors}%
\unskip\
\newblock
\APACrefYearMonthDay{2021}{}{}.
\newblock
{\BBOQ}\APACrefatitle {Gradual Domain Adaptation via Self-Training of Auxiliary
  Models} {Gradual domain adaptation via self-training of auxiliary
  models}.{\BBCQ}
\newblock
\APACjournalVolNumPages{arXiv preprint arXiv:2106.09890}{}{}{}.
\PrintBackRefs{\CurrentBib}

\bibitem [\protect \citeauthoryear {%
Zhu%
, Park%
, Isola%
\BCBL {}\ \BBA {} Efros%
}{%
Zhu%
\ \protect \BOthers {.}}{%
{\protect \APACyear {2017}}%
}]{%
CycleGAN}
\APACinsertmetastar {%
CycleGAN}%
\begin{APACrefauthors}%
Zhu, J\BHBI Y.%
, Park, T.%
, Isola, P.%
\BCBL {}\ \BBA {} Efros, A\BPBI A.%
\end{APACrefauthors}%
\unskip\
\newblock
\APACrefYearMonthDay{2017}{}{}.
\newblock
{\BBOQ}\APACrefatitle {Unpaired image-to-image translation using
  cycle-consistent adversarial networks} {Unpaired image-to-image translation
  using cycle-consistent adversarial networks}.{\BBCQ}
\newblock
\BIn{} \APACrefbtitle {Proceedings of the IEEE international conference on
  computer vision} {Proceedings of the ieee international conference on
  computer vision}\ (\BPGS\ 2223--2232).
\PrintBackRefs{\CurrentBib}

\end{thebibliography}

\appendix

\onecolumn

\aistatstitle{Supplementary Materials}

\section{Ablation studies for the classification benchmarks}

\subsection{Component analysis}
\label{appendix:component_analysis}

\paragraph{Influence of individual components} We begin our ablation studies by examining each component on the continual TTA benchmarks. The results are summarized in Table \ref{tab:ablation_component}. First, we investigate the effect of solely performing self-training and the advantages of filtering pseudo-labels. Compared to the BN--1 performance, self-training alone only improves the performance on CIFAR10C and CIFAR100C. For ImageNet-C and ImageNet-R, we experience error accumulation resulting in a drastic performance degredation on ImageNet-C. Filtering pseudo-labels according to our proposed threshold benefits all continual classification benchmarks surpassing the BN--1 performance. This is expected since less confident predictions are more likely to be incorrect and can significantly contribute to error accumulation due to large gradient magnitudes \cite{mummadi2021test}. Utilizing source data in the form of source replay stabilizes online test-time adaptation in the sense that the performance is improved on all investigated datasets. We also analyze source replay for TENT \cite{TENT} and find that it is also beneficial for methods based on entropy minimization. TENT-continual with source replay achieves an error rate of 18.1\%, 31.2\%, 61.3\%, and 57.6\% on CIFAR10C, CIFAR100C, ImageNet-C, and ImageNet-R, respectively. This corresponds to an error reduction of -2.6\%, -29.7\%, -1.3\%, and -1.0\%. Especially for CIFAR100C, this is a significant improvement, as TENT-continual without source replay suffers from heavy error accumulation. Last but not least we look into the effect of omitting self-training and solely doing mixup or style transfer. While training on the intermediate domains provided either by mixup or style transfer significantly improves the performance upon BN--1, both methods are limited in mitigating the domain gap. Adding self-training to the framework further benefits the results. This highlights that self-training can have a big performance improvement when provided with reliable pseudo-labels.

\paragraph{Performing multiple update steps}
As already shown in Table \ref{tab:continual-corruptions}, GTTA-MIX and GTTA-ST benefit from doing multiple update steps. In general, performing multiple updates does not necessarily lead to a performance improvement. Assuming that a proper learning rate for a single update step was chosen, performing, e.g., four updates does not improve the performance when only doing self-training, as presented in Table \ref{tab:ablation_component}. This especially becomes apparent for ImageNet-C and ImageNet-R, where a drastic performance degradation is experienced. This is not surprising as in the setting of online test-time adaptation only the current test batch $x_t^T$ is commonly utilized to perform adaptation. We denote this as over-adaptation, where as a result error accumulation can be very prominent. Naturally, filtering less confident predictions, which are more likely to be false predictions, reduces error accumulation. Source replay further helps in the setting of multiple updates to stabilize the adaptation and even results in a performance improvement on CIFAR10C and CIFAR100C compared to a single update. Finally, generating intermediate domains in the form of either mixup or style transfer is the key factor to benefit from multiple updates on CIFAR10C, CIFAR100C, and ImageNet-C. Since mixup has its limitations to create intermediate domains for natural shifts, as imposed by ImageNet-R, only style transfer benefits from multiple updates on this dataset.

\begin{table}[t]
    \newcolumntype{C}[1]{>{\centering\let\newline\\\arraybackslash\hspace{0pt}}m{#1}}
    \renewcommand{\arraystretch}{1.2}
    \centering
    \setlength{\tabcolsep}{3pt}
    \caption{Component analysis for the continual classification benchmarks. The classification error rate~(\%) is reported as an average over 5 runs. We investigate the effect of self-training, filtering pseudo-labels, source-replay, and performing mixup and style transfer without self-training (on the right).}
    \scalebox{1.0}{
        \begin{tabular}{l|C{1cm}|C{1cm}|C{1cm}|C{1cm}|C{1cm}|C{1cm}|C{1cm}|C{1cm}|C{1cm}||C{1.1cm}|C{1.1cm}}\hline
        & BN--1 & \multicolumn{2}{c|}{self-training} & \multicolumn{2}{c|}{self-training} & \multicolumn{2}{c|}{+ source replay} & \multicolumn{2}{c||}{+ mixup} & mixup & style transfer \\
        \hline
        self-training & n/a & \multicolumn{2}{c|}{\ding{51}} & \multicolumn{2}{c|}{\ding{51}} & \multicolumn{2}{c|}{\ding{51}} & \multicolumn{2}{c||}{\ding{51}} & \ding{55} & \ding{55} \\
        \hline
        threshold & n/a &  \multicolumn{2}{c|}{\ding{55}} & \multicolumn{2}{c|}{\ding{51}} & \multicolumn{2}{c|}{\ding{51}} & \multicolumn{2}{c||}{\ding{51}} & n/a & n/a \\
        \hline
        \diagbox{dataset}{updates} & - & 1 & 4 & 1 & 4 & 1 & 4 & 1 & 4 & 1 & 1  \\
        \hline
        CIFAR10C  & 20.4 & 19.9 & 24.8 & 18.4 & 19.4 & 18.1 & 17.6 & 17.2 & 15.6 & 17.7 & - \\
        CIFAR100C & 35.4 & 33.5 & 40.6 & 32.0 & 33.4 & 30.5 & 29.1 & 30.4 & 28.9 & 31.3 & - \\
        ImageNet-C & 68.6 & 81.9 & 97.7 & 66.6 & 94.1 & 65.2 & 88.9 & 60.3 & 57.1 & 63.1 & 62.2 \\
        ImageNet-R & 60.4 & 61.2 & 79.6 & 55.7 & 67.5 & 55.4 & 56.0 & 56.4 & 56.6 & 59.7 & 56.9 \\
        \hline
        \end{tabular}}
        \label{tab:ablation_component}
\end{table}

\subsection{Trade-off between efficiency and performance}
In Table \ref{tab:ablations_combined} (a) we further explore the effect of performing different numbers of update steps for our overall approaches GTTA-MIX and GTTA-ST, since for some applications it can make sense to neglect computational efficiency in favor of a higher accuracy and vice versa. For GTTA-MIX, CIFAR10C and CIFAR100C benefit increasingly from performing multiple update steps. They show the best performance at 8 update steps reducing the error rate to 15.0\% and 28.3\%, respectively. For ImageNet-C the best performance is achieved at 6 update steps reducing the error rate from 60.3\% for one update to 56.4\%. Since mixup shows its difficulties for natural domain shifts, as covered by ImageNet-R, performing more update steps does not necessarily result in a better performance. For GTTA-ST, the best performance for ImageNet-C is again achieved at 6 update steps, resulting in an error rate of 56.3\%. In contrast to mixup, style transfer can benefit from multiple update steps for ImageNet-R, achieving the best performance of 52.3\% performing 8 update steps.

\subsection{Amount of source  samples}
Since applications can vary in the amount of available memory, we show in Table \ref{tab:ablations_combined} (b) the error rate for different percentages of saved source samples. While the performance on ImageNet-C and ImageNet-R is only marginally affected by the amount of available source samples for both GTTA-MIX and GTTA-ST, the error rate increases slightly for CIFAR10C and CIFAR100C. Still, using only 10\% of the source data does not increase the error rate significantly for both CIFAR10C and CIFAR100C.

\begin{table*}[t]
    \renewcommand{\arraystretch}{1.2}
    \centering
    \setlength{\tabcolsep}{3pt}
    \caption{Classification error rate~(\%) for different: (a) numbers of update steps; (b) amounts of saved source samples.}
    \begin{tabular}{cc}
        (a) & (b) \\
        \scalebox{1.0}{
            \tabcolsep3pt
            \begin{tabular}{l|l|ccccc}\hline
             & \diagbox{dataset}{updates} & 1 & 2 & 4 & 6 & 8 \\
            \hline
            \multirow{4}{*}{GTTA-MIX} &
            CIFAR10C    & 17.2 & 16.5 & 15.6 & 15.2 & \textbf{15.0} \\
             & CIFAR100C   & 30.4 & 29.7 & 28.9 & 28.6 & \textbf{28.3} \\
             & ImageNet-C   & 60.3 & 58.4 & 57.1 & \textbf{56.4} & 56.6 \\
             & ImageNet-R   & 56.4 & \textbf{55.9} & 56.6 & 56.1 & 57.3 \\
            \hline
            \multirow{2}{*}{GTTA-ST}
             & ImageNet-C   & 59.7 & 58.0 & 57.0 & \textbf{56.3} & 57.0 \\
             & ImageNet-R   & 53.8 & 52.9 & 52.5 & 52.9 & \textbf{52.3} \\
            \hline
            \end{tabular}}
        &
        \scalebox{1.0}{
            \tabcolsep3pt
            \begin{tabular}{l|cccccc}\hline
            \diagbox{dataset}{\% source} & 100 & 50 & 25 & 10 & 5 & 1 \\
            \hline
            CIFAR10C  & \textbf{17.2} & \textbf{17.2} & \textbf{17.2} & 17.4 & 17.8 & 18.8 \\
            CIFAR100C & \textbf{30.4} & \textbf{30.4} & 30.5 & 30.6 & 30.9 & 32.0 \\
            ImageNet-C & 60.3 & \textbf{60.0} & 60.4 & 60.3 & 60.3 & 60.5 \\
            ImageNet-R & 56.4 & 56.4 & 56.1 & 56.3 & \textbf{55.9} & 56.5 \\
            \hline
            ImageNet-C   & 59.7 & 59.6 & 59.7 & 59.7 & \textbf{59.5} & 59.8 \\
            ImageNet-R   & \textbf{53.8} & 54.1 & 54.0 & 54.1 & 54.5 & 54.0 \\
            \hline
            \end{tabular}}
        \\
    \end{tabular}
    \label{tab:ablations_combined}
\end{table*}

\subsection{Mixup strength}
\label{appendix:mixup_strength}
In Table \ref{tab:ablation_mixup_strength} we investigate the effect of mixup, in particular, how much source or test image is taken into account to create intermediate samples. First, it can be seen that mixup is beneficial for all datasets. ImageNet-R plays a special role in the sense that only a small mixup strength $\lambda_\text{mix}=0.1$ marginally improves upon source replay, which corresponds to a mixup strength of $\lambda_\text{mix}=0$. For higher mixup strengths, the performance drastically decreases. For CIFAR10C and CIFAR100C $\lambda_\text{mix}=1/3$ is a good choice, corresponding to weighting source images twice as strong as test images. ImageNet-C further benefits from higher mixup strength and shows its best performance at $\lambda_\text{mix}=0.5$, corresponding to an equal weighting of source and test images.

\begin{table}[t]
    \renewcommand{\arraystretch}{1.2}
    \centering
    \setlength{\tabcolsep}{3pt}
    \caption{Classification error rate~(\%) averaged over 5 runs for different mixup strengths $\lambda_\text{mix}$. While $\lambda_\text{mix}=0$ corresponds to source replay, $\lambda_\text{mix}=0.5$ corresponds to the case of equally weighting source and test images.}
    \scalebox{1.0}{
        \tabcolsep3pt
        \begin{tabular}{l|cccccc}\hline
        \diagbox{dataset}{$\lambda_{\mathrm{mix}}$} & 0 & 1/10 & 1/4 & 1/3 & 2/5 & 1/2 \\
        \hline
        CIFAR10C & 18.1 & 18.1 & 17.7 & \textbf{17.2} & 17.4 & 17.7 \\
        CIFAR100C & 30.5 & \textbf{30.4} & \textbf{30.4} & \textbf{30.4} & 30.5 & 30.6 \\
        ImageNet-C & 65.2 & 64.0 & 61.5 & 60.3 & 59.9 & \textbf{59.6} \\
        ImageNet-R & 55.4 & \textbf{55.2} & 55.8 & 56.4 & 57.1 & 58.0 \\
        \hline
        \end{tabular}}
    \label{tab:ablation_mixup_strength}
\end{table}

\subsection{Single sample test-time adaptation}
\label{appendix:single_sample}
\begin{table*}[t]
\renewcommand{\arraystretch}{1.2}
\centering
\caption{Classification error rate~(\%) for batch size 1 using a sliding window approach with $b$ samples. The weight update is performed once after the complete buffer has changed.} 
\scalebox{0.95}{
\label{tab:ablation_sliding_window}
\tabcolsep3pt
\begin{tabular}{l|l|c|c|  ccccccccccccccc|c}\hline
& \multicolumn{3}{l|}{Time} & \multicolumn{15}{l|}{$t\xrightarrow{\hspace*{12.2cm}}$}& \\ \hline
& Method & \rotatebox[origin=c]{90}{ batch size } & \rotatebox[origin=c]{90}{buffer size} & \rotatebox[origin=c]{70}{Gaussian} & \rotatebox[origin=c]{70}{shot} & \rotatebox[origin=c]{70}{impulse} & \rotatebox[origin=c]{70}{defocus} & \rotatebox[origin=c]{70}{glass} & \rotatebox[origin=c]{70}{motion} & \rotatebox[origin=c]{70}{zoom} & \rotatebox[origin=c]{70}{snow} & \rotatebox[origin=c]{70}{frost} & \rotatebox[origin=c]{70}{fog}  & \rotatebox[origin=c]{70}{brightness} & \rotatebox[origin=c]{70}{contrast} & \rotatebox[origin=c]{70}{elastic\_trans} & \rotatebox[origin=c]{70}{pixelate} & \rotatebox[origin=c]{70}{jpeg} & Mean \\
\hline
\multirow{6}{*}{\rotatebox[origin=c]{90}{CIFAR10C}}& BN--0 (src.) & 1 & - & 72.3 & 65.7 & 72.9 & 46.9 & 54.3 & 34.8 & 42.0 & 25.1 & 41.3 & 26.0 & 9.3 & 46.7 & 26.6 & 58.5 & 30.3 & 43.5 \\
& BN--1     & 1 & 16 & 30.8 & 28.8 & 39.2 & 15.7 & 37.5 & 16.8 & 15.0 & 20.3 & 20.2 & 17.8 & 10.7 & 15.5 & 27.1 & 23.0 & 29.9 & 23.2 \\
& BN--1     & 1 & 32 & 29.3 & 27.3 & 37.7 & 14.2 & 36.5 & 15.4 & 13.8 & 19.0 & 18.8 & 16.8 & 9.4 & 14.3 & 25.2 & 21.4 & 28.8 & 21.9 \\
& GTTA-MIX & 1 & 16 & 26.5 & 21.5 & 28.9 & 12.7 & 30.5 & 13.9 & 12.9 & 16.8 & 15.0 & 13.1 & 9.2 & 10.6 & 23.0 & 17.5 & 22.9 & 18.3 \\
& GTTA-MIX & 1 & 32 & 25.5 & 21.4 & 28.2 & 11.9 & 30.1 & 13.2 & 11.9 & 15.8 & 14.8 & 12.8 & 8.6 & 10.6 & 21.4 & 16.5 & 23.0 & 17.7 \\
& GTTA-MIX & 200 & - & 26.0 & 21.5 & 29.7 & 11.1 & 30.0 & 12.2 & 10.5 & 15.1 & 14.1 & 12.3 & 7.5 & 10.0 & 20.4 & 15.8 & 21.4 & 17.2\\
\hline
\multirow{6}{*}{\rotatebox[origin=c]{90}{CIFAR100C}}& BN--0 (src.) & 1 & - & 73.0 & 68.0 & 39.4 & 29.3 & 54.1 & 30.8 & 28.8 & 39.5 & 45.8 & 50.3 & 29.5 & 55.1 & 37.2 & 74.7 & 41.2 & 46.4 \\
& BN--1     & 1 & 16 & 46.2 & 44.5 & 47.1 & 31.4 & 46.5 & 33.3 & 31.7 & 39.2 & 38.6 & 45.4 & 30.8 & 34.7 & 40.6 & 37.5 & 45.2 & 39.5\\
& BN--1     & 1 & 32 & 44.1 & 42.4 & 45.2 & 29.6 & 43.6 & 31.0 & 30.0 & 37.1 & 36.7 & 43.8 & 28.7 & 32.6 & 38.0 & 35.3 & 42.9 & 37.4 \\
& GTTA-MIX & 1 & 16 & 40.1 & 36.1 & 37.8 & 28.3 & 40.4 & 29.8 & 28.3 & 32.7 & 31.1 & 34.2 & 26.2 & 27.5 & 33.5 & 30.4 & 38.2 & 33.0 \\
& GTTA-MIX & 1 & 32 & 38.9 & 35.0 & 36.8 & 26.6 & 38.3 & 28.0 & 26.5 & 31.2 & 30.1 & 34.0 & 24.4 & 26.2 & 32.1 & 28.6 & 36.4 & 31.5 \\
& GTTA-MIX & 200 & - & 39.4 & 34.4 & 36.6 & 24.7 & 36.8 & 26.6 & 24.3 & 30.1 & 28.9 & 34.6 & 22.8 & 25.1 & 30.7 & 26.9
& 34.7 & 30.4\\
\hline
\multirow{9}{*}{\rotatebox[origin=c]{90}{ImageNet-C}} & BN--0 (src.) & 1 & -  & 95.3 & 94.6 & 95.3 & 84.9 & 91.1 & 86.9 & 77.2 & 84.4 & 79.7 & 77.3 & 44.4 & 95.6  & 85.2 & 76.9 & 66.7 & 82.4 \\
 & BN--1     & 1 & 16 & 86.6 & 85.2 & 86.0 & 86.6 & 86.5 & 75.7 & 65.1 & 67.8 & 70.5 & 55.2 & 37.8 & 84.6 & 59.1 & 55.0 & 63.7 & 71.0 \\
 & BN--1     & 1 & 32 & 85.4 & 84.0 & 84.9 & 85.6 & 85.0 & 74.1 & 61.9 & 66.8 & 68.4 & 52.6 & 36.1 & 83.9 & 57.1 & 52.3 & 61.7 & 69.3 \\
 & GTTA-MIX & 1 & 16 & 80.8 & 73.8 & 71.6 & 79.3 & 79.3 & 70.0 & 61.1 & 58.1 & 59.8 & 47.6 & 37.5 & 73.8 & 54.2 & 51.1 & 53.3 & 63.4\\
 & GTTA-MIX & 1 & 32 & 79.8 & 73.4 & 70.7 & 77.3 & 76.0 & 64.6 & 55.8 & 57.4 & 58.5 & 45.5 & 34.4 & 67.1 & 50.5 & 45.7 & 49.4 & 60.4 \\
 & GTTA-MIX & 64 & - & 80.5 & 74.7 & 72.4 & 77.8 & 75.7 & 64.3 & 54.0 & 57.0 & 58.6 & 44.6 & 33.9 & 67.5 & 49.4 & 44.7 & 49.3 & 60.3 \\
 & GTTA-ST & 1 & 16 & 80.9 & 74.3 & 74.7 & 77.7 & 77.3 & 66.3 & 58.2 & 59.0 & 60.4 & 47.6 & 37.6 & 62.9 & 53.5 & 48.8 & 52.6 & 62.1 \\
 & GTTA-ST & 1 & 32 & 80.5 & 74.4 & 74.0 & 77.3 & 74.8 & 63.5 & 54.3 & 56.4 & 57.6 & 45.2 & 33.5 & 61.8 & 50.0 & 45.7 & 50.7 & 60.0 \\
 & GTTA-ST & 64 & - & 80.6 & 74.1 & 74.3 & 76.8 & 74.9 & 62.3 & 53.9 & 56.4 & 58.0 & 44.1 & 33.4 & 62.2 & 48.6 & 44.9 & 50.4 & 59.7 \\
\hline
\end{tabular}}
\end{table*}

So far, we only considered the batch setting for the classification task. For some applications, timeliness can be critical, e.g., in autonomous driving, and production data arrives individually rather than in batches. Simply performing prediction and adaptation for a single sample for classification imposes some challenges. First, computing a good estimate of the batch normalization statistics is impossible and second, gradient updates are very noisy. One straightforward approach to overcome these challenges is to save recent data in a buffer and use a sliding window for performing prediction and adaptation. Specifically, we save the last $b$ test samples in a buffer and update the model every $b$ samples due to the strong correlation introduced by the buffer. For current test sample $x_{ti}^T$ at time step $t$, the sample is first added to the buffer replacing the oldest sample. Now, the whole buffer is forwarded to make a prediction for the current sample $x_{ti}^T$. Needless to say, this comes with a computational overhead, but allows a good estimate of the batch normalization statistics and better gradient updates.

Table \ref{tab:ablation_sliding_window} illustrates the results for the previously described setting on the continual corruption benchmarks using a buffer of size 16. Due to the much smaller batch size used in this setting, the performance of the baseline BN--1 slightly degrades, since the estimation of the batch statistics becomes more noisy. While the performance of our approach is also slightly worse compared to the results achieved in the batch setting of TTA, GTTA in the single-sample setting still performs on par with most state-of-the-art methods in the batch setting. Using a larger buffer size of 32 further increases the performance. On ImageNet-C the performance in the single sample setting is now comparable with the batch setting, being only 0.1\% and 0.3\% behind for GTTA-MIX and GTTA-ST, respectively.

\begin{table}[h]
\renewcommand{\arraystretch}{1.1}
\centering
\caption{Investigation of the amount of gradual shift. $\Delta t$ corresponds to the delta time used for all our standard sequences. We report the mIoU (\%) for the least common multiple, averaged over 5 runs. For comparison, BN--0 and BN--1 achieve 58.3 and 61.7 on the \textit{day2night} split and 39.4 and 49.1 on the \textit{dynamic} split, respectively.}
\scalebox{1.0}{\tabcolsep5pt
\label{tab:ablation_delta}
\begin{tabular}{l|ccccc}
    \hline
    & $4\Delta t$ & $2\Delta t$ & $\Delta t$ & $\Delta t/2$ & $\Delta t/4$  \\
    \hline
    day2night & 64.6 & 66.1 & 66.7 & 67.2 & \textbf{67.7} \\
    dynamic & 53.0 & 55.2 & 55.6 & \textbf{57.0} & 56.3  \\
\hline
\end{tabular}}
\end{table}
\begin{table*}[h]
    \renewcommand{\arraystretch}{1.1}
    \centering
    \caption{Mean intersection-over-union averaged over 5 runs for a) when no domain shift occurs and the test domain is equal to the source domain; b) the domain changes abruptly from day to night.}
    \begin{tabular}{cc}
        (a) & (b) \\
        \scalebox{0.89}{\tabcolsep5pt
        \begin{tabular}{l|ccccc}
        \hline
        Method & BN--0 & BN--0.1 & BN--1 & BN--EMA & GTTA-ST  \\
        \hline
        & \textbf{78.4} & \textbf{78.4} & 76.6 & 77.6 & 77.9$\pm$0.14 \\
        \hline
        \end{tabular}}
        &
        \scalebox{0.89}{\tabcolsep5pt
        \begin{tabular}{l|ccccc}
        \hline
        Method & BN--0 & BN--0.1 & BN--1 & BN--EMA & GTTA-ST  \\
        \hline
        & 43.8 & 52.3 & 52.8 & 54.3 & \textbf{59.7}$\pm$0.32\\
        \hline
        \end{tabular}}
        \\
         &  \\
    \end{tabular}
    \label{tab:ablation_carla}
\end{table*}
\newpage
\section{Ablation studies for CarlaTTA}
\label{appendix:ablation_carla}

\subsection{Gradual shifts for CarlaTTA}

Denoting $\Delta t$ as a proxy for the gradual shift, we generate a \textit{dynamic} and a \textit{day2night} sequence which allow to further study the benefits of gradual TTA. By sub-sampling the sequences, we achieve varying degrees of gradual shift. For comparison, we evaluate on the least common multiple. As shown in Table \ref{tab:ablation_delta} we further benefit from a slower gradual domain shift ($\Delta t/2$), gaining another 0.5\% on \textit{day2night} and 1.4\% on \textit{dynamic}. Having a faster domain shift corresponding to a bigger delta ($2\Delta t$, $4\Delta t$) leads to a reduced performance, indicating that exploiting small gradual shifts is indeed beneficial for our self-training setup. Considering the \textit{dynamic} sequence, for very small shifts ($\Delta t/4$) we can see a slight performance degradation in comparison to $\Delta t/2$, however, the mIoU is still 0.7\% better than $\Delta t$. 

\subsection{Investigating abrupt shifts and no shifts for CarlaTTA}

Since we do not have gradual shifts all the time in the real world, we investigate two additional settings: (a) no domain shift occurs, i.e., the test domain is equal to the source domain and (b) the domain changes abruptly. The results are reported in Table \ref{tab:ablation_carla} (a) and (b). For the setting of (a), where no domain shift occurs, BN--0, as expected, shows the best performance with a mIoU of 78.4\%. Updating the batch statistics through an exponential moving average, i.e., considering BN--EMA, leads to a performance decrease of 0.8\%. As a result, also GTTA-ST's performance cannot achieve BN--0. BN--1 performs even worse at a mIoU of 76.6\% illustrating that even in the setting of segmentation in an urban scene a perfect estimation of the batch normalization statistics is not possible for a single sample. For the investigation how GTTA-ST handles abrupt shifts, instead of starting at the first sample of the \textit{day2night} sequence, we begin after 300 samples, corresponding to a sun altitude of 0°. After one complete night-cycle, the sequence reaches its end. Our method still shows the capability to adapt, reaching a 5.4\% higher mIoU than the best BN adaptation approach, namely BN--EMA.

\begin{figure*}[h]
    \centering
    \includegraphics[scale=0.70]{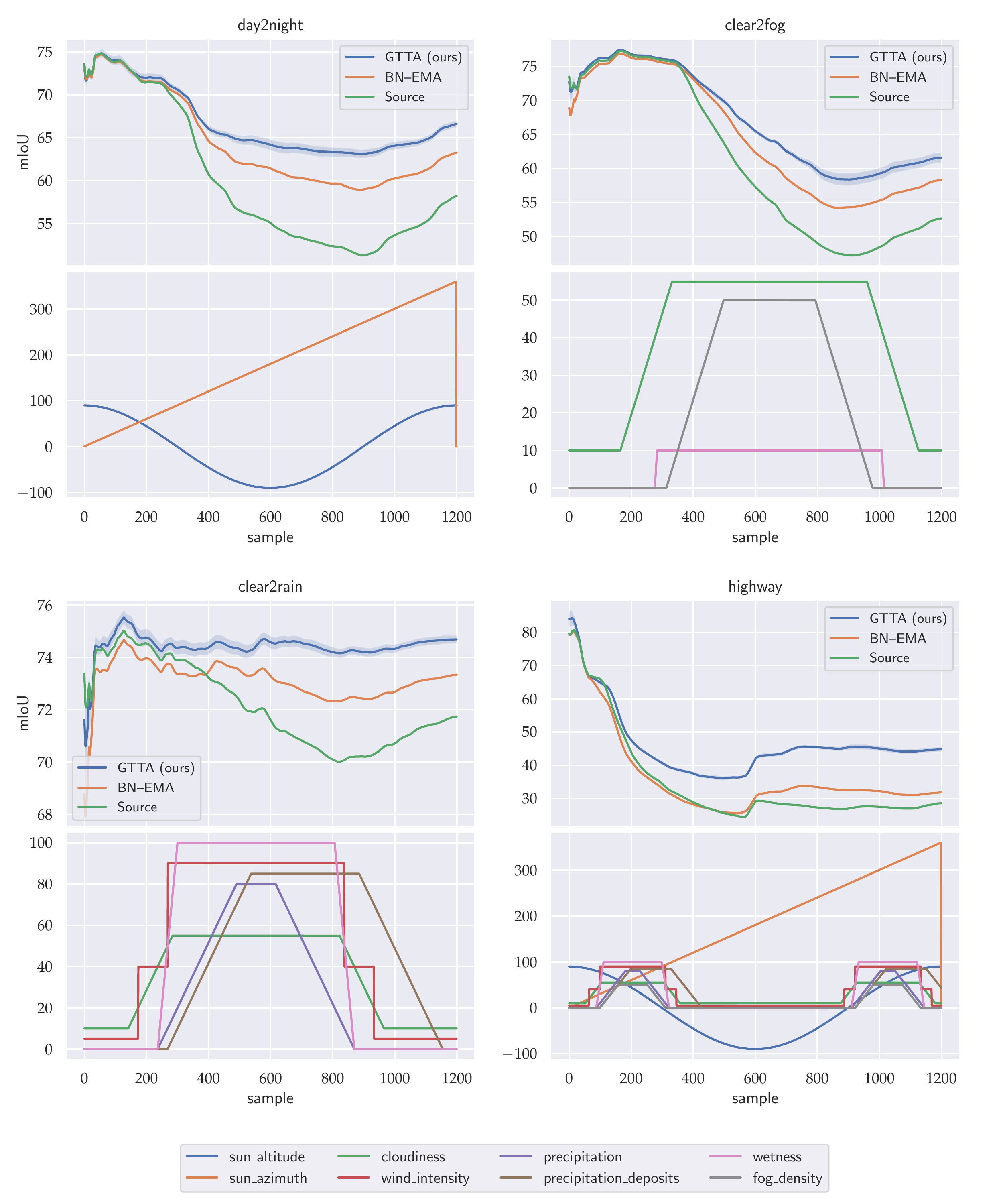}
    \caption{mIoU up to time step $t$ for \textit{day2night}, \textit{clear2fog}, \textit{clear2rain}, and \textit{highway}. Additionally, we illustrate the progression of the weather parameters to have a direct comparison when a domain shift occurs.}
    \label{fig:sequences_carla}
\end{figure*}
\subsection{Performance over time on CarlaTTA}
In Figure \ref{fig:sequences_carla} we illustrate the sequences \textit{day2night}, \textit{clear2fog}, \textit{clear2rain}, and \textit{highway} and their corresponding progression of environment conditions. At time step $t$ we visualize the mean intersection-over-union averaged from the beginning $t=0$ up to time step $t$. For the \textit{day2night} sequence a decline in performance can be seen at around sample 300, corresponding to the sun setting. In the setting \textit{clear2fog} the increase in cloudiness does not influence the model's performance much. With the fog density increasing, a steady decrease in performance can be seen. For \textit{clear2rain} the performance of GTTA-ST stays more or less constant, while the source performance steadily decreases with the start of the rain. The \textit{highway} sequence experiences the same weather behaviour as the \textit{dynamic} sequence, but additionally experiences a shift of urban scenery to a highway scene. As a result, the performance decreases drastically in the beginning. At around sample 550 an increase in performance can be experienced. This is due to the car driving back into the city, leaving the city again at sample 770.

\newpage
\section{Dataset: CarlaTTA}
\label{appendix:dataset}
\begin{figure}
  \centering
  \begin{tabular}{c}
    \includegraphics[width=6in]{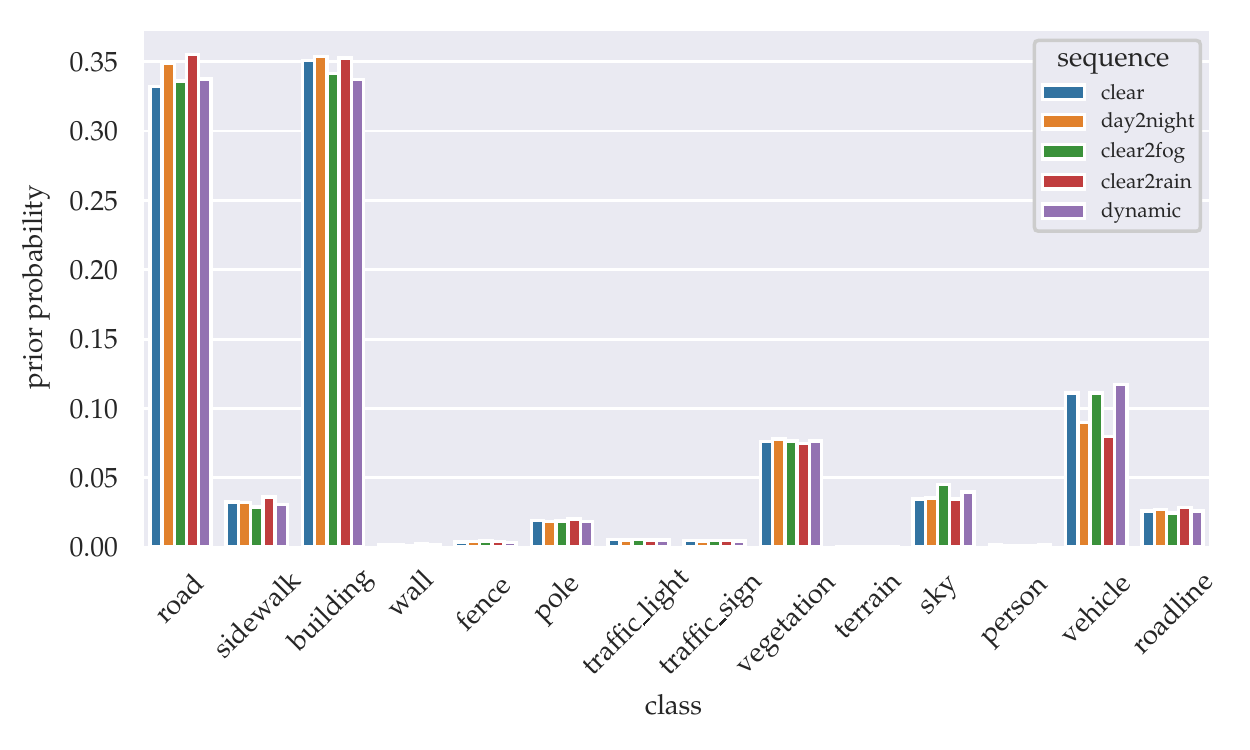}\\ \includegraphics[width=6in]{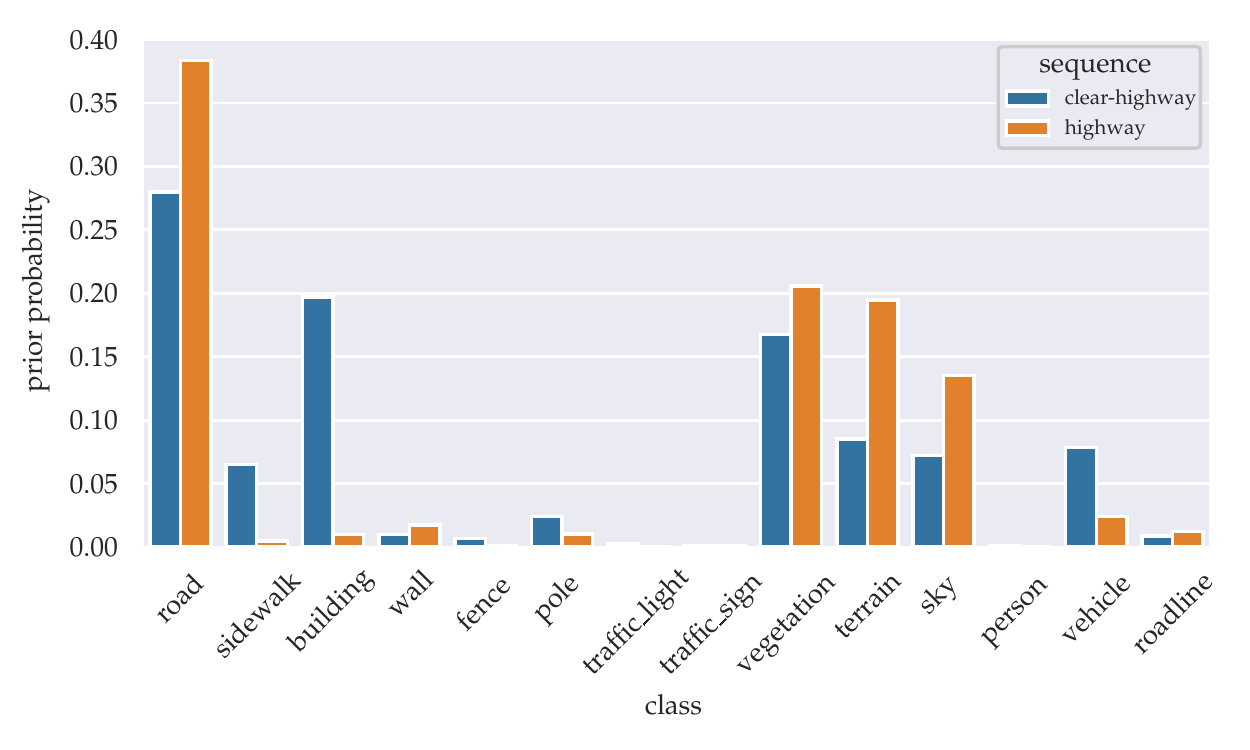}\\
  \end{tabular}
  \caption{Prior probabilities for the 14 classes. On the top we visualize the prior probabilities corresponding to data originating from Town10HD. On the bottom, we compare \textit{clear-highway} to \textit{highway}. The label distribution shift is clearly prominent as the test-time sequence evolves from an urban scenery into a highway setting.}
  \label{fig:priors}
\end{figure}

In the following, we will discuss the specifics about our dataset CarlaTTA. For the generation, CARLA 0.9.13 was used. The data is recorded using an RGB camera and a corresponding semantic segmentation sensor with a resolution of $1920\times1024$ and a field of view of 40°. Both sensors are positioned 0.5 m forward and 1.2 m upward relative to the ego-vehicle. 14 classes are considered. We visualize the class priors for all sequences in Figure \ref{fig:priors}.

\paragraph{clear}
\begin{table*}[h]
\renewcommand{\arraystretch}{1.1}
\centering
\caption{\textit{clear} weather parameters and maximum values used for rain and fog in the settings \textit{clear2rain} and \textit{clear2fog}, respectively.}
\scalebox{0.8}{\tabcolsep5pt
\label{tab:clear_parameters}
\begin{tabular}{l|cc|cc|ccc|ccc}
    \hline
    & \multicolumn{2}{c|}{sun} & \multicolumn{2}{c|}{} & \multicolumn{3}{c|}{rain} & \multicolumn{3}{c}{fog}  \\
    & altitude & azimuth & cloudiness & wind & precipitation & deposits & wetness & density & distance & falloff\\
    \hline
    unit & ° & ° & \% & \% & \% & \% & \% & \% & m & -\\
    \hline
    clear & 90 & 0 & 10 & 5 & 0 & 0 & 0 & 0 & 0 & 0 \\
    night (max) & -90 & 180 & - & - & - & - & - & - & - & - \\
    rain (max) & - & - & 55 & 90 & 80 & 85 & 100 & - & - & - \\
    fog (max) & - & - & 55 & 90 & - & - & 10 & 50 & 0 & 0.9 \\
    \hline
\end{tabular}}
\end{table*}
The source dataset \textit{clear} is the basis for the four domain changes: \textit{day2night}, \textit{clear2fog}, \textit{clear2rain}, and \textit{dynamic}. The data for \textit{clear} is recorded in Town10HD due to being the only town in CARLA with high resolution textures. To increase diversity we generate multiple sequences using different seeds. Specifically, for each seed, up to 40 vehicles and 20 pedestrians are randomly sampled from all (safe) blueprints. We end up with 3500 train and 500 test samples. \textit{clear} is recorded at noon (sun altitude of 90°) with a cloudiness of 10\% and a wind intensity of 5\%. All weather parameters are fixed in this setting. The complete overview of the weather parameters used for the \textit{clear} setting is given in Table \ref{tab:clear_parameters}.

\begin{figure}
  \centering
    \includegraphics[width=\textwidth]{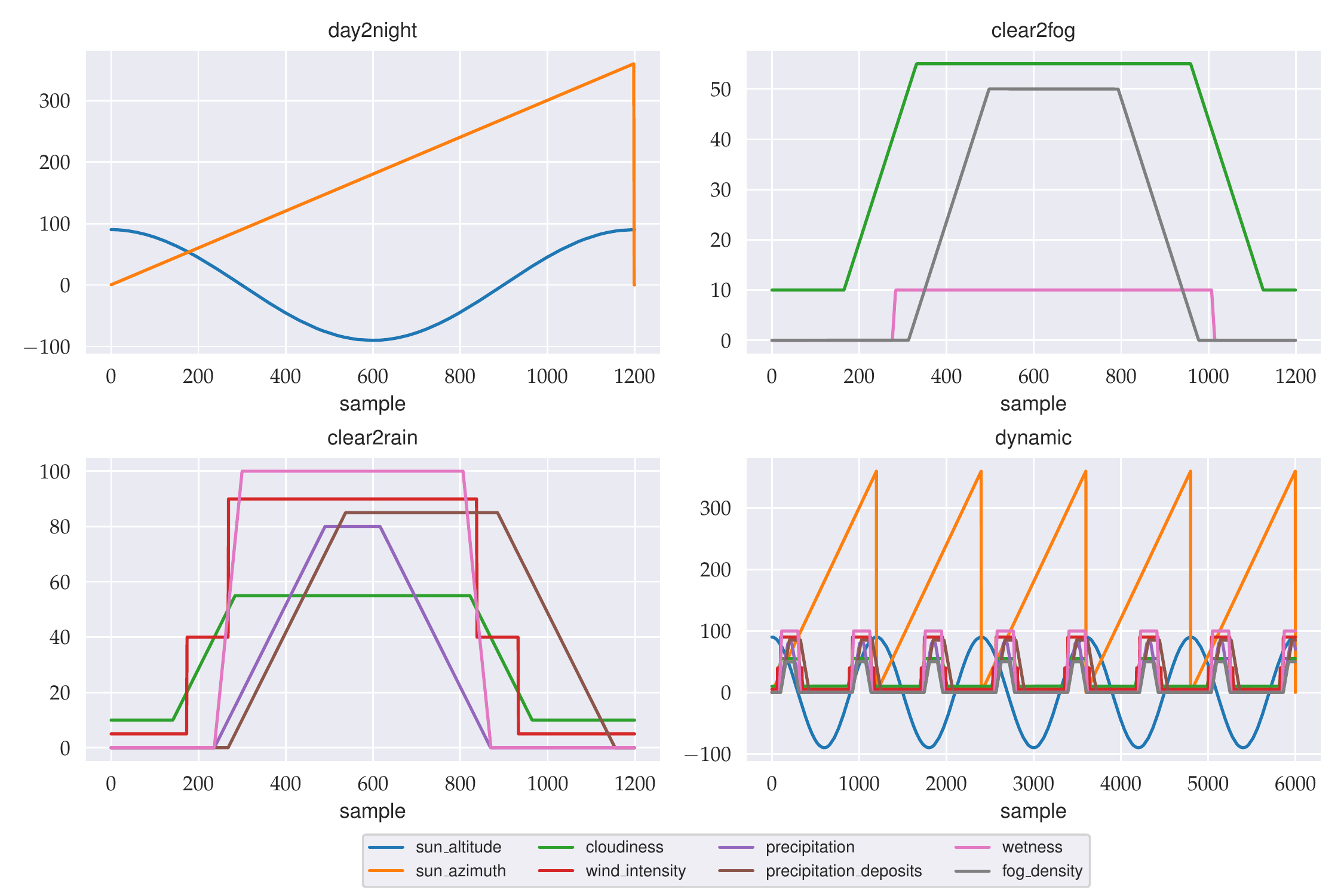}
  \caption{Progression of the weather parameters for the sequences \textit{day2night}, \textit{clear2fog}, \textit{clear2rain}, and \textit{dynamic}. Note that \textit{highway} follows the weather behavior of \textit{dynamic}.}
  \label{fig:weather_params}
\end{figure}

\begin{figure}
  \centering
  \setlength{\tabcolsep}{1pt}
  \begin{tabular}{ccccc}
    day2night & clear2fog & clear2rain & dynamic & highway\\
    \includegraphics[width=2.8cm]{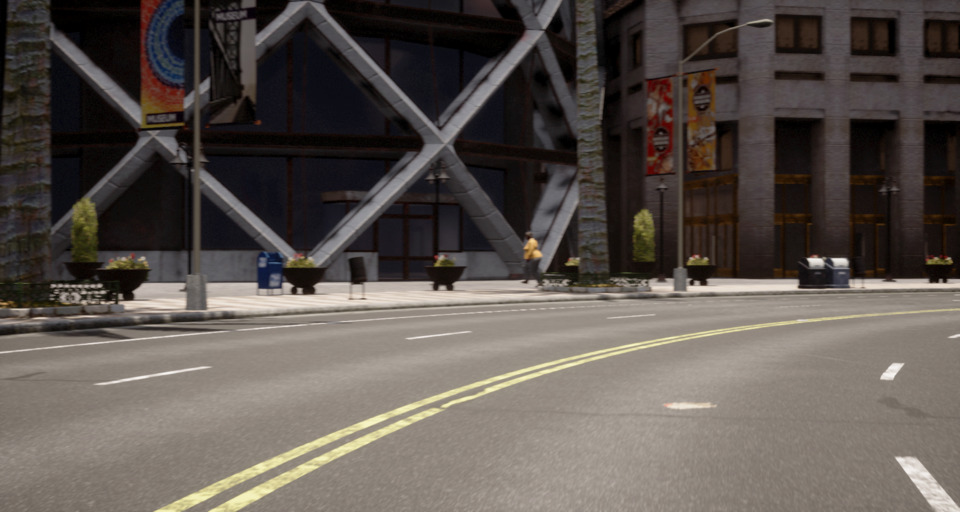} & \includegraphics[width=2.8cm]{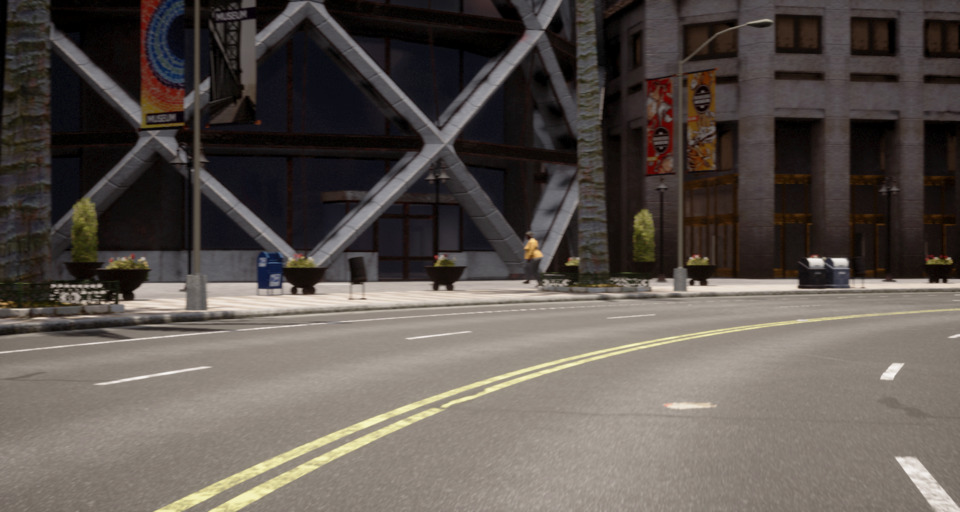} &
    \includegraphics[width=2.8cm]{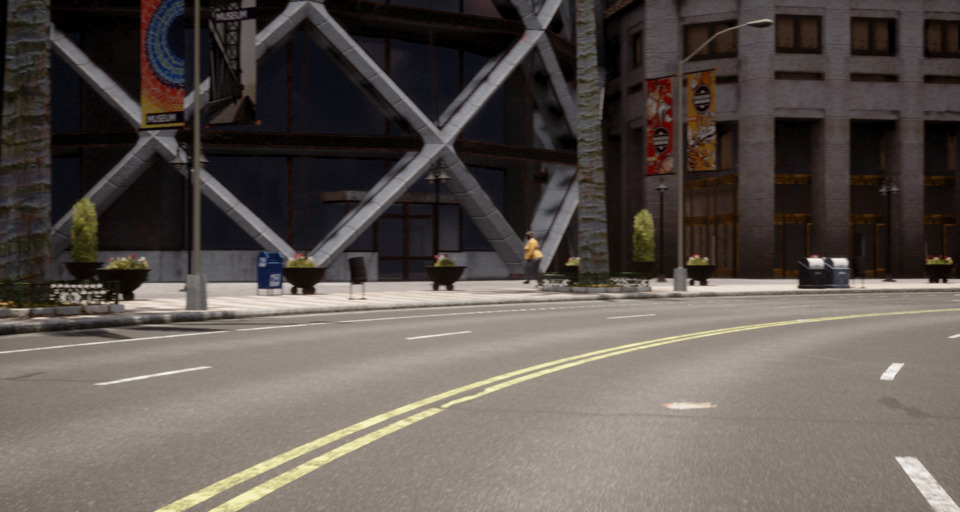} &
    \includegraphics[width=2.8cm]{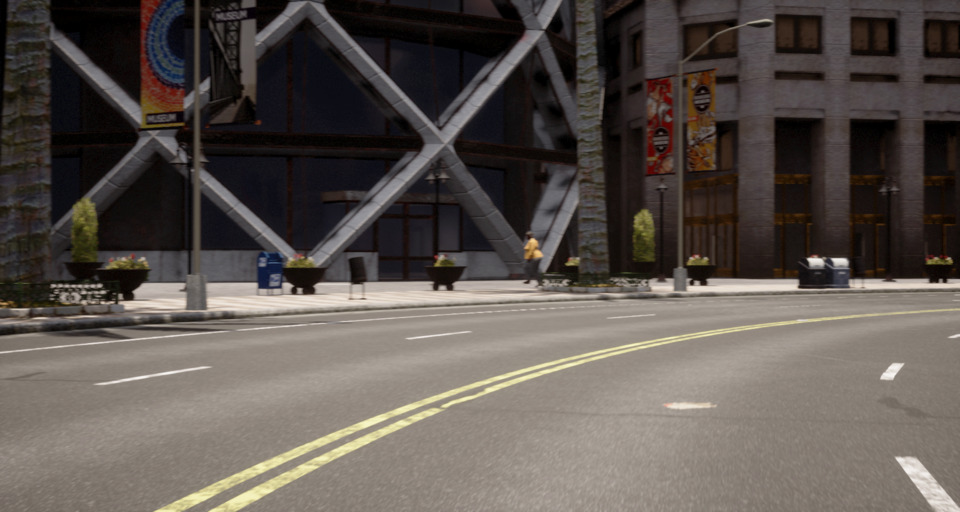} &
    \includegraphics[width=2.8cm]{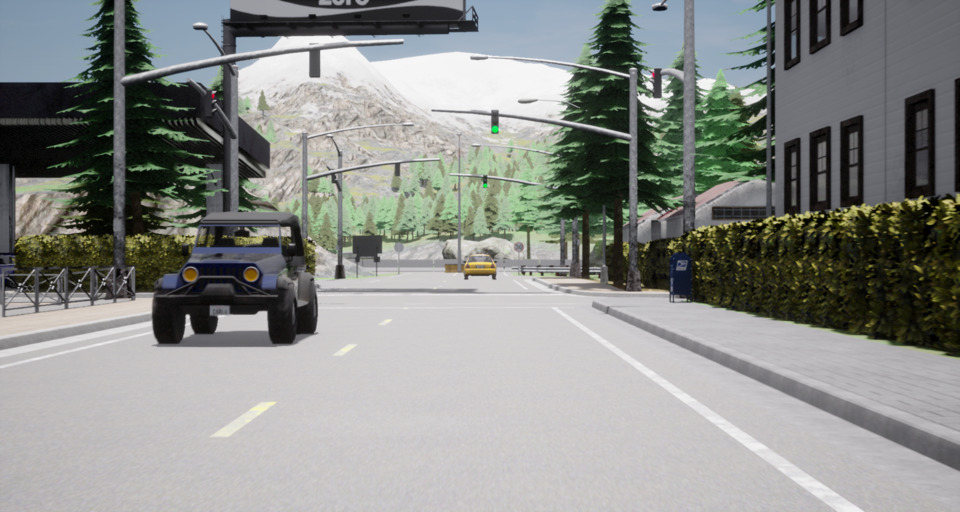} \\[-4pt]
    \includegraphics[width=2.8cm]{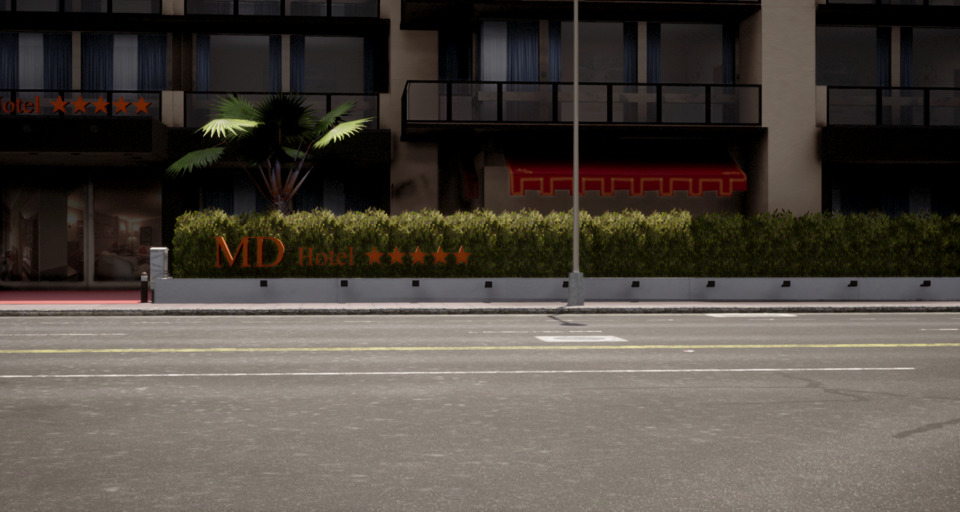} & \includegraphics[width=2.8cm]{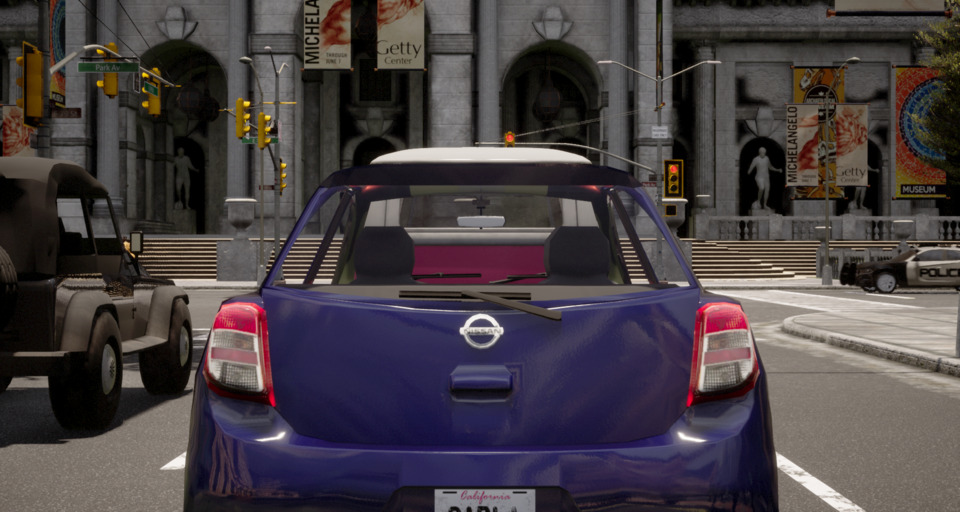} &
    \includegraphics[width=2.8cm]{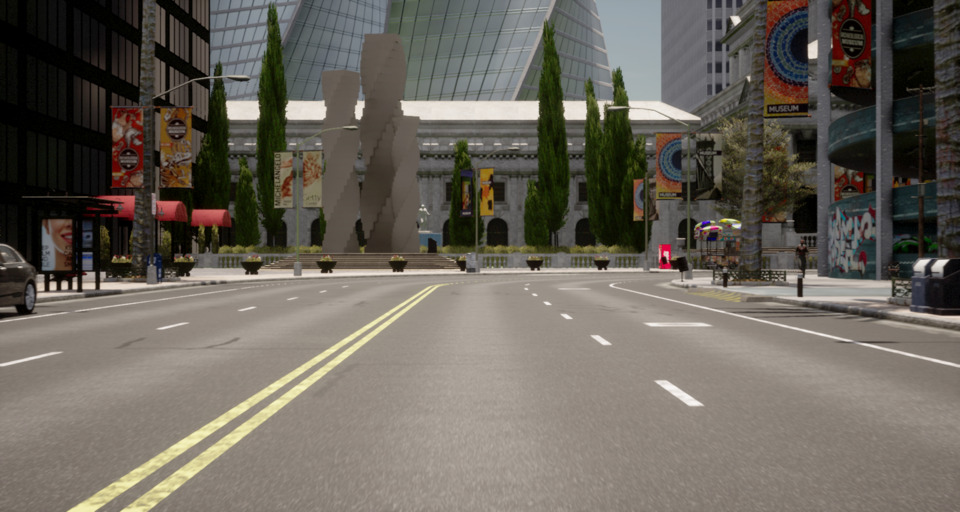} &
    \includegraphics[width=2.8cm]{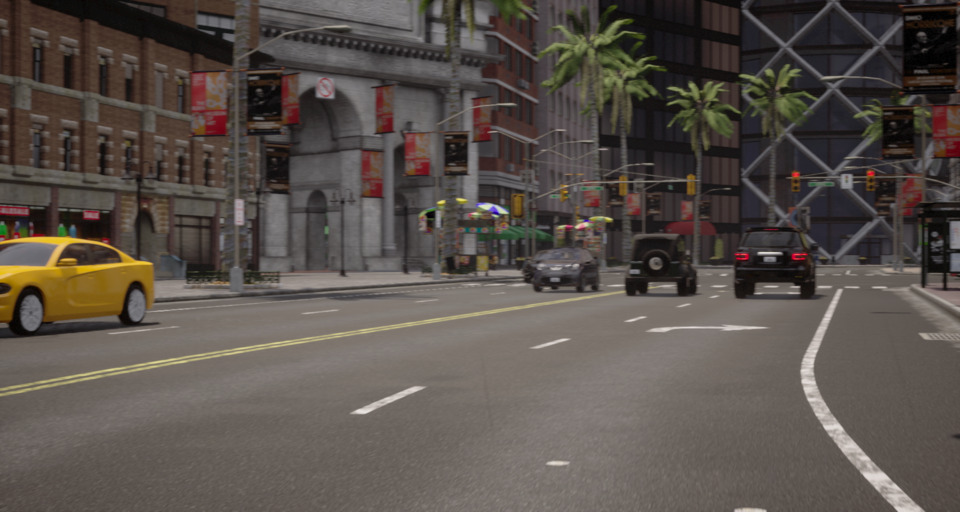} &
    \includegraphics[width=2.8cm]{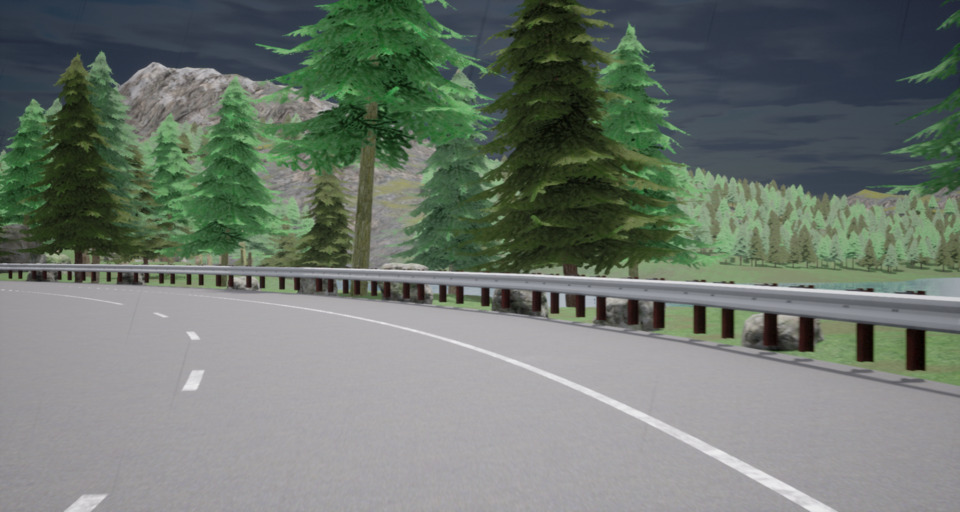} \\[-4pt]
    \includegraphics[width=2.8cm]{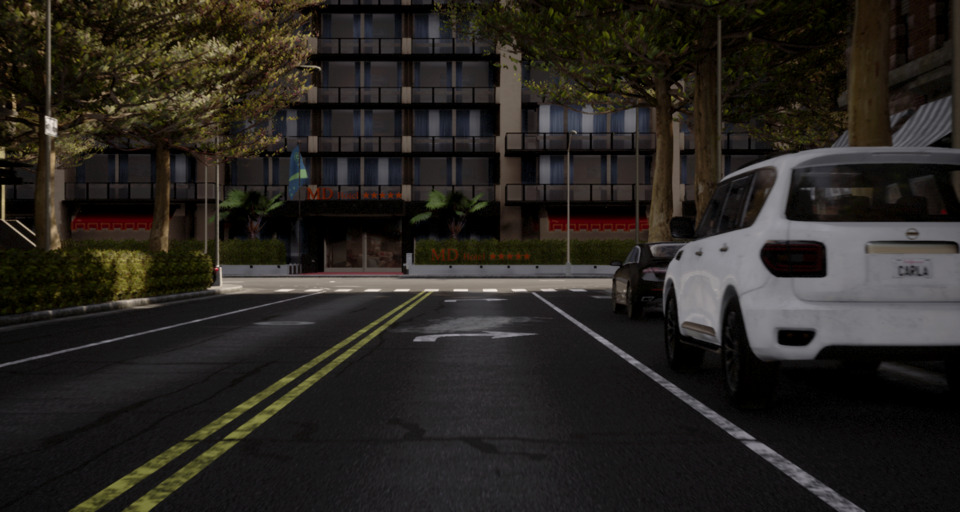} & \includegraphics[width=2.8cm]{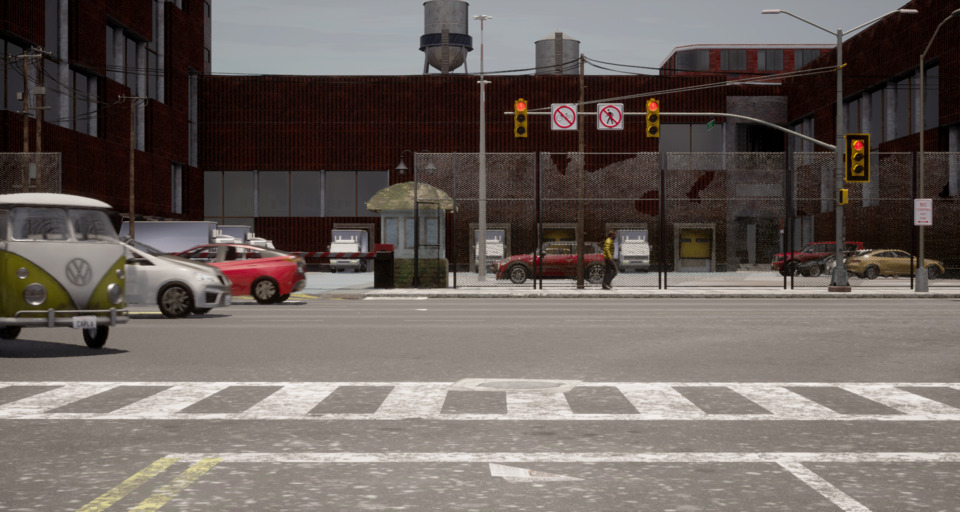} &
    \includegraphics[width=2.8cm]{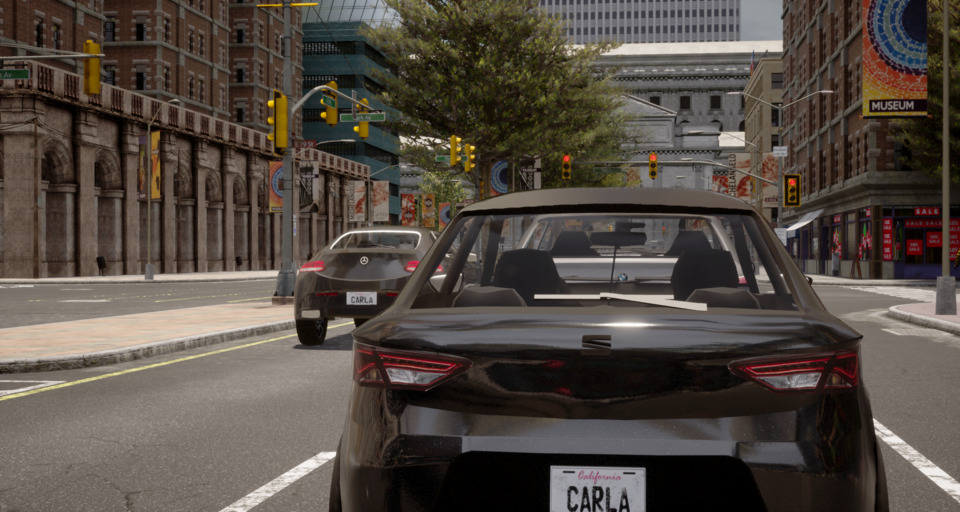} &
    \includegraphics[width=2.8cm]{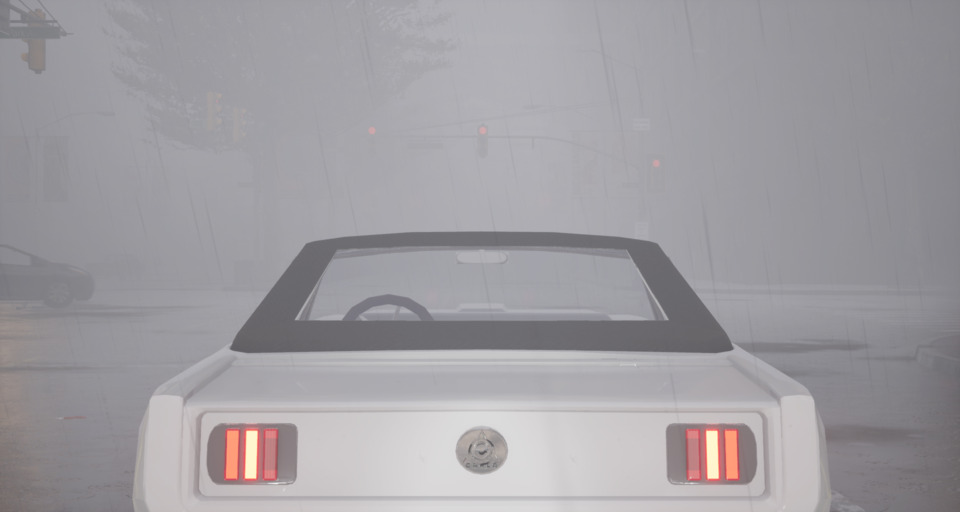} &
    \includegraphics[width=2.8cm]{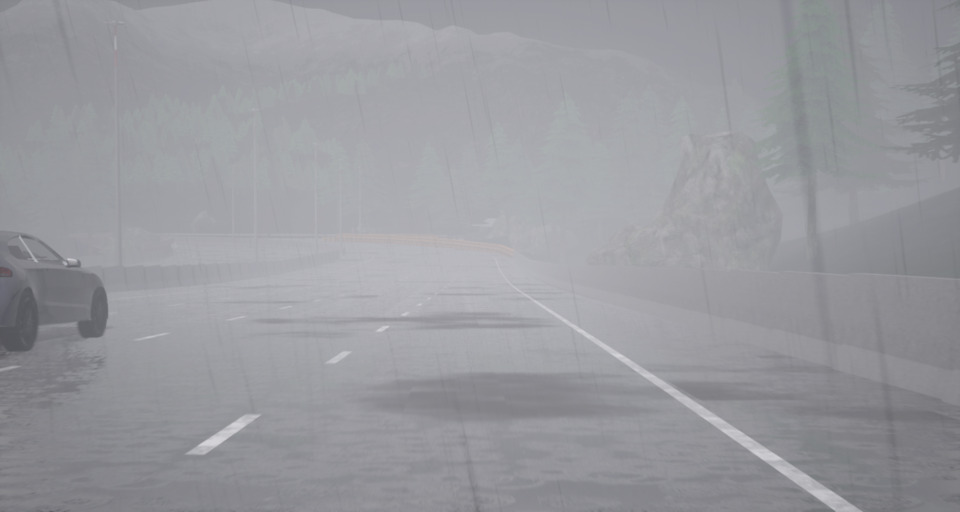} \\[-4pt]
    \includegraphics[width=2.8cm]{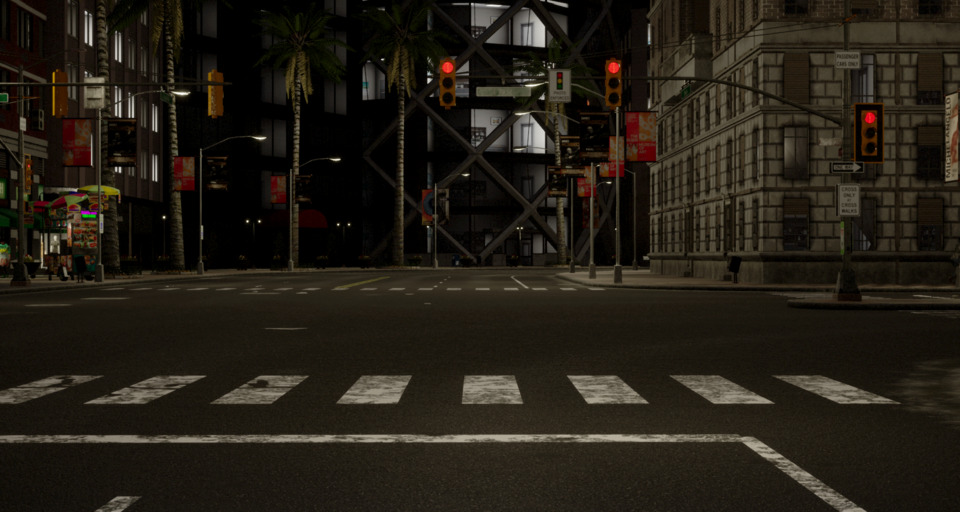} & \includegraphics[width=2.8cm]{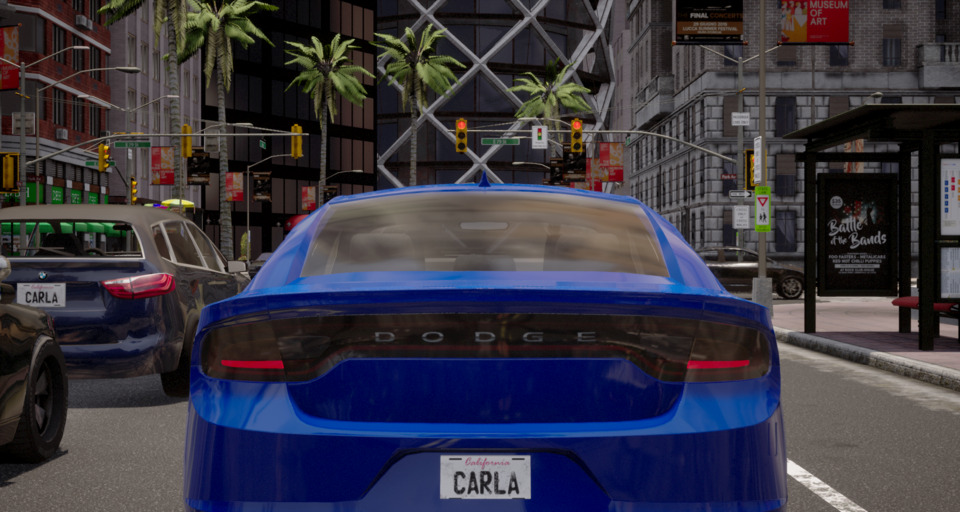} &
    \includegraphics[width=2.8cm]{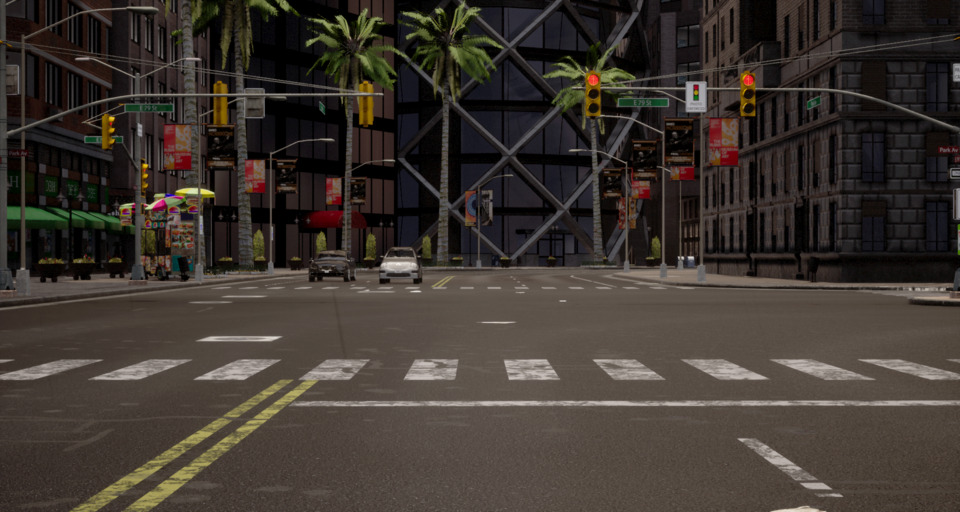} &
    \includegraphics[width=2.8cm]{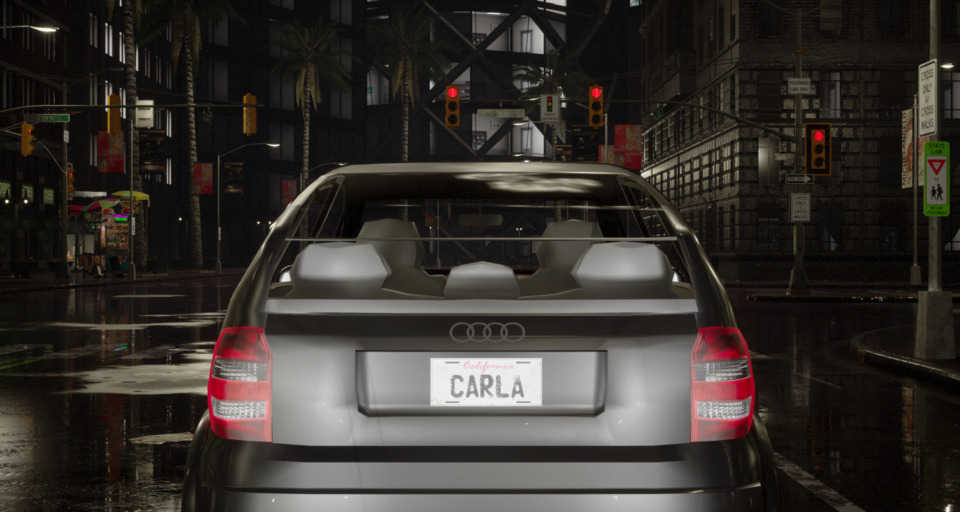} &
    \includegraphics[width=2.8cm]{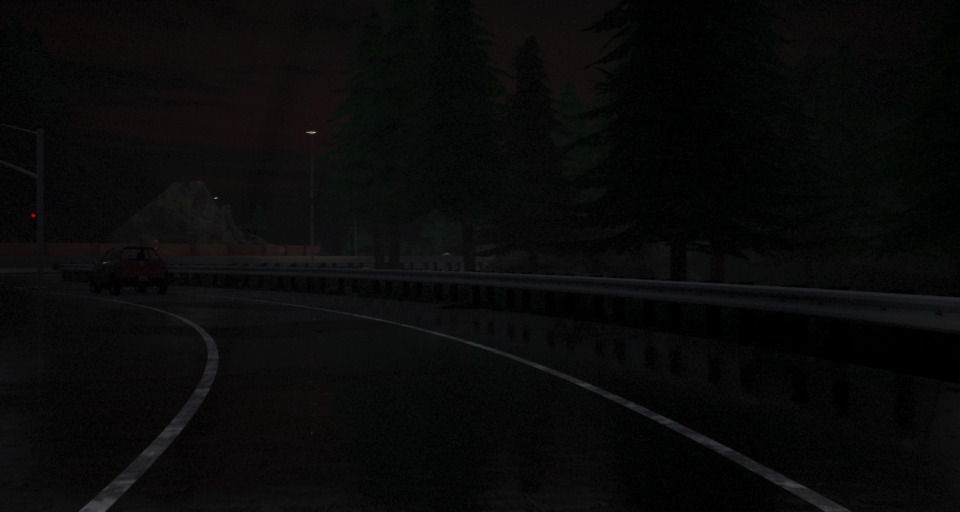} \\[-4pt]
    \includegraphics[width=2.8cm]{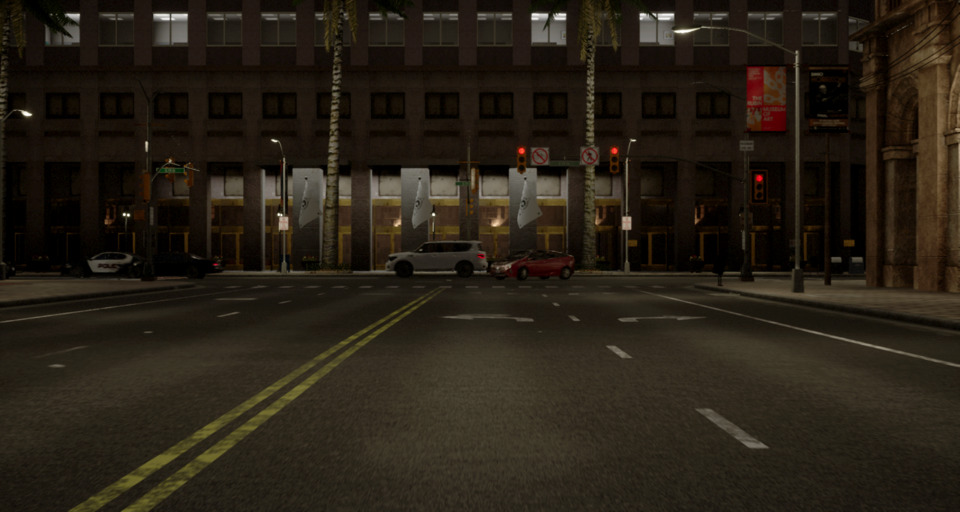} & \includegraphics[width=2.8cm]{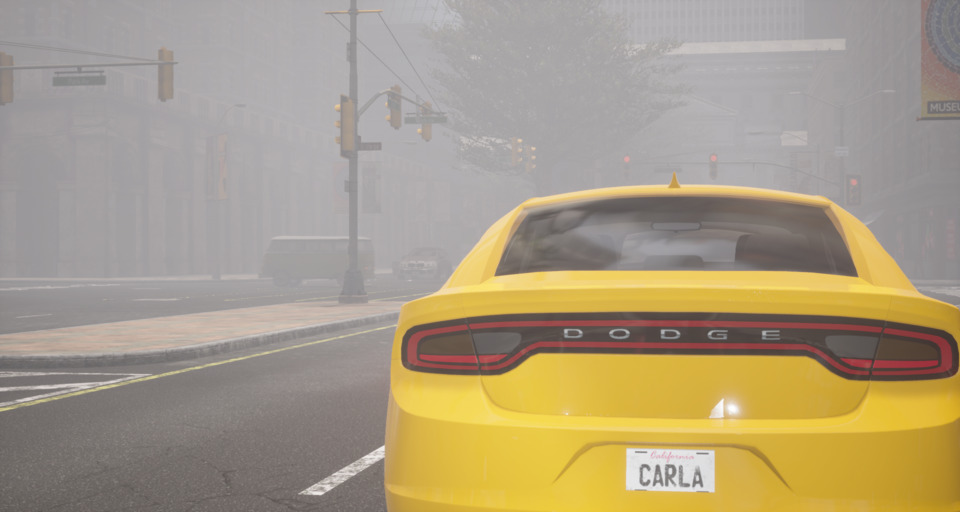} &
    \includegraphics[width=2.8cm]{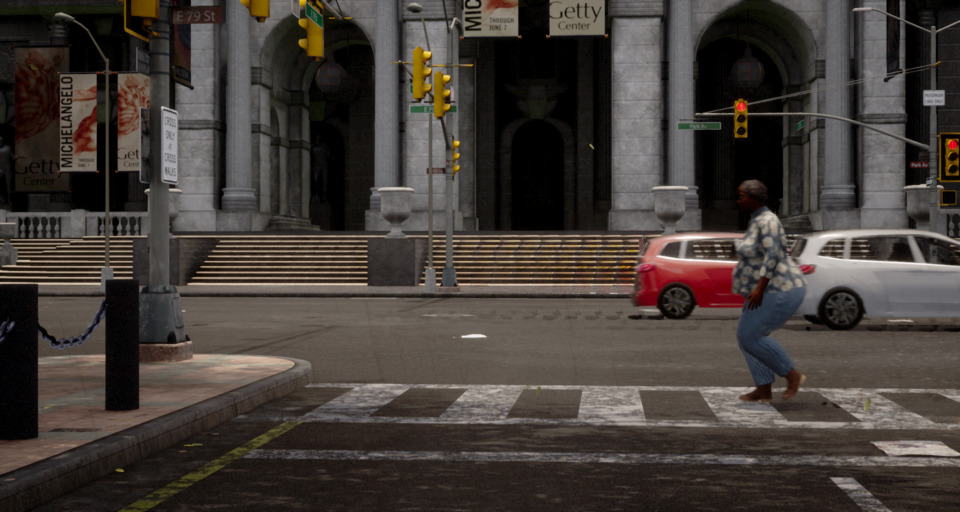} &
    \includegraphics[width=2.8cm]{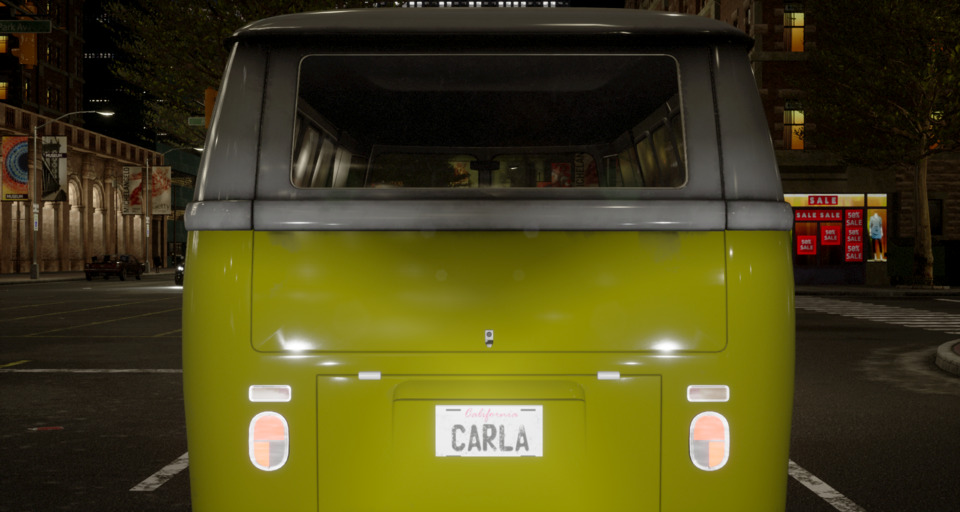} &
    \includegraphics[width=2.8cm]{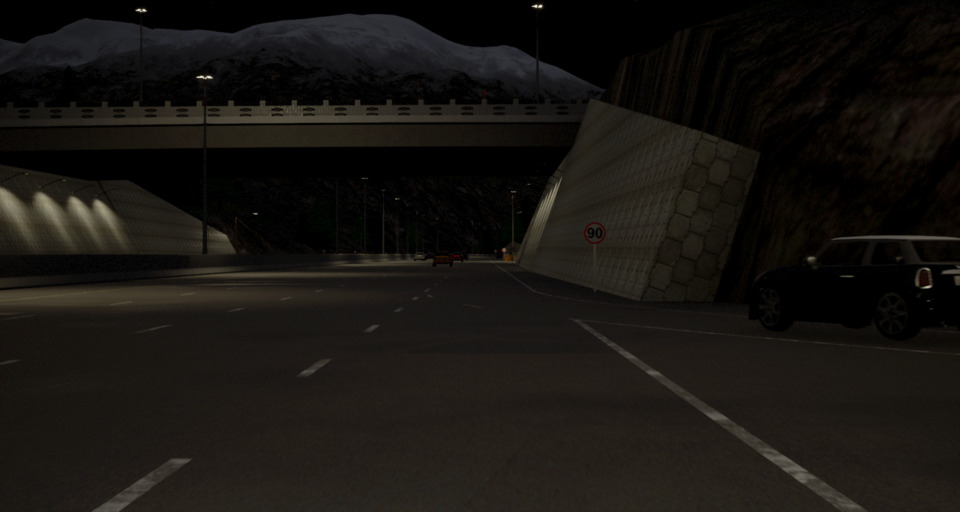} \\[-4pt]
    \includegraphics[width=2.8cm]{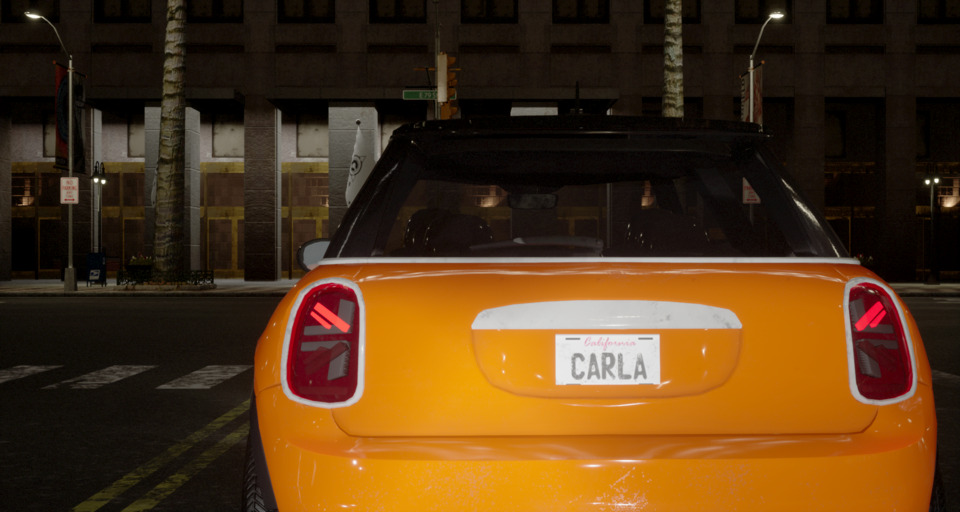} & \includegraphics[width=2.8cm]{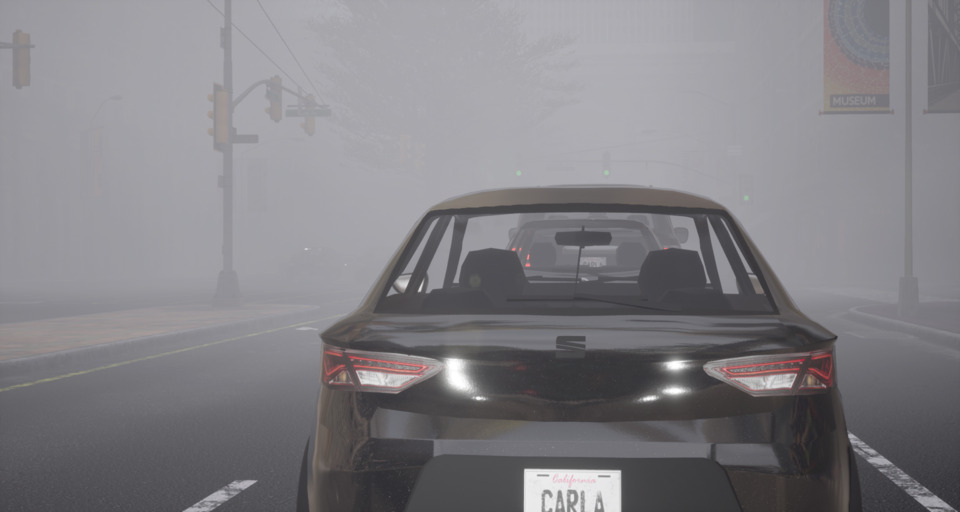} &
    \includegraphics[width=2.8cm]{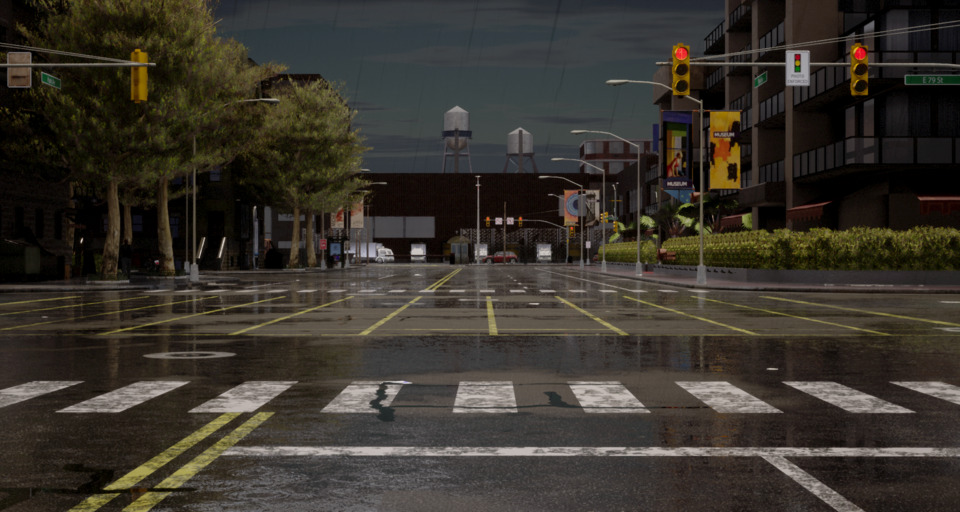} &
    \includegraphics[width=2.8cm]{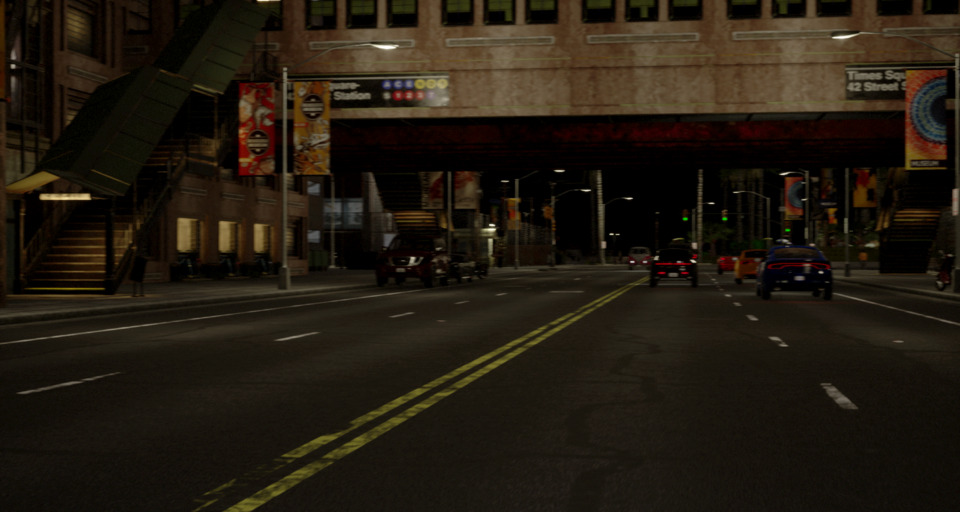} &
    \includegraphics[width=2.8cm]{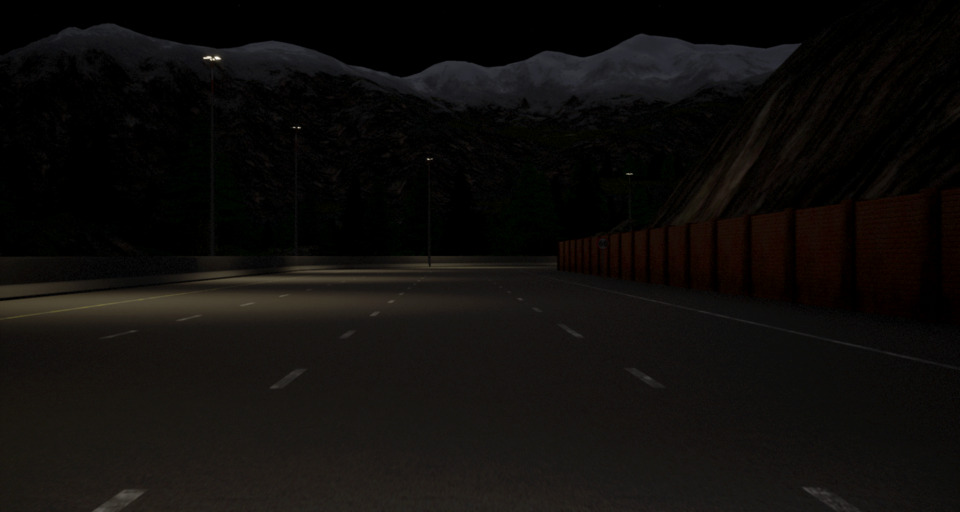} \\[-4pt]
    \includegraphics[width=2.8cm]{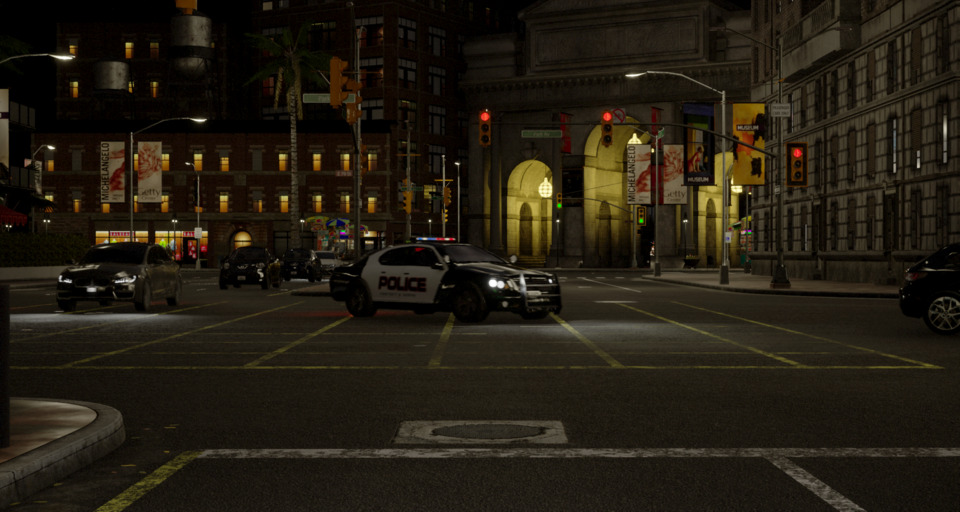} & \includegraphics[width=2.8cm]{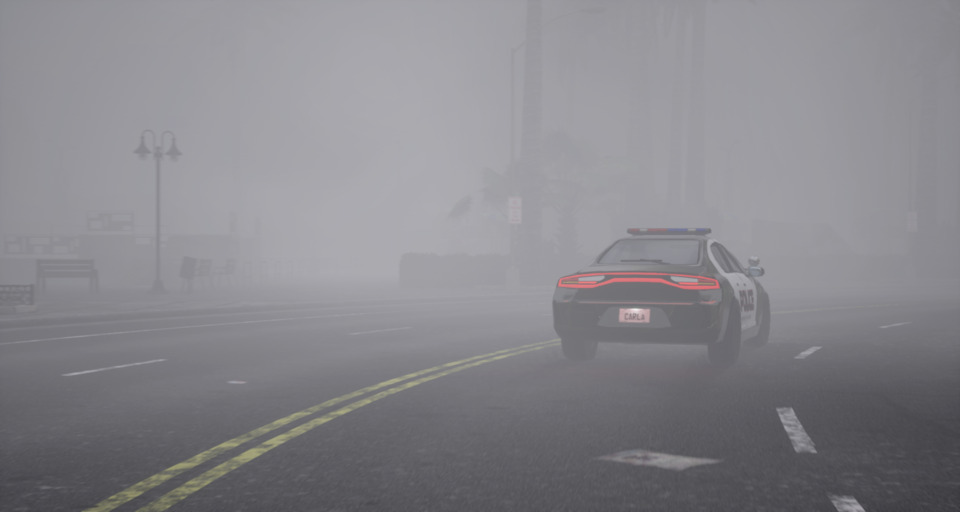} &
    \includegraphics[width=2.8cm]{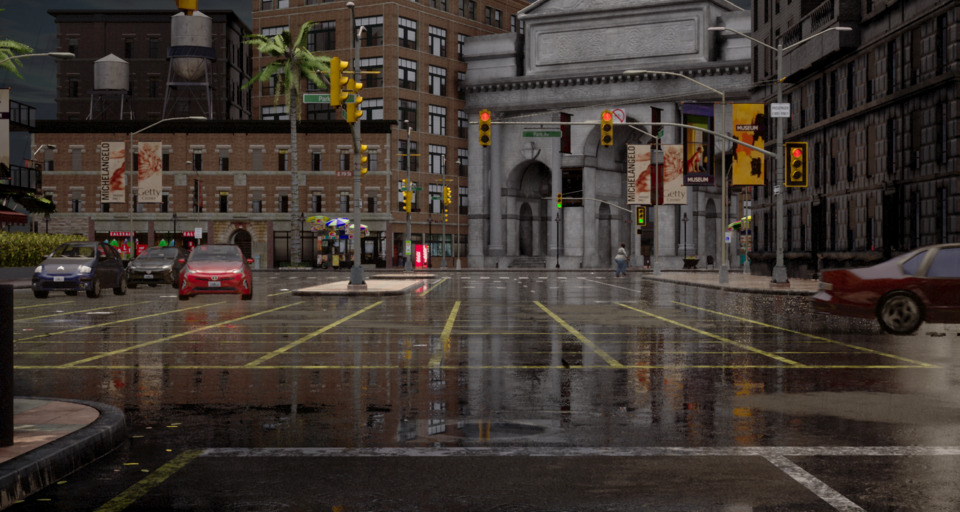} &
    \includegraphics[width=2.8cm]{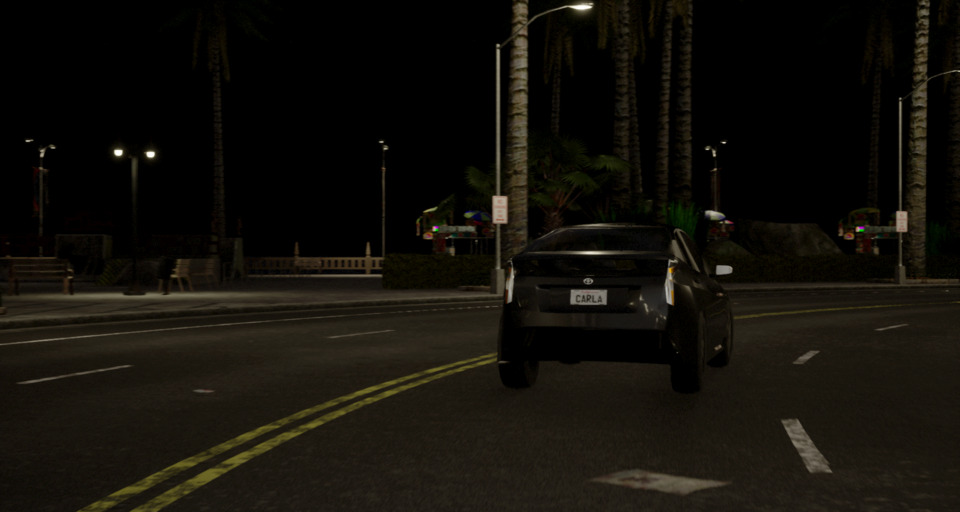} &
    \includegraphics[width=2.8cm]{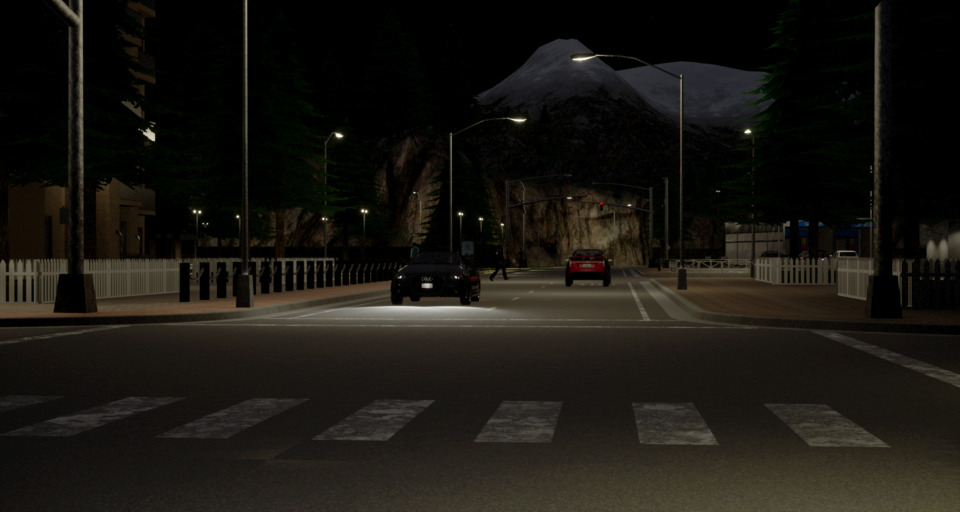} \\[-4pt]
    \includegraphics[width=2.8cm]{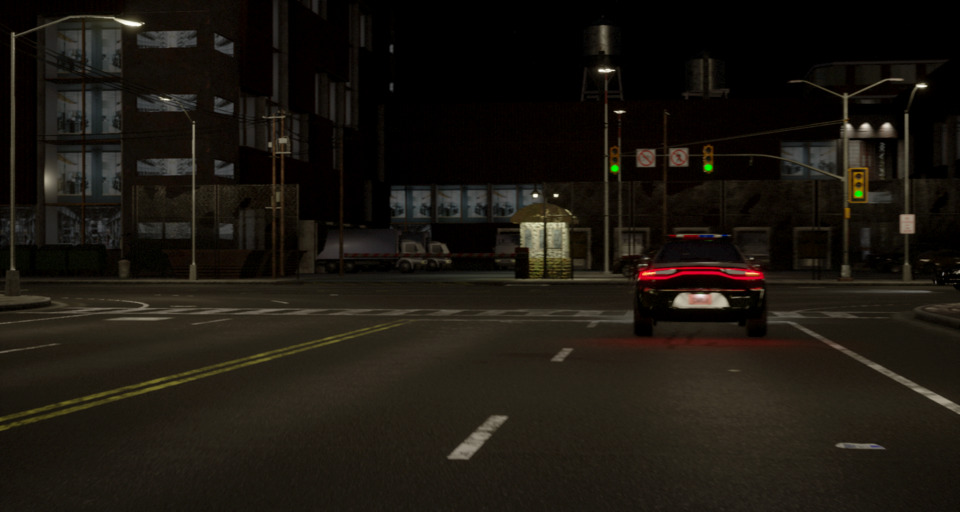} & \includegraphics[width=2.8cm]{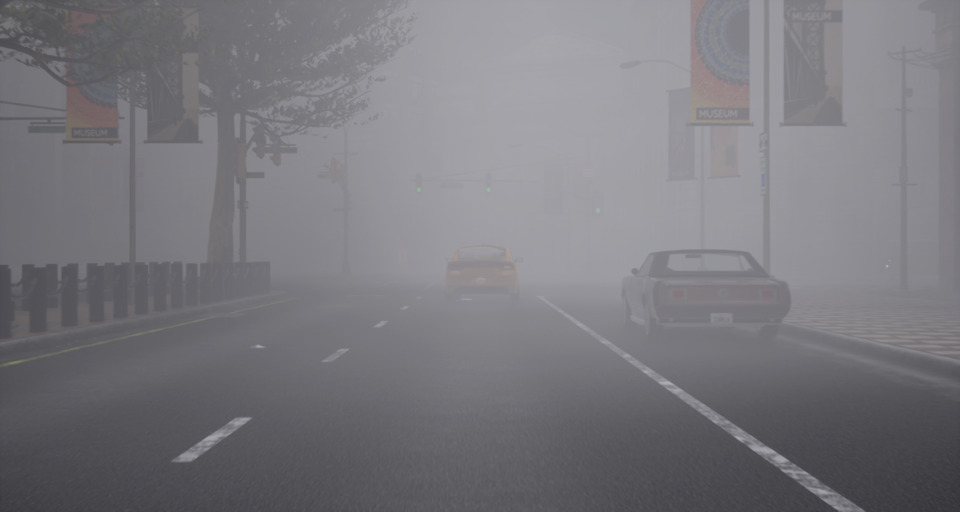} &
    \includegraphics[width=2.8cm]{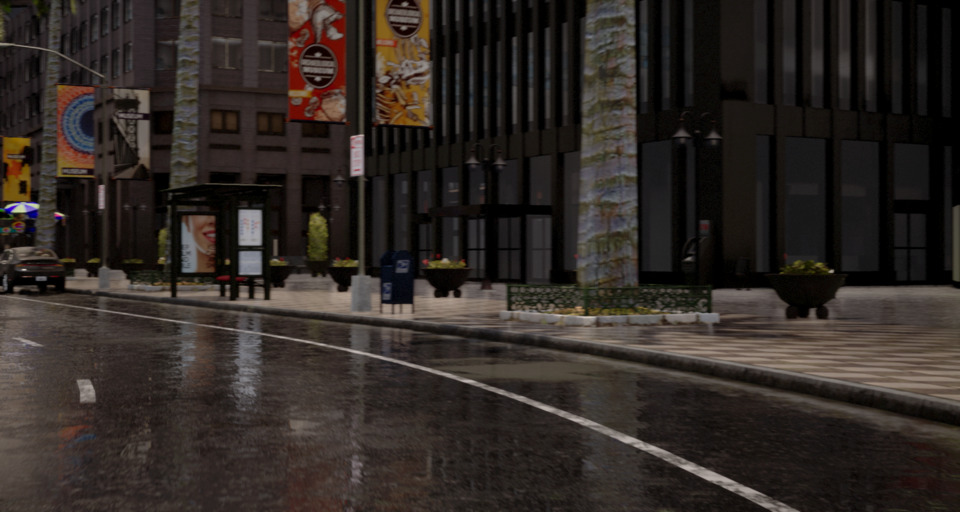} &
    \includegraphics[width=2.8cm]{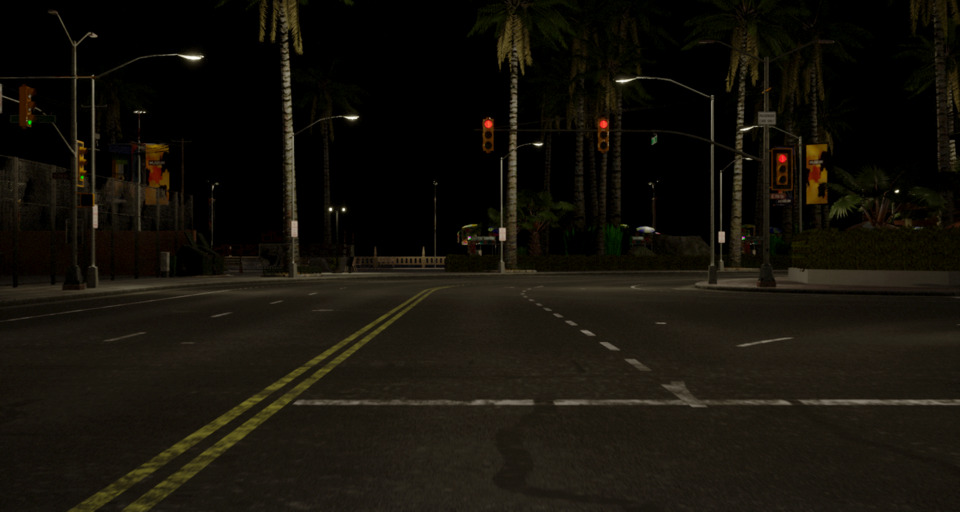} &
    \includegraphics[width=2.8cm]{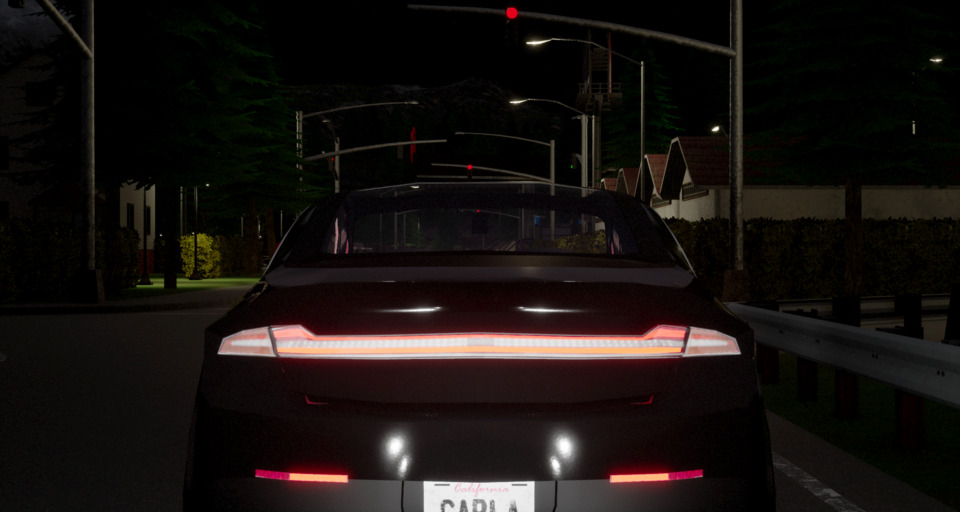} \\[-4pt]
    \includegraphics[width=2.8cm]{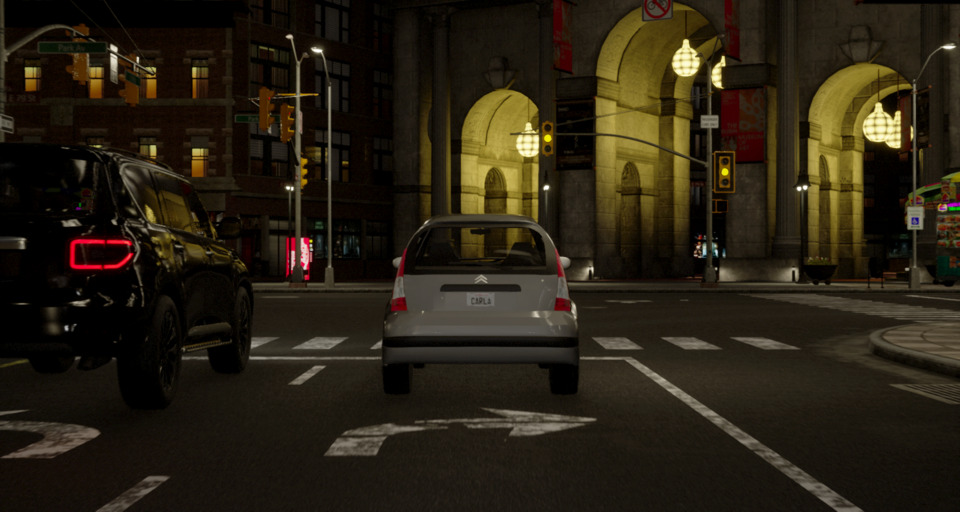} & \includegraphics[width=2.8cm]{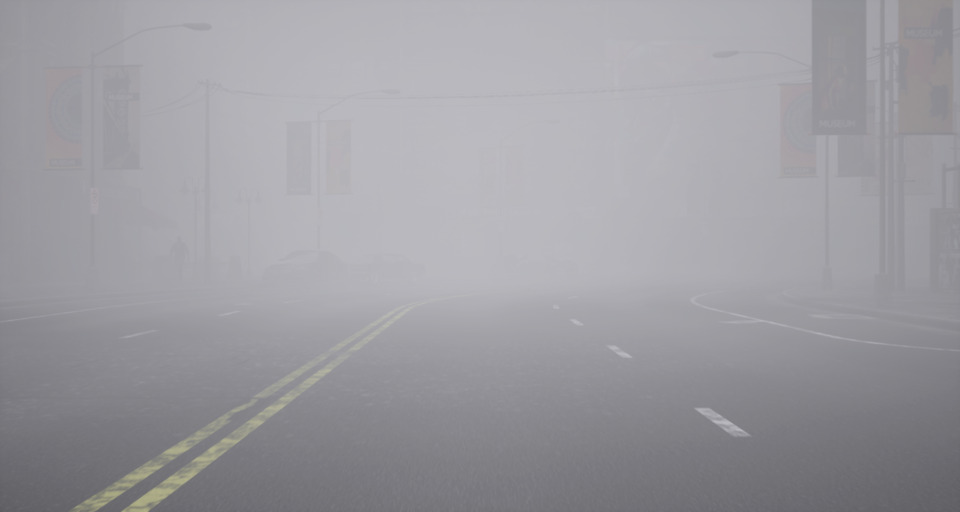} &
    \includegraphics[width=2.8cm]{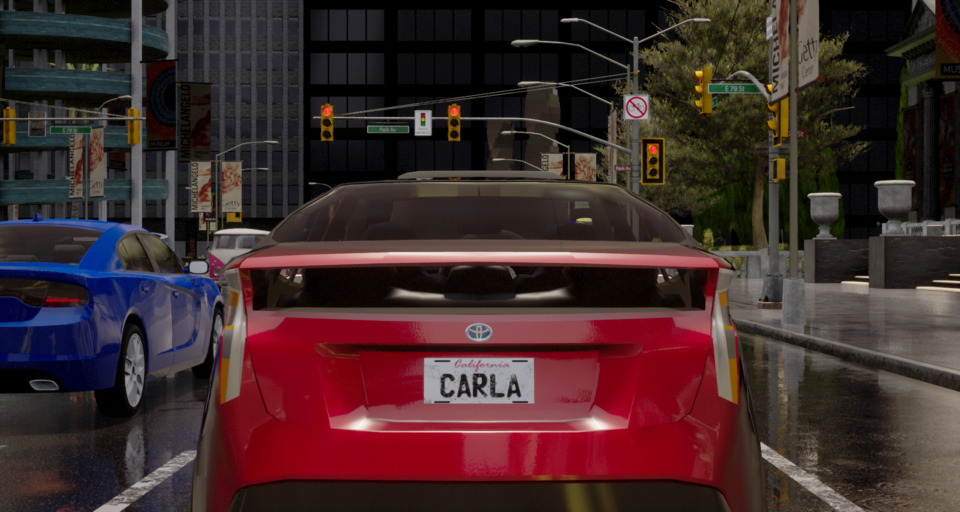} &
    \includegraphics[width=2.8cm]{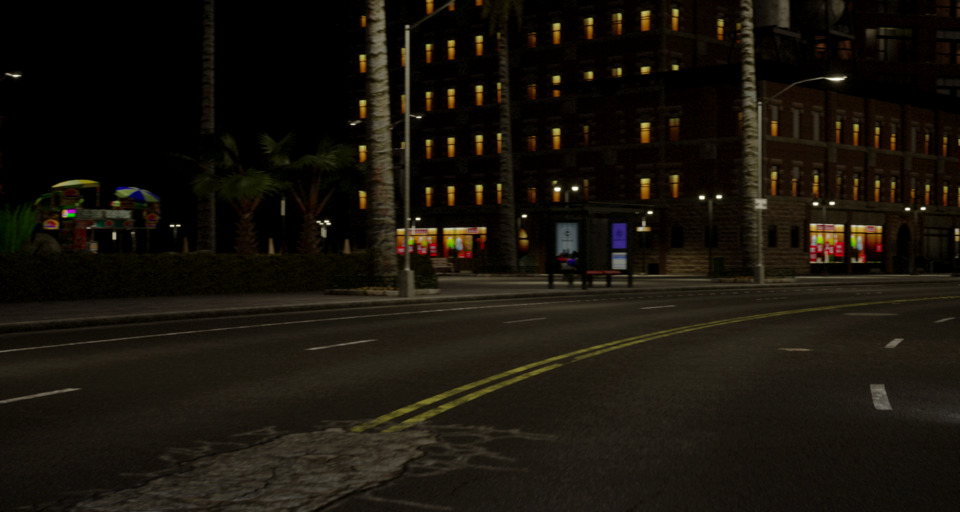} &
    \includegraphics[width=2.8cm]{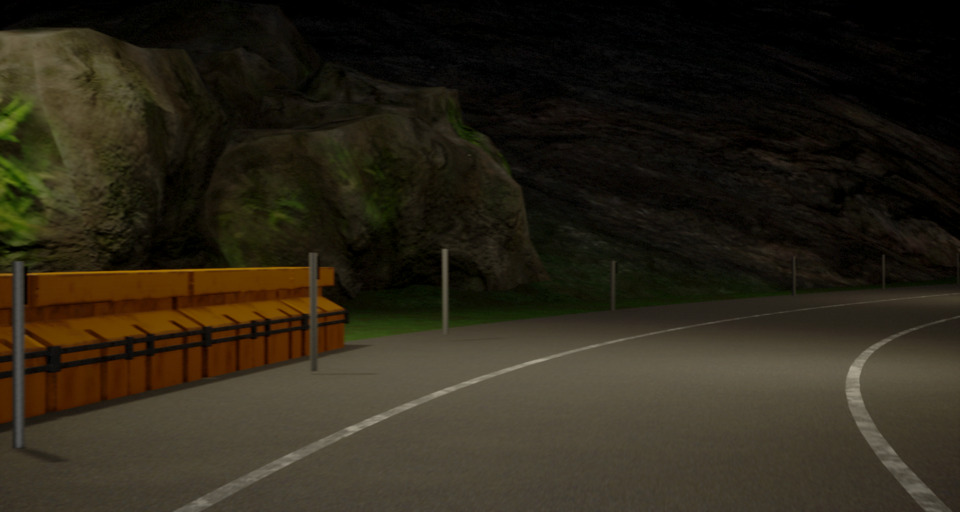} \\[-4pt]
    \includegraphics[width=2.8cm]{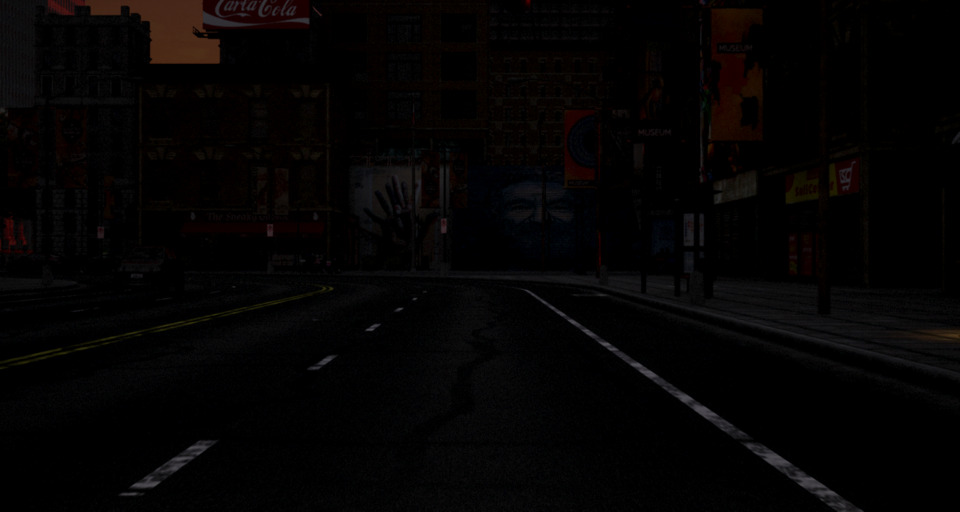} & \includegraphics[width=2.8cm]{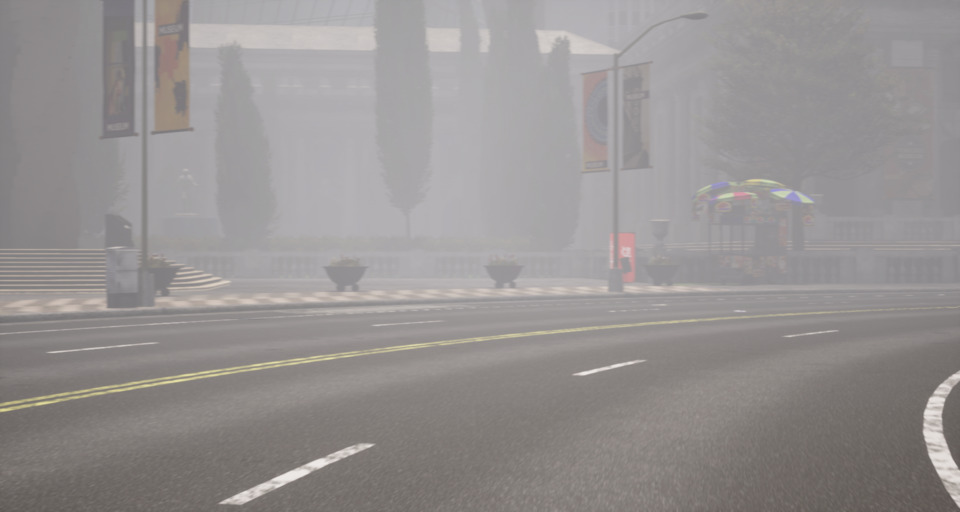} &
    \includegraphics[width=2.8cm]{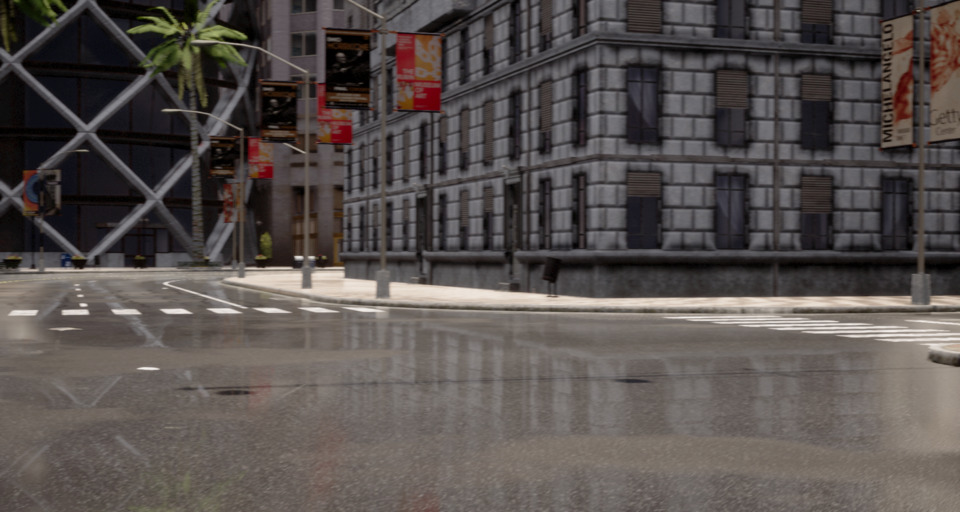} &
    \includegraphics[width=2.8cm]{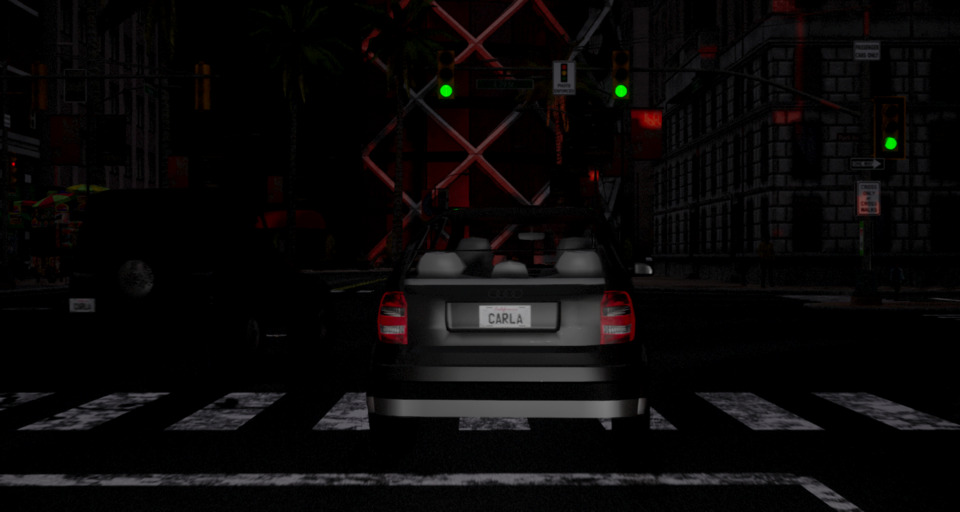} &
    \includegraphics[width=2.8cm]{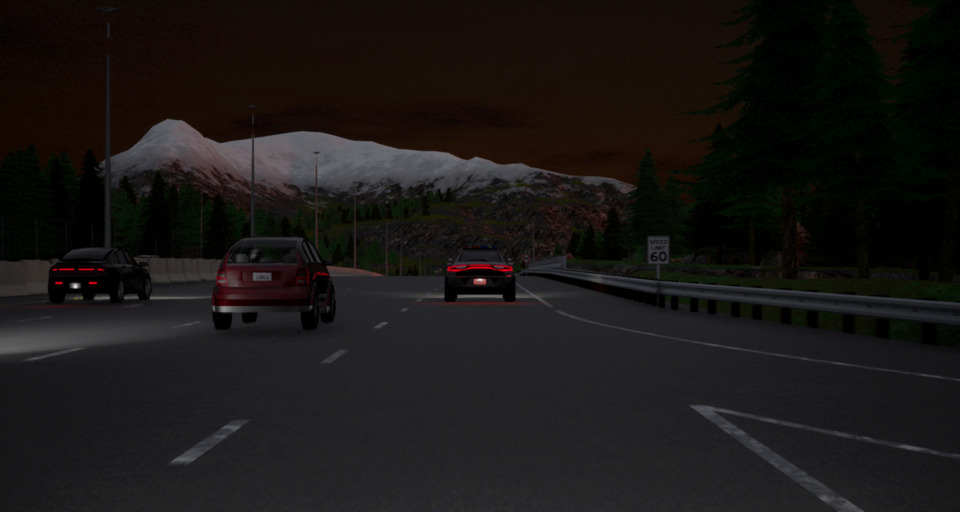} \\[-4pt]
    \includegraphics[width=2.8cm]{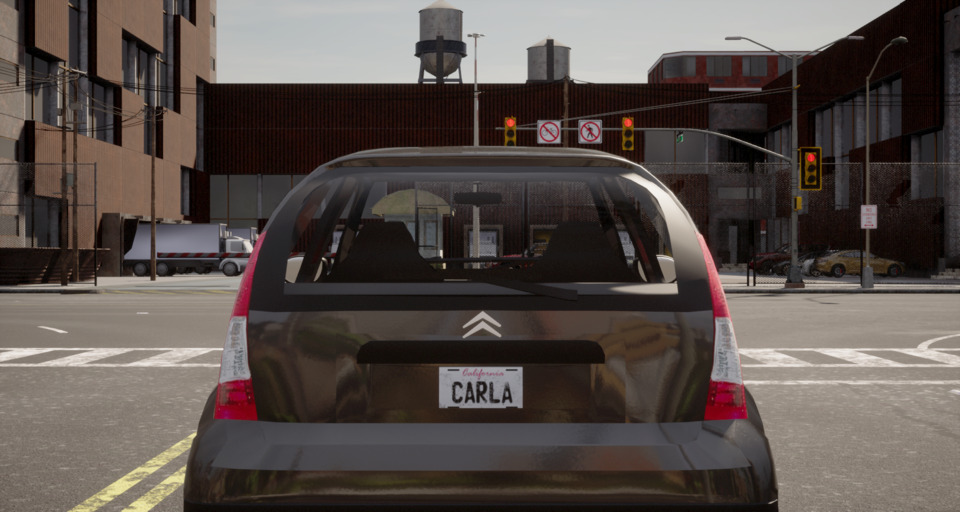} & \includegraphics[width=2.8cm]{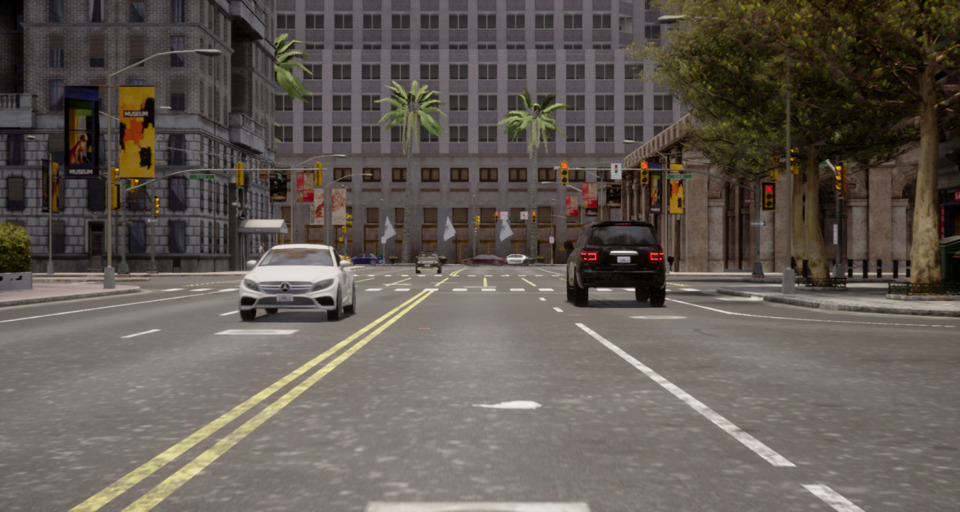} &
    \includegraphics[width=2.8cm]{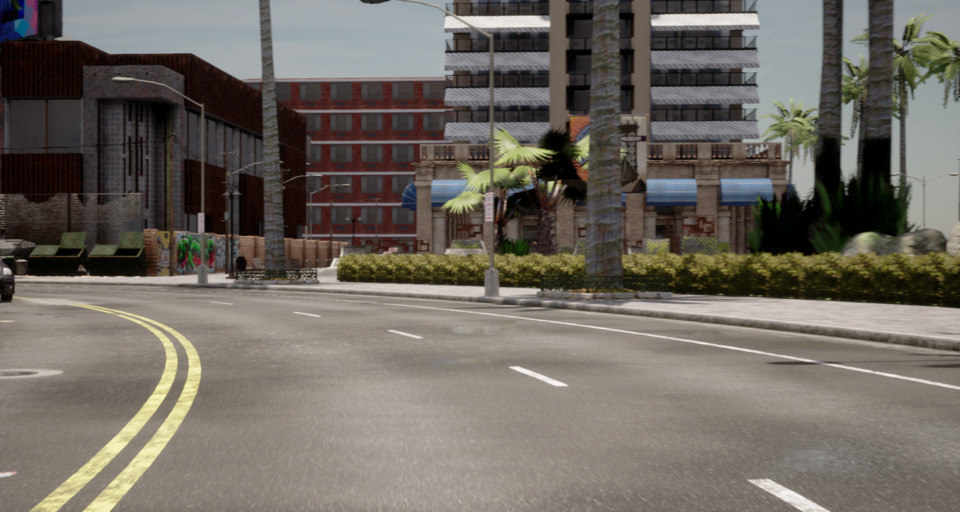} &
    \includegraphics[width=2.8cm]{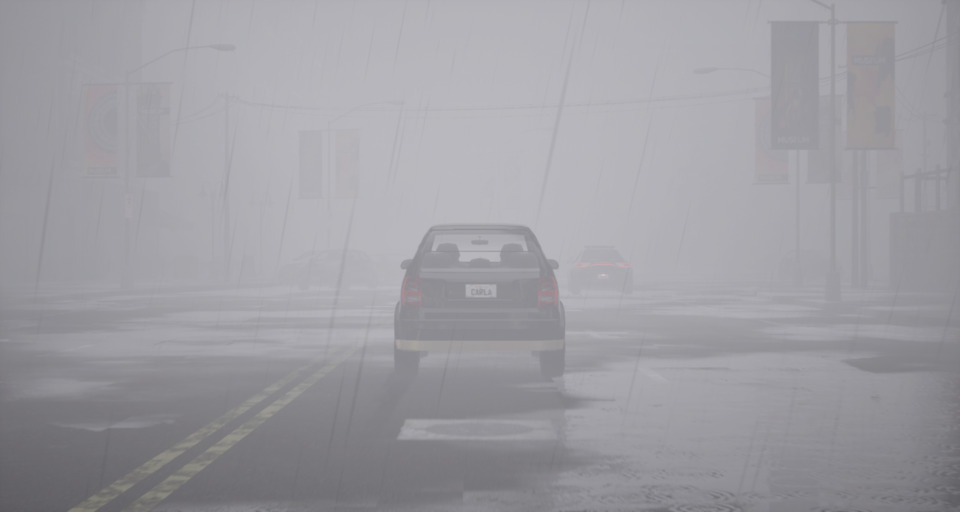} &
    \includegraphics[width=2.8cm]{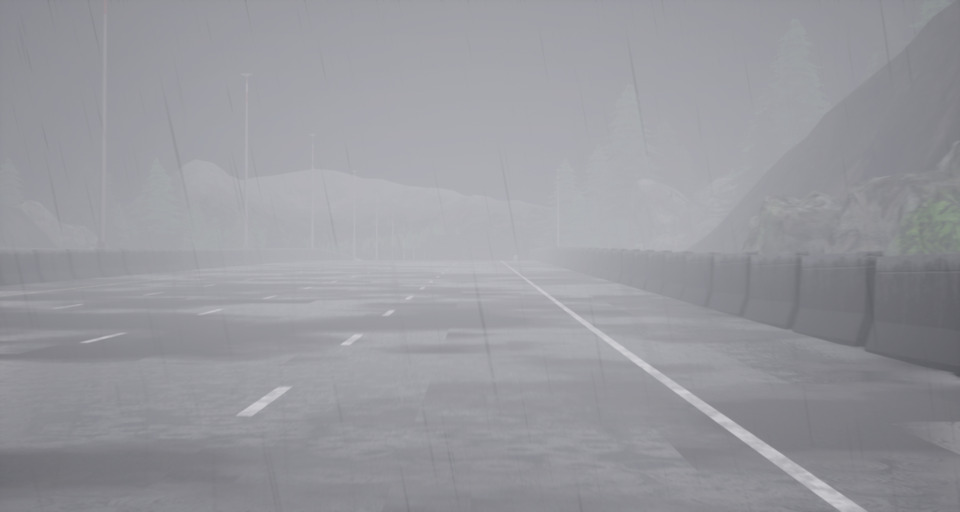} \\[-4pt]
    \includegraphics[width=2.8cm]{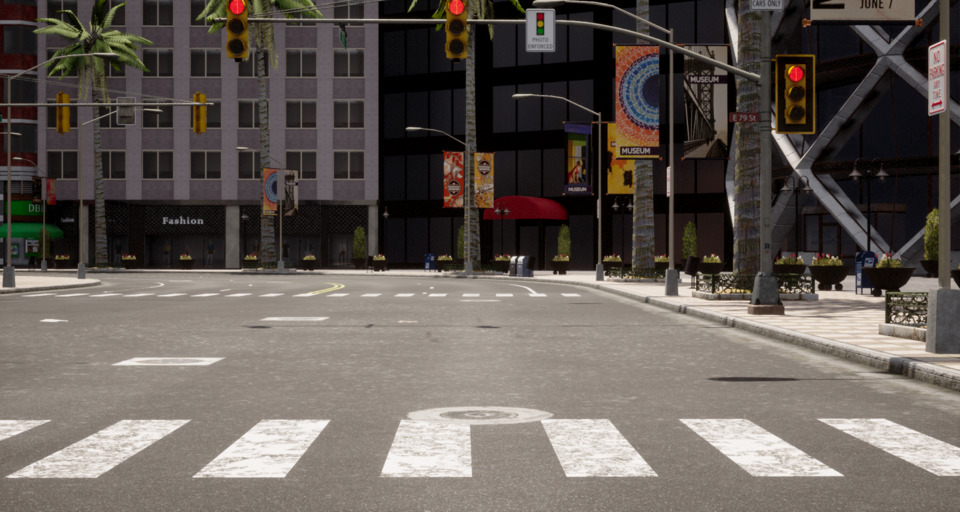} & \includegraphics[width=2.8cm]{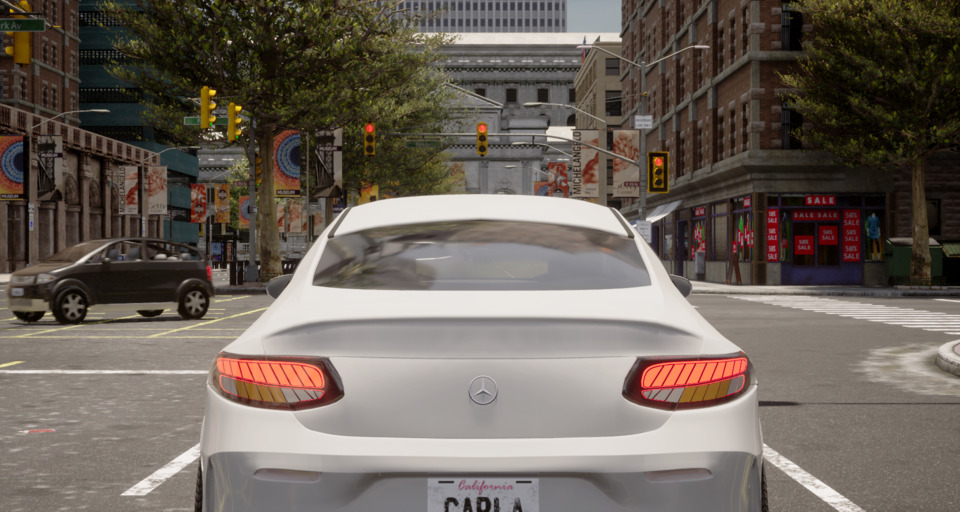} &
    \includegraphics[width=2.8cm]{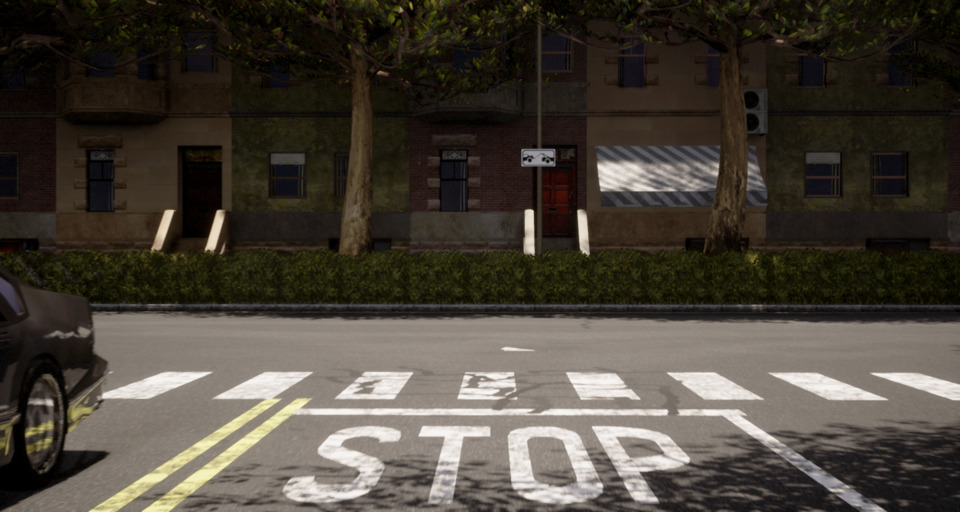} &
    \includegraphics[width=2.8cm]{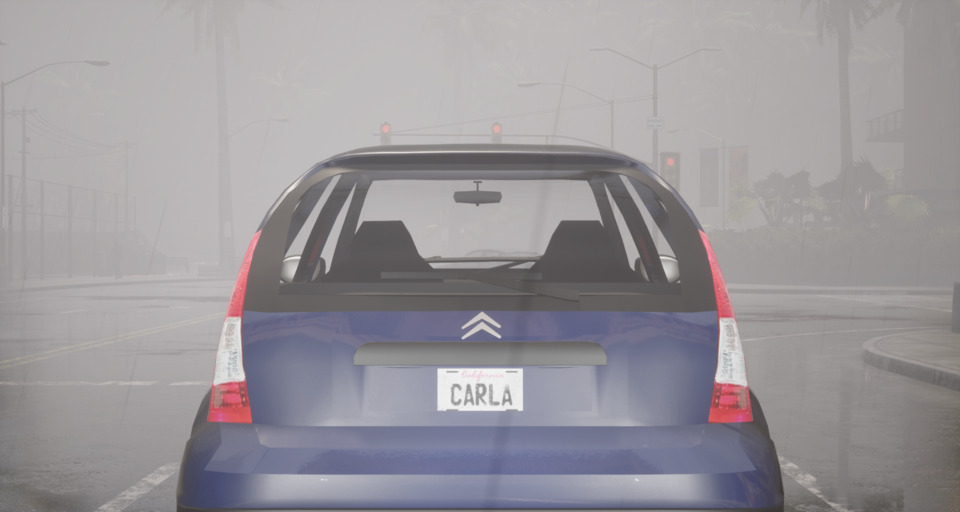} &
    \includegraphics[width=2.8cm]{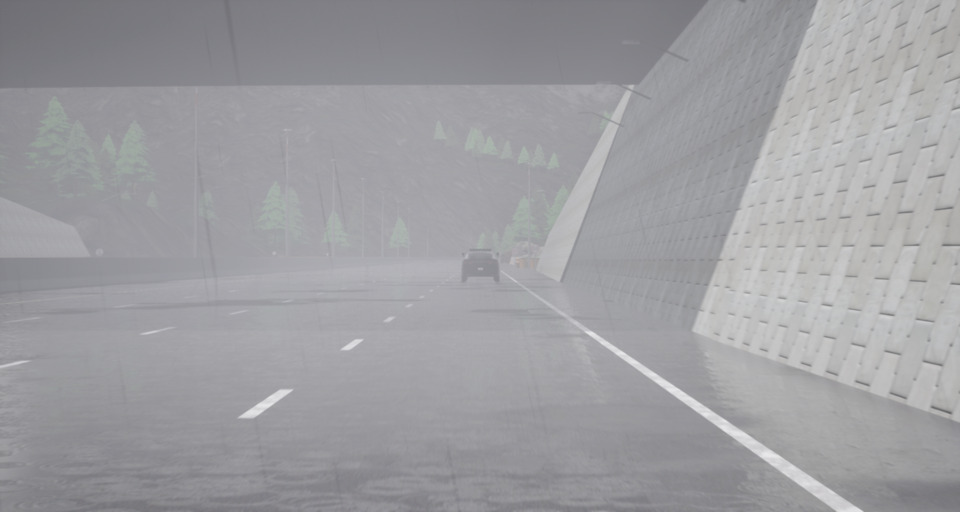} \\[-4pt]
    \includegraphics[width=2.8cm]{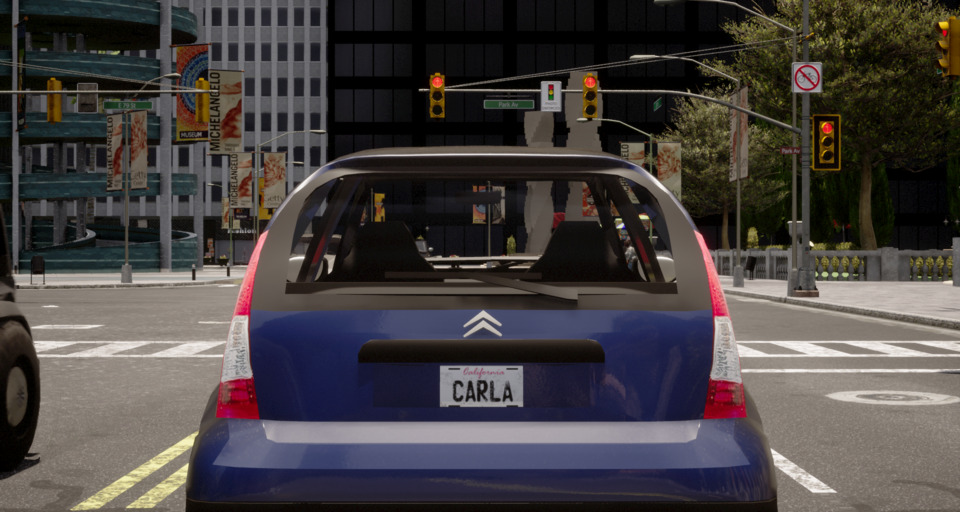} & \includegraphics[width=2.8cm]{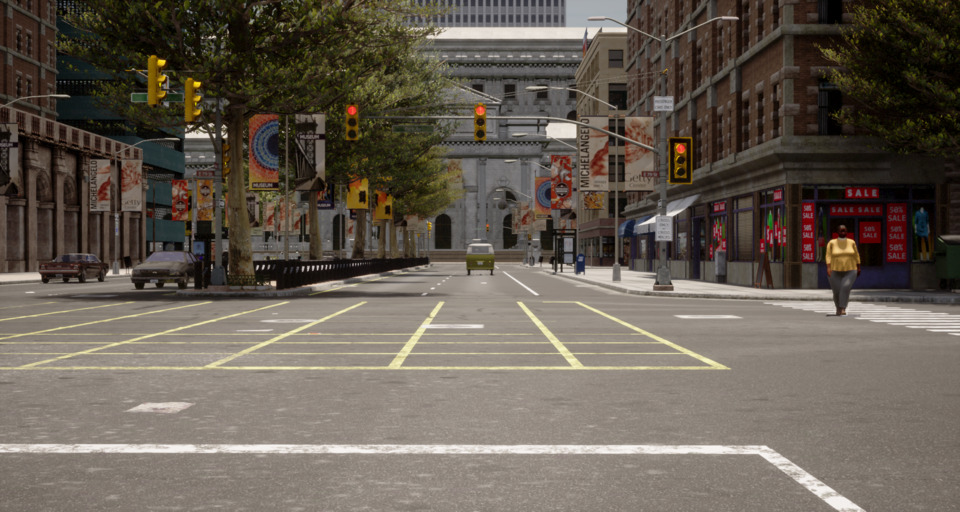} &
    \includegraphics[width=2.8cm]{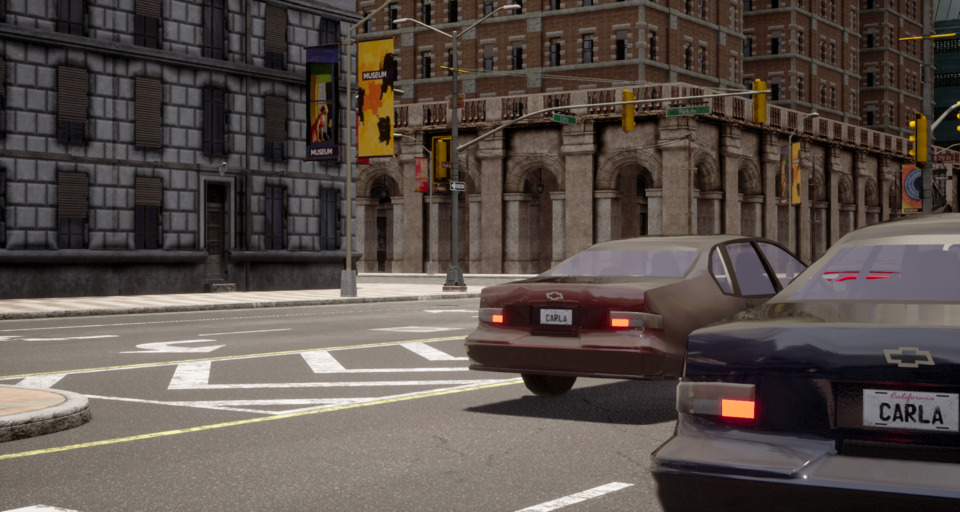} &
    \includegraphics[width=2.8cm]{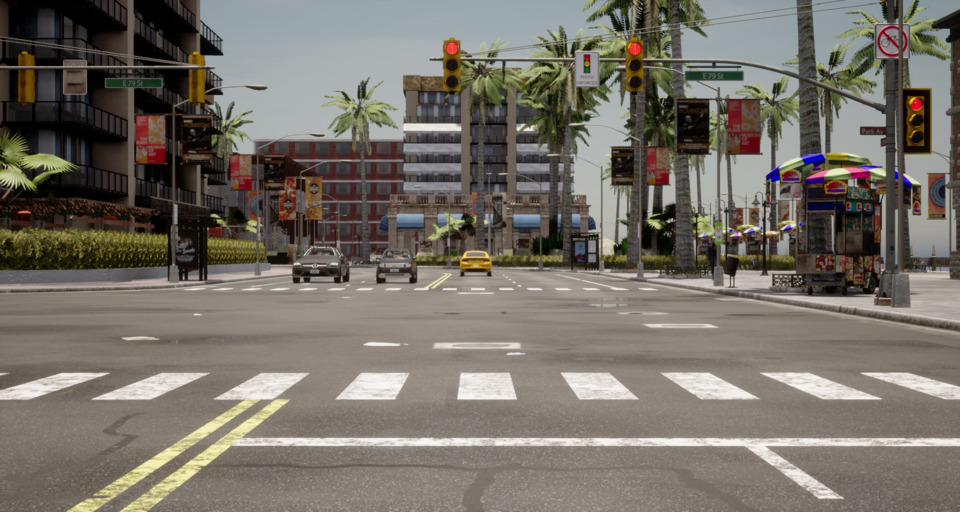} &
    \includegraphics[width=2.8cm]{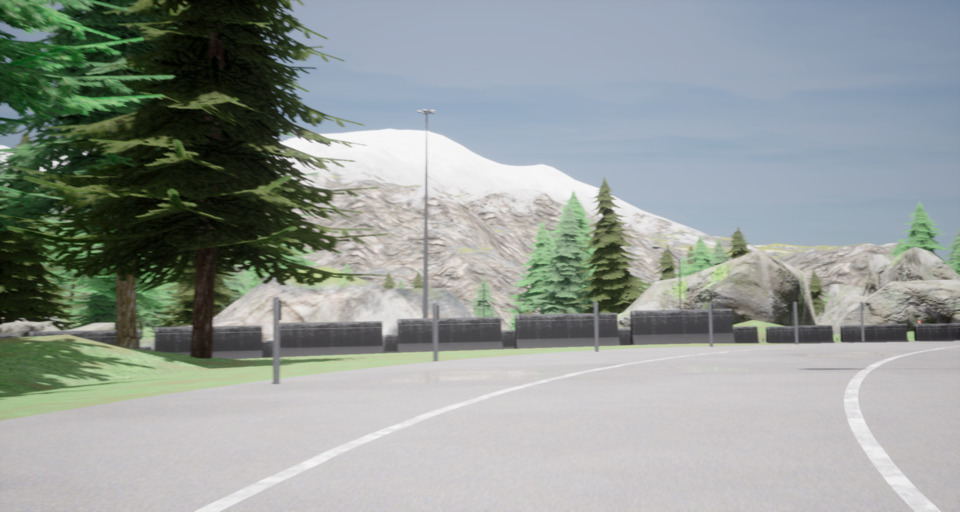} \\[-4pt]
  \end{tabular}
  \caption{Sequence for the different domain changes. Every 100th example is shown.}
  \label{fig:carla_dataset_appendix}
\end{figure}

\paragraph{day2night, clear2fog, clear2rain, dynamic}
Different domain changes are introduced by the sequences \textit{day2night}, \textit{clear2fog}, \textit{clear2rain}, and \textit{dynamic}. Each sequence starts with the weather parameters of \textit{clear} and follows the behavior illustrated in Figure \ref{fig:weather_params}. The weather model is based on the implementation of CARLA\footnote{\url{https://github.com/carla-simulator/carla/blob/0.9.13/PythonAPI/examples/dynamic\_weather.py}}. The maximum values used for the night, rain, and fog stetting are depicted in Table \ref{tab:clear_parameters}, with a bar indicating no change from the default parameters of \textit{clear}. While the default sequence length is 1200 samples, \textit{dynamic} contains 6000 samples to have the capability to study long-term behavior. Figure \ref{fig:carla_dataset_appendix} shows every 100th sample of the mentioned domain shifts.

\paragraph{day2night-slow and dynamic-slow}
\textit{day2night-slow} and \textit{dynamic-slow} are only considered for the gradual domain shift ablation study. Compared to the regular sequences, where 1200 samples correspond to one complete day-night cycle, in the slow setting, 4800 samples correspond to one day-night cyle. This enables the investigation of smaller domain shifts.

\paragraph{highway}
Since Town10HD does not contain any other settings than urban scenery, we use Town04 for the \textit{highway} sequence. For the corresponding source dataset, we use sequences generated from Town02 (urban setting, similar to Town04) and sequences from Town04 where the ego vehicles only drives in the city. The source dataset contains overall 3000 train and 500 test samples. The \textit{highway} sequence starts in the city of Town04 and shortly after continues on the highway. It follows the same weather behavior as \textit{dynamic}. Examples of the sub-sampled sequence are visualized in the last column in Figure \ref{fig:carla_dataset_appendix}.


\vfill

\end{document}